\documentclass[a4paper,onecolumn,12pt]{elsarticle}

\usepackage[T1]{fontenc}
\usepackage[english]{babel}
\usepackage{tgbonum} 
\usepackage{amsmath,amssymb,amsthm,mathrsfs,bm}
\usepackage{graphicx}
\usepackage{epstopdf}
\usepackage{booktabs}
\usepackage{multirow}
\usepackage{tabularx}
\usepackage{siunitx}
\usepackage[table]{xcolor}
\usepackage{float}
\usepackage{subcaption}
\usepackage{rotating}
\usepackage{algorithm}
\usepackage{algpseudocode}
\usepackage{enumitem,enumerate}
\usepackage{eucal}
\usepackage{dsfont}
\usepackage{ragged2e}
\usepackage{xcolor}
\usepackage[citecolor=blue]{hyperref}

\usepackage[legalpaper, lmargin=.5in, rmargin=.5in, tmargin=0.8in, bmargin=0.8in]{geometry}

\usepackage[labelfont=,labelsep=period]{caption}
\newcommand\highlightReference[1]{%
  \expandafter\newcommand\csname highlightReference-#1\endcsname{}%
}
\let\oldbibitem\bibitem
\def\bibitem#1 #2\par{%
  \expandafter\ifx\csname highlightReference-#1\endcsname\relax
    \oldbibitem{#1}#2\par
  \else
    \oldbibitem{#1}\highlight{#2}\par
  \fi
}
\newcommand\highlight[1]{\textcolor{blue}{#1}}

\allowdisplaybreaks

\begin{document}
	\begin{frontmatter}
		\title{A-PINN: Auxiliary Physics-informed Neural Networks for Structural Vibration Analysis in Continuous Euler-Bernoulli Beam}
		
		\author[a]{Shivani Saini}
		\ead{shivanis.nith@gmail.com}
		
		\author[a]{Ramesh Kumar Vats}
		\ead{rkvats@nith.ac.in}
		
		\author[b]{Arup Kumar Sahoo$^{*}$}
		\ead{arupnitr.jrfmath@gmail.com}

		\cortext[cor3]{Corresponding author}
		\affiliation[a]{organization={Department of Mathematics and Scientific Computing,\\ National Institute of Technology Hamirpur},
			postcode={177005},
			city={Himachal Pradesh},
			country={India}}
			
			\affiliation[b]{organization={Hatter Department of Marine Technologies, University of Haifa},
				postcode={3498838},
				city={Haifa},
				country={Israel}}

\begin{abstract}\justifying	
Recent advancements in physics-informed neural networks (PINNs) and their variants have garnered substantial focus from researchers due to their effectiveness in solving both forward and inverse problems governed by differential equations. In this research, a modified Auxiliary physics-informed neural network (A-PINN) framework with balanced adaptive optimizers is proposed for the analysis of structural vibration problems. In order to accurately represent structural systems, it is critical for capturing vibration phenomena and ensuring reliable predictive analysis.
So, our investigations are crucial for gaining deeper insight into the robustness of scientific machine learning models for solving vibration problems. Further, to rigorously evaluate the performance of A-PINN, we conducted different numerical simulations to approximate the Euler-Bernoulli beam equations under the various scenarios. The numerical results substantiate the enhanced performance of our model in terms of both numerical stability and predictive accuracy. Our model shows improvement of at  least 40\% over the baselines.

\begin{figure}[H]
	\centering
	\fbox{\includegraphics[width=17cm]{ 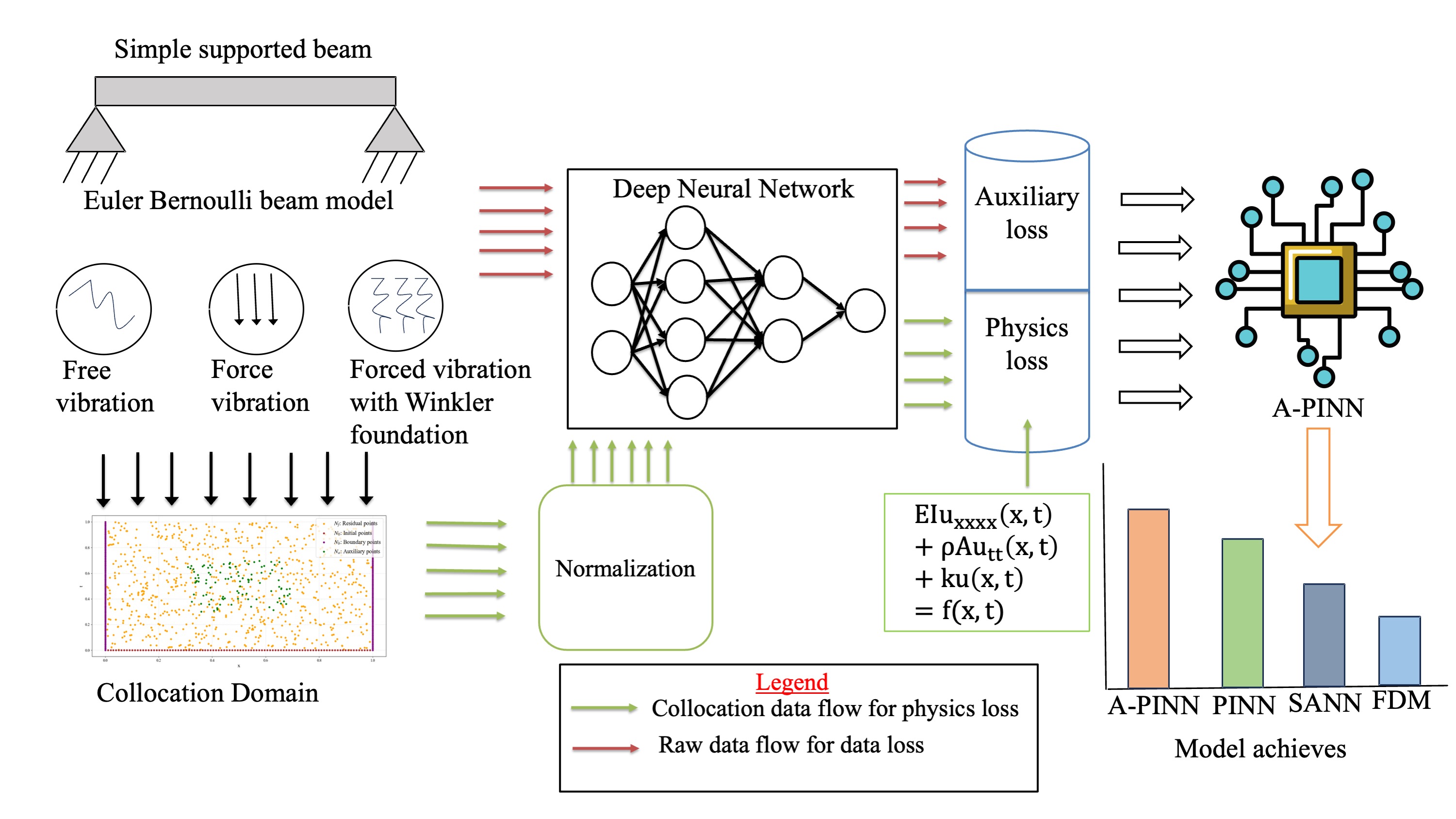}}  
	\caption*{Graphical abstract of A-PINN}
\end{figure}
\setcounter{figure}{0}
\end{abstract}

\begin{keyword}\justifying
	Scientific Machine Learning;  Auxiliary Physics-informed Neural Networks; Euler-Bernoulli Beam; Structural Vibration; Adaptive Optimization. 
\end{keyword} 
\end{frontmatter}

\begin{table}[H]
	\centering
	\textbf{List of Abbreviations} \\[2mm] 
	\begin{tabular}{ll}
		\hline
		\textbf{Abbreviation} & \textbf{Definition} \\
		\hline
		EBB   & Euler-Bernoulli Beam \\
		DOF   & Degree of Freedom \\
        TVB  & Transverse Vibration of Beam\\
        DE & Differential Equation\\
		PDE & Partial Differential Equation\\
		BC    & Boundary Condition \\
		IC    & Initial Condition \\
		GT  &  Ground Truth\\
		FDM & Finite Difference Method\\ 
	
		AI & Artificial Intelligence\\
        ML & Machine Learning\\
        SciML & Scientific Machine Learning\\
        
		ANN & Artificial Neural Network\\
		PINN & Physics-informed Neural Network \\
		A-PINN & Auxiliary Physics-informed Neural Network \\
        SANN & Symplectic Artificial Neural Network\\
		AD    & Automatic Differentiation \\
        AE & Absolute Error\\
		MSE   & Mean Square Error \\
		RMSE  & Root Mean Square Error \\
		MAE   & Mean Absolute Error \\
		\hline
	\end{tabular}
\end{table}

\section{Introduction}\justifying
Artificial intelligence (AI) is not the future; it is the present. The world is shifting towards fully functional AI services through various national projects, including the UAE's AI Strategy 2031, Israel's National AI Program, India's AI for All initiative, the Pan-Canadian AI Strategy, the USA's Winning the AI Race, and the UK's National AI Strategy, among others. As such, significant attention has shifted toward research in AI and automation. Among AI techniques, artificial neural networks (ANNs) stand out as widely used tools due to their countless applications in fields such as data mining, robotics, face detection, pattern recognition, and many other applications. Additionally, ANN has increasingly been adopted in high-stakes domains such as healthcare systems, weather forecasting, airport traffic management, military operations, and modern age warfare, where inaccuracy can profoundly impact human lives \cite{goodfellow2016deep,kumar2025comparative, song2024admm, kumar2025robust}. This demonstrates their adaptability and efficiency in resolving challenging issues.  When dealing with real-world dynamical phenomena, such as structural dynamics, fluid dynamics, wave propagation, and robotic motion, the integration of AI with mathematical modeling becomes especially beneficial.

The research in the vibration spans a wide range of domains, including structural health monitoring \cite{randall2021vibration}, dynamic response prediction \cite{skudrzyk1980mean}, stability assessment \cite{chu2018human}, and mechanical and civil engineering systems \cite{caetano2011vision}. Mechanical systems usually consist of the structural components, like rods, plates, beams, and shells, that possess distributed mass and elasticity. Unlike lumped parameter systems with finite degrees of freedom (DOF), these structural elements are continuous systems characterized by an infinite number of DOF, and their dynamic behavior is described by partial differential equations (PDEs). Analyzing vibrations in such systems is essential to ensure the performance and reliability, which leads to a wide range of research for developing analytical, numerical, and computational approaches. Beam-type structures are widely used in many branches of the mechanical, civil, and aerospace engineering \cite{rao2019vibration}. Among these problems, the transverse vibrations of the beam (TVB) hold fundamental importance in structural engineering. Moreover, the governing equation for TVB is a fourth-order PDE derived from the classical Euler-Bernoulli Beam (EBB) Model \cite{rao2019vibration, weaver1991vibration}, which assumes small deflections, linear elastic behavior, and negligible shear deformation. Solutions to this equation depend on boundary conditions (BCs); simply supported, clamped, and free ends produce characteristic mode shapes and natural frequencies that inform more complex structural analyses. For decades, the finite element method (FEM) \cite{wu2013analytical}, finite difference method (FDM) \cite{thomas2013numerical}, differential quadrature method (DQM) \cite{jena2018free}, and spectral methods \cite{goldman1999vibration} have been used for solving beam vibration problems. While effective, these approaches face challenges including mesh-based domains, high computational costs, discretization errors, and reduced efficiency for high-frequency or long-duration simulations. These limitations highlight the need for alternative approaches capable of combining physical fidelity with computational efficiency \cite{yadav2025application}.

On the other hand, vibration equations in structural dynamics have been solved with greater accuracy using purely data-driven neural networks (NNs) \cite{alli2003solutions}. However, if the dataset is noisy or scarce, their performance suffers. To overcome these challenges, in 2019, a research group from Brown university introduced groundbreaking work in the domain of scientific machine learning (SciML), viz. PINNs \cite{raissi2019physics}. Unlike conventional data-driven NNs, PINNs embed the governing physical laws in the form of DEs, along with available BCs and initial conditions (ICs), directly into the objective function of the NN \cite{kumar2025deep}. This allows PINNs to approximate the vibration solutions with improved generalization and reduced dependence on the large datasets. Beyond structural dynamics, PINNs are effectively utilized across a broad spectrum of problems, including fluid dynamics \cite{cai2021physics, kumar2023physics}, structural mechanics \cite{sahoo2024unsupervised}, biomedical modeling \cite{roquemen2025recent}, wave propagation \cite{chakraverty2025artificial, Rao2024}, and metamaterials \cite{chen2020physics}. Crucially, PINNs enable NNs to move beyond the data-driven, providing powerful tools for scientific computing and numerical simulation, particularly in cases where conventional methods face limitations of scalability and data quality. 

Even though PINNs are becoming more and more popular, they frequently experience stability problems and long-term error accumulation, particularly when subjected to external forcing factors.  A-PINNs combined with balanced adaptive optimizers and an extra auxiliary loss function have been suggested as a solution to these gaps. It is generally known that the choice of optimizer, in a composite loss function integrated architecture, can significantly influence the performance of NNs. During training, the optimizer adaptively modifies the contribution of each loss component, effectively balancing differences in magnitude among multiple loss components and improving convergence stability. The efficiency of our proposed model has been validated by solving the EBB problem under classical BCs, demonstrating its ability to accurately model beam vibrations while mitigating long-term error accumulation. Moreover, we focus on the simply supported beams subjected to three types of undamped EBB dynamic scenarios: free vibration, forced vibration, and vibration on a Winkler foundation. The A-PINN integrates auxiliary variable strategies designed to control error growth, improve stability, and enhance predictive accuracy across all three EBB models. The results demonstrate that A-PINN provides more reliable and robust predictions compared to conventional PINNs, making it well-suited to serve as an effective alternative to traditional numerical solvers for structural vibration analysis.

The rest of the paper is organized as follows: Section \ref{sec2} provides an overview of prior studies employing PINNs for vibration analysis and related ODE and PDE-based problems. Section \ref{sec4} presents the formulation of the loss function for the proposed modified adaptive optimizer A-PINN framework. Section \ref{sec5} reports three case studies of the EBB model, where fourth-order PDEs are used to evaluate the approximation capability of A-PINN. Section \ref{sec6} provides a detailed discussion of the obtained results, and Section \ref{sec7} concludes the paper with final remarks and potential directions for future research. For interested readers, the theory of FDM and symplectic artificial neural network (SANN)  and the absolute error (AE) results for the three EBB problems are given in Appendix A and Appendix B.

\section{Related Work}\label{sec2}\justifying

This section reviews prior research on the application of PINNs for vibration analysis, with particular emphasis on EBB theory and structural dynamics.  
Recent research in vibration analysis has increasingly adopted PINNs for solving PDEs arising in structural dynamics. Early studies benchmarked PINNs against classical methods, showing that embedding physical constraints within the loss function can significantly reduce dependence on labeled data.

Over time, advancements in PINNs have motivated researchers to incorporate nonlinearity and develop various extensions to address the limitations of the PINN framework. Xu et al.\cite{xu2023transfer} were among the first to propose a new variant of PINN for solving inverse problems. In a pioneering work, they proposed a transfer learning-based PINN framework for solving inverse problems in engineering structures under different types of loading scenarios. The combination of multitask learning and uncertainty weighting improves training efficiency and accuracy. It estimates external loads using limited displacement measurements and enables rapid adaptation to new structures through transfer learning. The study by Chen et al.~\cite{chen2023second} proposed an ML-based structural analysis using PINNs for the second-order analysis of steel beam columns. The approach integrates physical constraints into the learning process, enabling accurate predictions with minimal data and improved numerical efficiency through adaptive loss weighting and transfer learning.

Following the development of PINN, several variants have been proposed to incorporate domain decomposition, adaptive sampling, hard enforcement of ICs and BCs, and specialized activation functions. Each of these extensions enhances certain aspects of the PINN framework, enabling it to tackle more challenging PDEs. Recently, Kapoor et al.~\cite{kapoor2024transfer} introduced a causal PINN with transfer learning to simulate EBB, and Timoshenko beams with Winkler foundations. By reusing weights of related problems instead of training from scratch, their method improved generalizability, accelerated convergence, and outperformed conventional PINNs, even in extended spatio-temporal domains with noisy data. Zhou et al.~\cite{zhou2024data} presented a two-phase framework, Data Guided-PINN (DG-PINNs), combining data-driven and physics-informed approaches to efficiently solve inverse problems in the EBB equation. Zhang et al.~\cite{zhang2020differences} compared EBB and the Timoshenko beam for analyzing the effects of moving loads on periodically supported beams, such as railway tracks. Their results showed that the EBB model underestimated parametric excitation by a factor of about three compared to the Timoshenko beam due to shear deformation effects, which become significant for short span lengths. A 2.5D finite element model was also used as a reference, demonstrating closer agreement with the Timoshenko formulation. In Söyleyici et al.~\cite{soyleyici2025physics}, a physics-informed deep neural network (DNN) has demonstrated success in simulating beam vibrations and estimating system parameters, confirming the effectiveness of neural networks in capturing complex dynamic responses. Kapoor et al.~\cite{kapoor2024physics} further developed a PINN approach for both forward and inverse analyses of EBB and Timoshenko beam systems, achieving accurate and computationally efficient predictions.

Several studies have investigated stochastic, FEM, and meshless approaches for the analysis of micro and nanobeam structures. Jena and Chakraverty~\cite{jena2024structural} researched in the direction of uncertainty modeling in nanobeams, microbeams, and functionally graded beams, with emphasis on vibration and stability analysis under imprecisely defined parameters. Olawale et al.~\cite{olawale2023response} studied stochastic responses of EBB under random external forces using FDM combined with Monte Carlo simulations. Ye et al.~\cite{ye2025mixed} employed a mixed FEM to study vibration equations of structurally damped beams and plates, establishing convergence and error estimates for both $L^2$-norms and $H^1$-norms. Cheng et al.~\cite{cheng2025exponential} developed a space semi-discretized finite difference scheme for axially moving EBB under nonlinear boundary feedback control, ensuring uniform exponential stability. However, all these models are problem-dependent and mesh-based. Xiao et al.~\cite{xiao2025meshless} proposed a meshless Runge-Kutta-based PINN framework for structural vibration analysis, improving temporal flexibility, reducing computational cost, and efficiently handling high-frequency vibrations in EBB problems. All the above discussed PINN models, along with some other variants, are listed in Table \ref{PINN_variants}. 

According to our literature review, PINNs are effective for solving forward, inverse, and parameter identification problems in EBB dynamics, covering scales from nanobeams to large-scale moving-load structures. However, open challenges remain in handling higher-order derivatives, ensuring numerical stability in long-duration simulations, and achieving robust generalization under diverse boundary and load conditions. So to fill these gaps, we apply both the PINN and our A-PINN framework to solve the EBB equation. The A-PINN introduces auxiliary variables to systematically address higher-order vibration equations, which improves the accuracy and stability of structural dynamics simulations. The contributions of our research can be summarized as follows:

\begin{itemize}

\item A-PINN reduces the steepness of the target solution, allowing the NN to more effectively exploit its approximation capability.  

\item A-PINN decreases the sensitivity to network architecture design, including choices of width, depth, and activation functions. Even simple NNs can achieve satisfactory accuracy under this formulation.  

\item An adaptive optimization strategy combining the Adam and L-BFGS algorithms is employed to ensure faster convergence and superior generalization. The adaptive switching between these optimizers balances global exploration and local refinement during training.

\item A-PINN preserves the fundamental principles of the original PINN framework, while the proposed modifications are lightweight, generic, and simple to implement.

\end{itemize}

\begin{table}[H]
\centering
\caption{Representative variants of PINN reported in the literature.}
\label{PINN_variants}
\scriptsize
\begin{tabularx}{\textwidth}{lXX}
\toprule
\textbf{Variant (with Reference)} & \textbf{Features} & \textbf{Main Approach} \\
\midrule

PINN~\cite{raissi2019physics} 
& Foundational deep learning framework coupling data and physics. 
& Embeds PDE residuals, ICs, and BCs into a unified composite loss to solve both forward and inverse problems. \\

XPINN~\cite{jagtap2020extended} 
& Space-time domain decomposition for nonlinear PDEs. 
& Uses multiple subdomain networks for parallelization and improved efficiency over PINN and cPINN. \\

hpPINN~\cite{kharazmi2021hp} 
& Adaptive accuracy through $h$ and $p$ refinements. 
& Combines domain partitioning ($h$-refinement) with higher-order polynomial basis enrichment ($p$-refinement) for precision control. \\

gPINN~\cite{yu2022gradient} 
& Gradient-enhanced residual learning for smooth convergence. 
& Includes first and higher-order derivatives in the loss to improve accuracy for stiff or rapidly varying PDE solutions. \\

PINN-DeepONet~\cite{wang2021learning} 
& Physics-informed operator learning. 
& Uses PDE-driven residual minimization within DeepONet to approximate nonlinear solution operators efficiently. \\

VT-PINN~\cite{zheng2024vtpinn} 
& PINN improved via variable transformation. 
& Applies variable substitutions to reformulate PDEs with smoother solutions, improving training stability and accuracy for both forward and inverse problems. \\

CL-PINN~\cite{sahoo2025cl} 
& Curriculum learning-based PINN for geophysical wave modeling. 
& Combines curriculum learning with PINNs to solve Boussinesq-type equations for tsunami propagation, improving accuracy and training efficiency. \\

wbPINN~\cite{cao2025wbpinn} 
& Weight-balanced adaptive PINN for multi-objective learning. 
& Employs correlation and penalty terms with adaptive loss weighting to balance multi-objective optimization, enhancing convergence and accuracy on complex PDEs. \\

ADMM-PINN~\cite{song2024admm} 
& ADMM-based framework for nonsmooth PDE-constrained optimization. 
& Combines ADMM with PINNs to decouple PDE constraints and regularization terms for efficient and scalable optimization. \\

\bottomrule
\end{tabularx}
\end{table}


\section{Problem Statement}\label{sec3}
After outlining the broader context, this section introduces the nominal model based on the EBB theory \cite{rao2019vibration}, commonly referred to as the classical beam theory. In this work, the undamped form of the EBB model is considered, neglecting any material or viscous damping effects to isolate the inherent vibrational characteristics of the structure. For dynamic cases, the formulation further incorporates inertial effects, elastic foundations, and external excitations.\\ 
Accordingly, for a beam subjected to a distributed transverse load \( F(x,t) \) and resting on a Winkler-type elastic foundation with stiffness \( k \), the governing equation of undamped transverse vibration is given by
\begin{equation}
	EI\,\frac{\partial^4 u(x,t)}{\partial x^4}
	+ \rho A\,\frac{\partial^2 u(x,t)}{\partial t^2}
	+ k\,u(x,t)
	= F(x,t),
	\label{general_beam}
\end{equation}
where \( u(x,t) \) denotes the transverse deflection of the beam. The physical quantities and their corresponding notations are summarized in Table~\ref{notation}.

\begin{table}[h!]
\centering
\caption{List of notations used in the Euler-Bernoulli beam model.}
\begin{tabular}{clc}
\hline
\textbf{Symbol} & \textbf{Description} & \textbf{Unit} \\
\hline
$E$         & Young’s modulus of elasticity                    & Pa \\
$I$         & Area moment of inertia of the cross-section      & m$^4$ \\
$\rho$      & Mass density of the beam material                & kg/m$^3$ \\
$A$         & Cross-sectional area of the beam                 & m$^2$ \\
$EI$        & Flexural rigidity                                & N·m$^2$ \\
$k$         & Winkler foundation stiffness per unit length      & N/m$^2$ \\
$u(x,t)$    & Transverse displacement                           & m \\
\hline
\end{tabular}
\label{notation}
\end{table}

Rearranging Eq.~\eqref{general_beam} gives:
\begin{equation}
	c^{2}\,\frac{\partial^{4} u}{\partial x^{4}}(x,t)
	+ \frac{\partial^{2} u}{\partial t^{2}}(x,t)
	+ \kappa\,u(x,t)
	= f(x,t),
	\label{scaled_beam}
\end{equation}
where
\[
c^{2} = \frac{EI}{\rho A}, 
\qquad 
\kappa = \frac{k}{\rho A}, 
\qquad 
f(x,t) = \frac{F(x,t)}{\rho A}.
\]

\begin{figure}[H]
    \centering

    \begin{subfigure}[b]{0.30\textwidth}
        \centering
        \includegraphics[width=\textwidth]{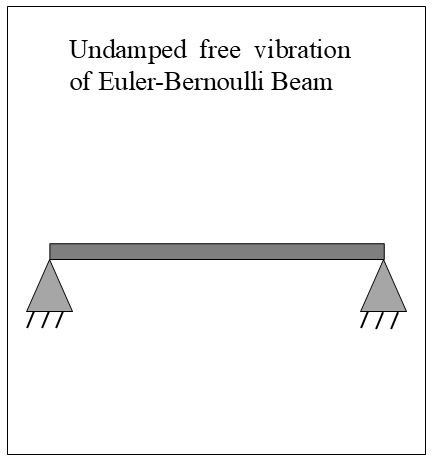}
        \caption{}
        \label{free}
    \end{subfigure}
    \hfill
    \begin{subfigure}[b]{0.30\textwidth}
        \centering
        \includegraphics[width=\textwidth]{ 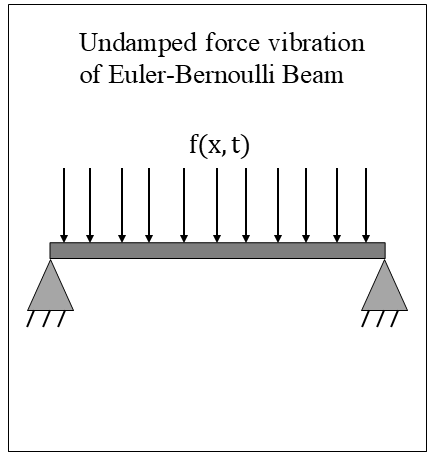}
        \caption{}
        \label{force}
    \end{subfigure}
    \hfill
    \begin{subfigure}[b]{0.30\textwidth}
        \centering
        \includegraphics[width=\textwidth]{ 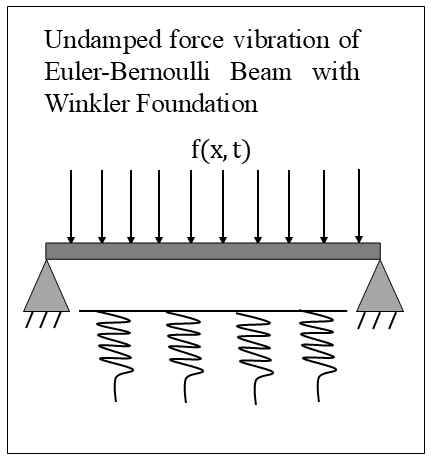}
        \caption{}
        \label{winkler}
    \end{subfigure}

    \caption{Schematic illustration of the simply supported EBB for three undamped vibration cases:
    (a) free, 
    (b) forced, 
    and (c) forced with a Winkler foundation.}
    \label{Simply supported}
\end{figure}

Where \(c\) is a stiffness, \(\kappa\) is the Winkler-foundation stiffness, and \(f(x,t)\) is an external transverse force. The present work focuses on analyzing three fundamental cases of transverse vibration for a simply supported EBB as presented in Figure \ref{Simply supported}. By systematically varying the governing equation, we distinguish undamped vibration cases - free vibration, forced vibration, and vibration with Winkler foundation, while maintaining identical ICs and BCs.

The three special cases studied here are obtained from Eq. \eqref{scaled_beam} by appropriately choosing \(\kappa\) and \(f(x,t)\):
\begin{itemize}
	\item If \(f(x,t) = 0\) and \(\kappa = 0\),  
	Eq.~\eqref{scaled_beam} reduces to the free vibration of the EBB as illustrated in Figure \ref{free}
	\begin{equation}
		c^{2}\,\frac{\partial^{4} u}{\partial x^{4}}(x,t) 
		+ \frac{\partial^{2} u}{\partial t^{2}}(x,t) = 0,
		\label{eq:free_vibration_scaled}
	\end{equation}
	
	\item  If \(\kappa = 0\) and \(f(x,t) \neq 0\),  
	Eq.~\eqref{scaled_beam} represents the EBB under external excitation as illustrated in Figure \ref{force} 
	\begin{equation}
		c^{2}\,\frac{\partial^{4} u}{\partial x^{4}}(x,t) 
		+ \frac{\partial^{2} u}{\partial t^{2}}(x,t) = f(x,t),
		\label{eq:forced_vibration_scaled}
	\end{equation}
	
	\item  If \(\kappa \neq 0\) and \(f(x,t) \neq 0\),  
	Eq.~\eqref{scaled_beam} becomes the EBB resting on a Winkler foundation with applied load as illustrated in Figure \ref{winkler}
	\begin{equation}
		c^{2}\,\frac{\partial^{4} u}{\partial x^{4}}(x,t) 
		+ \frac{\partial^{2} u}{\partial t^{2}}(x,t) 
		+ \kappa\,u(x,t) = f(x,t).
		\label{eq:winkler_scaled}
	\end{equation}
\end{itemize}

\noindent
The beam's displacement and velocity at $t = 0$ are given by the following ICs:

\begin{equation}
	\begin{aligned}
		u(x,0) &= \phi(x), \\
		\frac{\partial u}{\partial t}(x,0) &= \phi_{t}(x),
	\end{aligned}
\end{equation}
where \(\phi(x)\) and \(\phi_{t}(x)\) are prescribed functions representing the initial deflection and velocity distributions along the beam. 
The BCs correspond to a simply supported beam at $x=0$ and $x=L$. At the supports, the transverse displacement is zero and the bending moment vanishes, which is expressed mathematically as
\begin{equation}
	\begin{aligned}
		u(0,t) &= u(L,t) = 0, \\
		\frac{\partial^{2} u}{\partial x^{2}}(0,t) &= \frac{\partial^{2} u}{\partial x^{2}}(L,t) = 0.
	\end{aligned}
\end{equation}

\section{The Description of A-PINN}\label{sec4}
In this section, we describe the overall framework of the A-PINN. We begin with the basic PINN architecture based on DNNs, followed by the use of automatic differentiation (AD) for constructing physics-driven losses. Subsequently, the formulation of the A-PINN and the design of its composite loss function are presented in detail.

\subsection{PINN Architecture}
\begin{figure}[H]
	\centering
	\includegraphics[width=14cm]{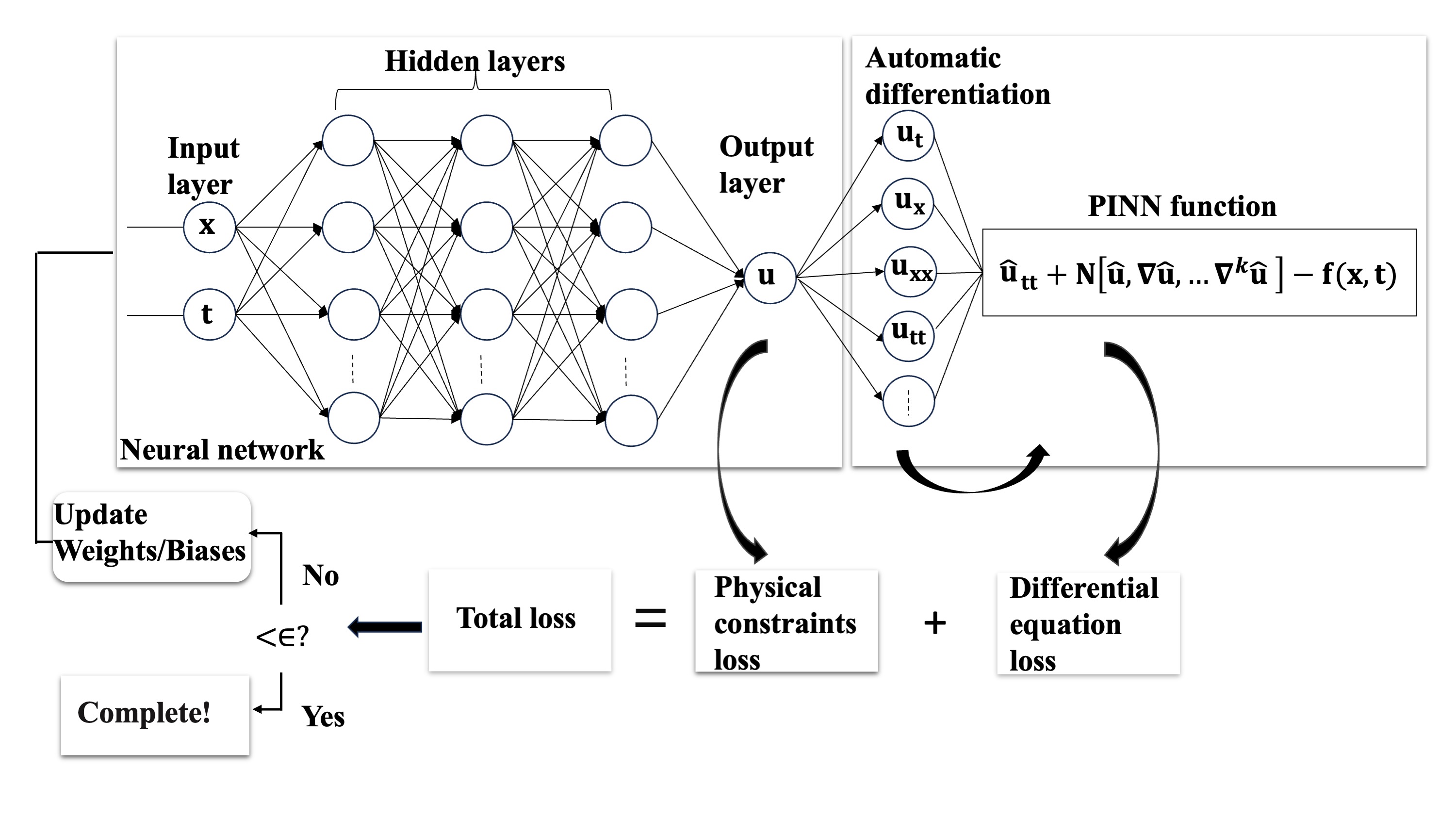} 
	\caption{Architecture of physics-informed neural network}
	\label{PINN}
\end{figure}

PINNs are a branch of SciML that improves the generalization of NN by integrating physics-based knowledge of the system during training. It is achieved by minimizing a composite loss function that calculates the difference between predicted and ground truth (GT). The loss function incorporates several problem-specific parameters, such as ICs and BCs, the spatial and temporal domain, and collocation points. 
Although not identical, the training strategy of PINN residual loss can be compared to unsupervised learning, since the model derives patterns and constraints from unlabeled data guided by physical laws. In contrast to most deep learning models, PINNs adopt a mesh-free approach, reformulating the governing DE into a loss minimization problem. In general, the architecture of a PINN is composed of three essential components: a NN that approximates the solution, an AD module that computes the necessary chain derivatives of the NN output with respect to the inputs, and a loss function that embeds the PDE residuals with the ICs and BCs. Figure~\ref{PINN} illustrates this generic architecture, highlighting how these components are placed and interact with one another.

\textbf{Deep Neural Networks:} 
A DNN is a fully connected feedforward architecture consisting of an input layer, several hidden layers, and an output layer. The connection between layers can be expressed as
\begin{align}
y_i &= \sigma\!\left(\mathbf{W}_i\,y_{i-1} + \mathbf{b}_i\right), \qquad 1 \le i \le n, \\
z   &= \mathbf{W}_{n+1}\,y_n + \mathbf{b}_{n+1},
\end{align}

where \(y_i\) denotes the output of the \(i\)-th layer, \(\sigma(\cdot)\) is the nonlinear activation function, and \((\mathbf{W}_i, \mathbf{b}_i)\) are the trainable weights and biases. The quantity \(z\) represents the final output of the NN.
 The DNN learns complex mappings through iterative optimization, and with activation functions, the derivatives of the NN output with respect to the inputs can be efficiently computed via AD.

\textbf{Automatic Differentiation:}
In PINNs, the solution of a DE requires the computation of derivatives of the NN output with respect to its input variables. AD computes derivatives systematically by applying the chain rule over a computational graph, yielding exact results up to machine precision while maintaining computational efficiency and scalability.  Unlike numerical differentiation, AD does not rely on approximations of neighbouring points. Instead, it decomposes the function into a sequence of differentiable sub-functions, with intermediate variables storing results at each stage. The chain rule is then applied in forward or backward mode to propagate derivatives through the NN.

In the context of PINNs as shown in Figure \ref{PINN}, AD is used to evaluate the required partial derivatives of the NN output $\hat{u}(\mathbf{x},t;\theta)$ with respect to spatial and temporal variables, where $\theta$ denotes the set of all trainable parameters of the NN, including its weights and biases. These derivatives are then substituted into the governing PDE, ICs, and BCs. The residual of the PDE is computed and incorporated into the loss function, which is minimized throughout the training process. This enables the NN to learn solutions aligned with both the data and the governing physics.

\textbf{Formulation of loss function:}
Consider a PDE on the domain $\Omega \subset \mathbb{R}^d$:

\begin{equation}\label{M_1}
	u_{tt}(\mathbf{x}, t) + \mathcal{N}[u(\mathbf{x}, t), \nabla u(\mathbf{x}, t), \ldots, \nabla^k u(\mathbf{x}, t)] = f(\mathbf{x}, t), \quad \mathbf{x} \in \Omega, \ t \in (0, T]
\end{equation}
\begin{equation}
	\mathcal{B}[u(\mathbf{x}, t), \nabla u(\mathbf{x}, t), \ldots, \nabla^m u(\mathbf{x}, t)] = g(\mathbf{x}), \quad \mathbf{x} \in \partial\Omega,
\end{equation}
\begin{equation}
	\mathcal{I}[u(\mathbf{x}, t), \partial_t u(\mathbf{x}, t), \ldots, \partial_t^r u(\mathbf{x}, t)] \Big|_{t=0} = h(\mathbf{x}), \quad \mathbf{x} \in \Omega,
\end{equation}
where, \( \mathcal{N}[\cdot] \) denote nonlinear differential operators, \( f(\mathbf{x}, t) \) is force term, \( g(\mathbf{x}) \) defines the BCs, and \( h(\mathbf{x}) \) specifies the ICs. The objective is to approximate solution of Eq. \eqref{M_1}. Let us \( \widehat{u}(\mathbf{x}, t;\theta) \) denote an approximate solution of the NN, which approximately satisfies:
\begin{equation}\label{R_PINN}
	\widehat{u}_{tt}(\mathbf{x}, t;\theta) + \mathcal{N}[\widehat{u}(\mathbf{x}, t;\theta), \nabla \widehat{u}(\mathbf{x}, t;\theta), \ldots, \nabla^k \widehat{u}(\mathbf{x}, t;\theta)] \approx f(\mathbf{x}, t), \quad \mathbf{x} \in \Omega, \ t \in (0, T].
\end{equation}
The residual \( \mathcal{R}(\mathbf{x}, t) \) of Eq. \eqref{R_PINN} is defined as:
\begin{equation}
	\mathcal{R}(\mathbf{x}, t) := 
	\widehat{u}_{tt}(\mathbf{x}, t;\theta) 
	+ \mathcal{N}[\widehat{u}(\mathbf{x}, t;\theta), \nabla \widehat{u}(\mathbf{x}, t;\theta), \ldots, \nabla^k \widehat{u}(\mathbf{x}, t;\theta)] 
	- f(\mathbf{x}, t).
\end{equation}
Here, the total loss function comprises four components, namely the PDE residual loss, BCs loss, ICs loss, and data loss. These are defined as
\begin{align}
\mathcal{L}_f(\theta) 
&= \frac{1}{N_f} \sum_{i=1}^{N_f} 
\Big| 
\widehat{u}_{tt}(\mathbf{x}_i^f, t_i^f;\theta)
+ \mathcal{N}\!\big[\widehat{u}(\mathbf{x}_i^f, t_i^f;\theta), 
\nabla \widehat{u}(\mathbf{x}_i^f, t_i^f;\theta), 
\ldots, 
\nabla^k \widehat{u}(\mathbf{x}_i^f, t_i^f;\theta)\big] 
- f(\mathbf{x}_i^f, t_i^f;\theta)
\Big|^2, \\[0.25em]
\mathcal{L}_b(\theta) 
&= \frac{1}{N_b} \sum_{i=1}^{N_b} 
\Big| 
\mathcal{B}\!\big[\widehat{u}(\mathbf{x}_i^b, t_i^b;\theta), 
\nabla \widehat{u}(\mathbf{x}_i^b, t_i^b;\theta), 
\ldots, 
\nabla^m \widehat{u}(\mathbf{x}_i^b, t_i^b;\theta)\big] 
- g(\mathbf{x}_i^b)
\Big|^2, \\[0.25em]
\mathcal{L}_0(\theta) 
&= \frac{1}{N_0} \sum_{i=1}^{N_0} 
\Big| 
\mathcal{I}\!\big[\widehat{u}(\mathbf{x}_i^0, 0;\theta), 
\partial_t \widehat{u}(\mathbf{x}_i^0, 0;\theta), 
\ldots, 
\partial_t^r \widehat{u}(\mathbf{x}_i^0, 0;\theta)\big] 
- h(\mathbf{x}_i^0)
\Big|^2, \\[0.25em]
\mathcal{L}_d(\theta) 
&= \frac{1}{N_d} \sum_{i=1}^{N_d} 
\Big| 
\widehat{u}(\mathbf{x}_i^d, t_i^d;\theta) - u(\mathbf{x}_i^d, t_i^d) 
\Big|^2.
\end{align}
The overall loss is expressed as a weighted combination of the above terms:
\begin{equation}
	\mathcal{L}(\theta) = \tilde{w}_f \, \mathcal{L}_f(\theta) 
	+ \tilde{w}_b \, \mathcal{L}_b(\theta) 
	+ \tilde{w}_0 \, \mathcal{L}_0(\theta) 
	+ \tilde{w}_d \, \mathcal{L}_d(\theta),
\end{equation}
where \( (\mathbf{x}_i^f, t_i^f;\theta) \), \( (\mathbf{x}_i^b, t_i^b;\theta) \), \( (\mathbf{x}_i^0, 0;\theta) \), and \( (\mathbf{x}_i^d, t_i^d;\theta) \) are collocation points for enforcing the PDE residual, BCs, ICs, and data supervision, respectively. 
The terms \( N_f \), \( N_b \), \( N_0 \), and \( N_d \) denote the number of training points for each component, and \( \tilde{w}_f \), \( \tilde{w}_b \), \( \tilde{w}_0 \), and \( \tilde{w}_d \) are adaptive scalar weights balancing influence of each loss term.

\subsection{Auxiliary Physics-informed Neural Networks}

\begin{figure}[H]
	\centering
	\includegraphics[width=15cm]{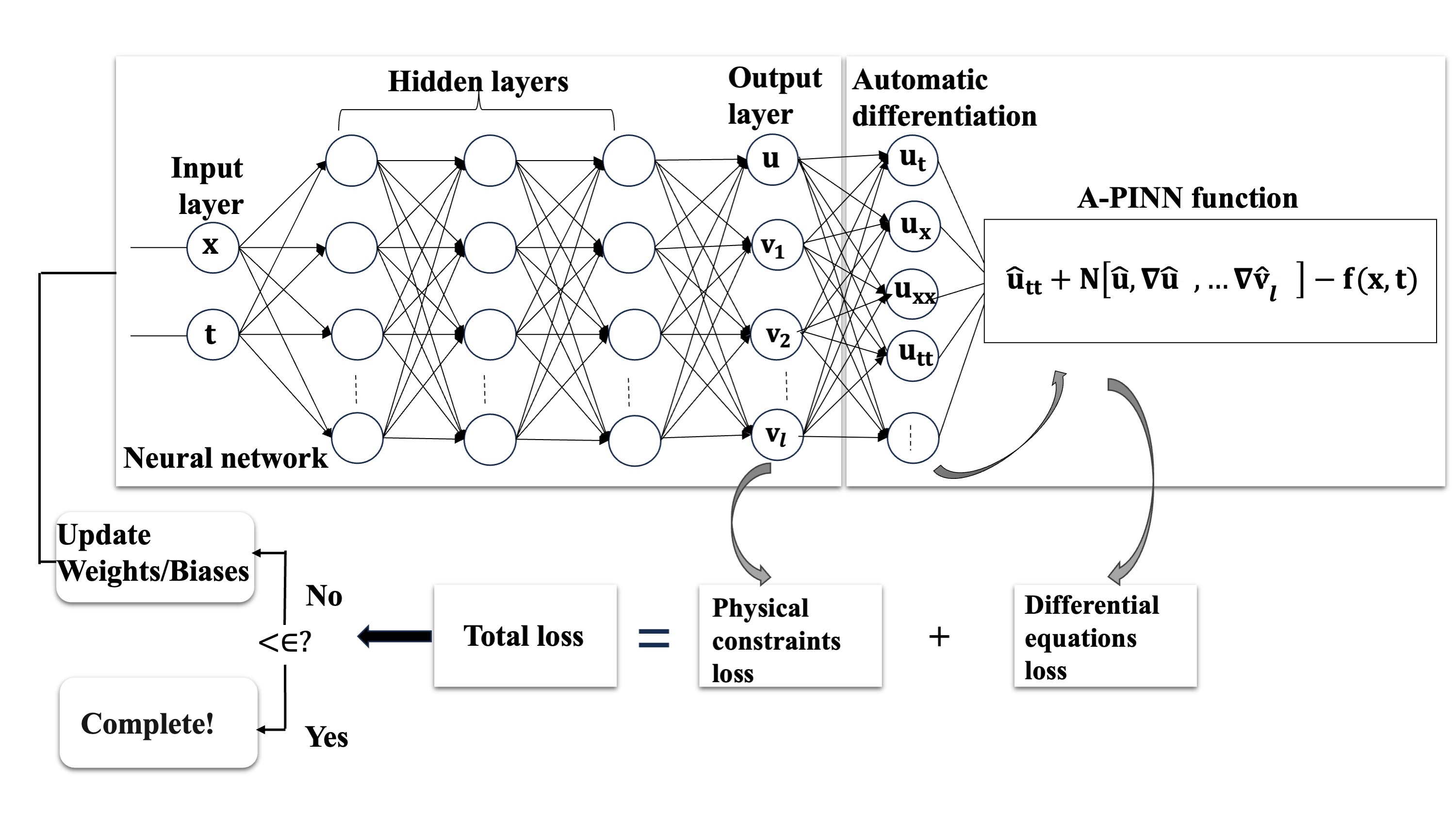}
	\caption{Architecture of Auxiliary physics-informed neural network}
	\label{A-PINN}
\end{figure}

A-PINNs extend the PINN framework to address the challenges in solving higher-order PDEs. In conventional PINNs, higher-order derivatives are obtained directly through AD, which often accumulates numerical errors and reduces both stability and accuracy during training. A-PINNs introduce the auxiliary variables that explicitly reflect these higher-order derivatives in order to get around this restriction.  This transformation improves numerical conditioning and enables more effective optimization by reducing the original PDE into an equivalent system of lower-order equations.  The A-PINN architecture in Figure~\ref{A-PINN} employs a multi-output NN that simultaneously approximates the principal solution field and its auxiliary variables while enforcing physical consistency through the governing equations.  This approach achieves accuracy and computing economy, speeds up convergence, and guarantees a more stable learning process.

In higher-order structural PDEs such as the EBB, directly computing fourth-order derivatives within PINNs often causes numerical instability. By adding auxiliary variables, A-PINN solves this problem and produces learning that is more precise and stable.
\begin{itemize}
    \item It improves the level of accuracy of the predicted dynamic responses (displacement, velocity, curvature, bending moment).  
    \item It captures the complex vibration modes of the continuous structures more reliably than PINNs.  
    \item It ensures physical consistency between the primary and auxiliary fields, which is essential in structural vibration problems.  
    \item It can be extended to more advanced structural models, plates, Timoshenko beams, and shells, where higher-order derivatives appear.  
\end{itemize}

Therefore, while balancing accuracy, computing economy, and physical behavior, A-PINNs offer a reliable and effective framework for simulating the dynamics of continuous structural systems, including EBB vibrations.
\\

\textbf{Formulation of the Loss Function in A-PINN:}

Consider the higher-order PDE in \( \Omega \subset \mathbb{R}^d \):
\begin{equation}
	{u}_{tt}(\mathbf{x}, t) 
	+ \mathcal{N}[{u}(\mathbf{x}, t), \nabla {u}(\mathbf{x}, t), \ldots, \nabla^k {u}(\mathbf{x}, t)] 
	= f(\mathbf{x}, t), 
	\quad \mathbf{x} \in \Omega,\ t \in (0,T],
\end{equation}
\begin{equation}
	\mathcal{B}[{u}(\mathbf{x}, t), \nabla {u}(\mathbf{x}, t), \ldots, \nabla^m {u}(\mathbf{x}, t)] 
	= g(\mathbf{x}), 
	\quad \mathbf{x} \in \partial\Omega,
\end{equation}
\begin{equation}
	\mathcal{I}[{u}(\mathbf{x}, t), \partial_t {u}(\mathbf{x}, t), \ldots, \partial_t^r {u}(\mathbf{x}, t)]\Big|_{t=0} 
	= h(\mathbf{x}), 
	\quad \mathbf{x} \in \Omega.
\end{equation}
A multi-output NN is defined as
\begin{equation}
\mathcal{NN}(\mathbf{x}, t;\theta)
=
\big[
    \widehat{u}(\mathbf{x}, t;\theta),\,
    \nabla \widehat{u}(\mathbf{x}, t;\theta),\,
    \ldots,
    \widehat{v}_{1}(\mathbf{x}, t;\theta),\,
    \widehat{v}_{2}(\mathbf{x}, t;\theta),\,
    \ldots,\,
    \widehat{v}_{l}(\mathbf{x}, t;\theta)
\big],
\end{equation}

where \(\widehat{u}(\mathbf{x}, t;\theta)\) approximates the solution and \( \widehat{v}_{j}(\mathbf{x}, t;\theta) \) approximate derivatives of order \( k_j \).

\medskip
The residual of the PDE is defined as
\begin{equation}
	\mathcal{R}(\mathbf{x}, t) := 
	\widehat{u}_{tt}(\mathbf{x}, t;\theta)
	+ \mathcal{N}[\widehat{u}(\mathbf{x}, t;\theta), 
	\nabla \widehat{u}(\mathbf{x}, t;\theta), 
	\ldots, 
	\widehat{v}_{1}(\mathbf{x}, t;\theta), 
	\ldots, 
	\widehat{v}_{l}(\mathbf{x}, t;\theta)]
	- f(\mathbf{x}, t).
\end{equation}

\noindent
The total loss function comprises different components, namely the BC loss, IC loss, data loss, and auxiliary loss. These are defined as:

\begin{align}
\mathcal{L}_f(\theta) 
&= \frac{1}{N_f} \sum_{i=1}^{N_f} 
\Big| \widehat{u}_{tt}(\mathbf{x}_i^f, t_i^f;\theta) 
+ \mathcal{N}\!\big[\widehat{u}(\mathbf{x}_i^f, t_i^f;\theta), \nabla \widehat{u}(\mathbf{x}_i^f, t_i^f;\theta), \ldots, \widehat{v}_{l}(\mathbf{x}_i^f, t_i^f;\theta)\big] 
- f(\mathbf{x}_i^f, t_i^f) \Big|^2, \\[0.25em]
\mathcal{L}_b(\theta) 
&= \frac{1}{N_b} \sum_{i=1}^{N_b} 
\Big| \mathcal{B}\!\big[\widehat{u}(\mathbf{x}_i^b, t_i^b;\theta), \nabla \widehat{u}(\mathbf{x}_i^b, t_i^b;\theta), \ldots, \widehat{v}_{j}(\mathbf{x}_i^b, t_i^b;\theta)\big] 
- g(\mathbf{x}_i^b) \Big|^2, \\[0.25em]
\mathcal{L}_0(\theta) 
&= \frac{1}{N_0} \sum_{i=1}^{N_0} 
\Big| \mathcal{I}\!\big[\widehat{u}(\mathbf{x}_i^0, 0;\theta), \partial_t \widehat{u}(\mathbf{x}_i^0, 0;\theta), \ldots, \partial_t^r \widehat{u}(\mathbf{x}_i^0, 0;\theta)\big] 
- h(\mathbf{x}_i^0) \Big|^2, \\[0.25em]
\mathcal{L}_d(\theta) 
&= \frac{1}{N_d} \sum_{i=1}^{N_d} 
\Big| \widehat{u}(\mathbf{x}_i^d, t_i^d;\theta) - u(\mathbf{x}_i^d, t_i^d) \Big|^2, \\[0.25em]
\mathcal{L}_{\text{a}}(\theta) 
&= \frac{1}{N_\text{a}} \sum_{i=1}^{N_\text{a}} \sum_{j=1}^l 
\Big| \nabla^{k_j} \widehat{u}(\mathbf{x}_i^\text{a}, t_i^\text{a};\theta) - \widehat{v}_{j}(\mathbf{x}_i^\text{a}, t_i^\text{a};\theta) \Big|^2.
\end{align}
Where, \( \nabla^{k_j} \widehat{u}(\mathbf{x}, t;\theta) \) is obtained via AD. Finally, the overall weighted loss function is expressed as:
\begin{equation}
	\mathcal{L}(\theta) = 
	\tilde{w}_f \,\mathcal{L}_f(\theta) 
	+ \tilde{w}_b \,\mathcal{L}_b(\theta) 
	+ \tilde{w}_0 \,\mathcal{L}_0(\theta) 
	+ \tilde{w}_d \,\mathcal{L}_d(\theta) 
	+ \tilde{w}_{\text{a}} \,\mathcal{L}_{\text{a}}(\theta),
    \label{A_Loss}
\end{equation}
where, \( \tilde{w}_f, \tilde{w}_b, \tilde{w}_0, \tilde{w}_d, \tilde{w}_{\text{a}} \) 
are adaptive positive weights balancing the contributions of each loss term.

The A-PINN architecture, depicted in Figure \ref{A-PINN}, processes inputs \( (\mathbf{x}, t) \) to produce primary and auxiliary outputs, which are used to compute the composite loss.

\begin{algorithm}
	\caption{A-PINN Algorithm for the EBB model}
    \label{algorithm}
	\begin{algorithmic}[1]
		\State \textbf{Inputs:} Initial weights $w_0$, biases $b_0$
		\State \textbf{Given:} ICs 
\[
u(x,0) = \psi(x), \quad u_t(x,0) = \psi_t(x),
\]
and BCs 
\[
u(a,t)=0, \quad u_{xx}(a,t)=0, \quad u(b,t)=0, \quad u_{xx}(b,t)=0
\]

		\State \textbf{Setup:} Domain $\Omega = [a,b]\times[c,d]$, tolerance $\epsilon$, maximum iterations $max\_iter$
		\State \textbf{Initialize:} Main network parameters $\theta_u \gets (w_0, b_0)$, auxiliary parameters $\theta_\text{a} \gets (w_0, b_0)$, iteration counter $k \gets 0$
	\State Generate the collocation points $(x_j^f,t_j^f)$, initial points $(x_j^0,0)$, boundary points $(a,t_j^b),(b,t_j^b)$, data points $(x_j^d,t_j^d)$, and auxiliary points $(x_j^\text{a},t_j^\text{a})$

		\While{not converged}
		    \State Predict $u(x,t;\theta)$ and auxiliary outputs $v(x,t;\theta)$
		    \State Compute losses: $\mathcal{L}_f$, $\mathcal{L}_b$, $\mathcal{L}_0$, $\mathcal{L}_d$, $\mathcal{L}_\text{a}$
		    
		    \State Compute total loss:  $\mathcal{L}_{total}$
		       \State \textbf{Optimization:} Adaptive Adam for the global optimization, followed by L\texttt{--}BFGS for local refinement.

		    \State Update $\theta_u,\theta_\text{a}$ 
		    \If{$|\Delta \mathcal{L}_{total}| < \epsilon$}
		        \State \textbf{break}
		    \EndIf
		    \State $k \gets k+1$
		\EndWhile
		\State \Return $(\theta_u, \theta_\text{a})$
	\end{algorithmic}
\end{algorithm}


\section{Numerical Experiments}\justifying\label{sec5}

The subsequent sections detail the experimental setup and results for three case studies of the EBB model, governed by the fourth-order PDE. Moreover, the three cases are undamped free vibration without external force (P1), undamped force vibration with an applied load (P2), and undamped force vibration on a Winkler foundation (P3) that introduces an additional elastic restoring effect ~\cite{deng2023dynamic}.
 A comparative assessment is carried out against the GT, along with the baselines, viz. FDM, SANN, and PINN. The numerical results demonstrate the framework's reliability and accuracy, emphasizing the potential of data-driven methods in structural analysis. The experiments simulate the primary case and use the extracted parameters to evaluate the modified A-PINN performance. Various numerical simulations are carried out on benchmark problems involving structural vibrations based on the EBB models. These simulations are used to compare the accuracy of A-PINNs with PINNs. Both models employ fully connected FFNNs that have hyperparameters listed in Table \ref{hyperparameters}. During the training, the Adam and L-BFGS combination is employed as an adaptive optimizer to update the NN parameters.

\begin{table}[H]
\centering
\caption{Hyperparameters used for training the A-PINN model}
\begin{minipage}{0.6\textwidth}
\centering
\begin{tabularx}{\textwidth}{l>{\raggedright\arraybackslash}X}
\toprule
\textbf{Hyperparameter} & \textbf{Value} \\
\midrule
Number of hidden layers & 4 \\
Neurons per layer       & 55 \\
Activation function     & Tanh \\
Epochs                  & 20{,}000 (for P1), 10{,}000 (for P2 and P3) \\
Learning rate           & 0.1  \\
Batch size              & 500 \\
\bottomrule
\end{tabularx}
\end{minipage}
\label{hyperparameters}
\end{table}

The hardware configuration setup used during the training is presented in Table~\ref{CPU}. The A-PINN training procedure is elaborated in Algorithm \ref{algorithm}
\begin{table}[H]
\centering
\caption{Hardware and software setup: CPU specifications for model training}
\begin{minipage}{0.6\textwidth}
\centering
\begin{tabularx}{\textwidth}{l>{\raggedright\arraybackslash}X}
\toprule
\textbf{Component} & \textbf{Specification} \\
\midrule
Processer      & AMD Ryzen 7 5700U (1.80 GHz) \\
Cores / Threads  & 8 cores / 16 threads \\
Memory Size      & 8 GB RAM \\
Operating System & Windows 11 \\
Python Version   & 3.11.2 \\
\bottomrule
\end{tabularx}
\end{minipage}
\label{CPU}
\end{table}

To quantitatively model and analyze the accuracy of the predicted solutions, several error metrics are employed. These metrics provide complementary insights into the model performance by capturing both local and global discrepancies between the predicted solution \(u_{\text{pred}}\) and the GT \(u_{\text{GT}}\). 
The pointwise AE highlights the local deviations at each discretization point of the given domain, and the relative error quantifies the overall discrepancy in an averaged sense, and the \(L^\infty\) error measures across the given domain. 
The definitions of these metrics are given below:
\begin{itemize}
	\item \textbf{Absolute Error $(E_1)$:}  
	\begin{equation}
		E_1 
		= \left| u_{\text{pred}}(x_i) - u_{\text{GT}}(x_i) \right|, 
		\quad i = 1,
        2, \dots, N,
		\label{eq:absolute_error}
	\end{equation}
	where \(u_{\text{pred}}\) is the predicted PDE solution by the NN, \(u_{\text{GT}}\) is the GT solution, and \(N\) is the number of discretization points in the spatial or spatio-temporal domain.

    	\item \textbf{Mean Squared Error $(E_2)$:}
	\begin{equation}
		E_2 = 
		\frac{1}{N} \sum_{i=1}^{N} 
		\left( u_{\text{pred}}(x_i) - u_{\text{GT}}(x_i) \right)^2,
		\label{eq:mse_error}
	\end{equation}
	where \(E_2\) quantifies the average of square differences between the predicted and true solutions, thus measuring the overall fitting accuracy of the model.
	\item \textbf{Relative Error \(L^2\)  $(E_3)$:}  
	\begin{equation}
		E_3 = 
		\frac{\| u_{\text{pred}} - u_{\text{GT}} \|_2}{\| u_{\text{GT}} \|_2} 
		= \frac{\sqrt{\sum_{i=1}^{N} \left( u_{\text{pred}}(x_i) - u_{\text{GT}}(x_i) \right)^2}}
		{\sqrt{\sum_{i=1}^{N} \left( u_{\text{GT}}(x_i) \right)^2}}, 
		\quad i = 1, 2, \dots, N.
		\label{eq:l2_error}
	\end{equation}

	\item \textbf{\(L^\infty\) Error $(E_4)$:}  
	\begin{equation}
		E_4 = 
		\max_{1 \leq i \leq N} \left| u_{\text{pred}}(x_i) - u_{\text{GT}}(x_i) \right|,
		\label{eq:linf_error}
	\end{equation}
	which represents the maximum absolute error between the predicted and GT solutions over the discretization domain.
\end{itemize}

\subsection{Undamped Free Vibration of the Euler-Bernoulli Beam}

The EBB model is a classical framework for analyzing the transverse vibration of beams. In the absence of external excitation, governed by
\begin{equation} \label{P_1}
u_{tt}(x,t) + u_{xxxx}(x,t) = 0, \quad x \in [0,1], \; t \in [0,1].
\end{equation}
The system is subjected to simply supported BCs, namely 
$u(0,t) = u(1,t) = 0$ and $u_{xx}(0,t) = u_{xx}(1,t) = 0$, which ensure zero displacement and bending moment at both ends of the beam. The ICs are prescribed as $u(x,0) = \sin(\pi x)$ and $u_t(x,0) = 0$, corresponding to an initial deflection profile with zero initial velocity.

In the A-PINN formulation, an auxiliary variable $v(x,t)$ is introduced to approximate the second spatial derivative $u_{xx}(x,t)$, allowing the fourth-order EBB equation to be written as a system of lower-order relations. Then the total loss combines the PDE residual, the auxiliary constraint, and the ICs and BCs. 
It is defined as

\begin{align} \label{eq35}
\mathcal{L}(\theta) 
&= \tilde{w}_f \left( \frac{1}{N_f} \sum_{i=1}^{N_f} 
\left| \widehat{u}_{tt}(x_i^f,t_i^f;\theta) + \widehat{v}_{xx}(x_i^f,t_i^f;\theta) \right|^2 \right) \notag \\
&\quad + \tilde{w}_\text{a} \left( \frac{1}{N_\text{a}} \sum_{i=1}^{N_\text{a}} 
\left| \widehat{v}(x_i^\text{a},t_i^\text{a};\theta) - \widehat{u}_{xx}(x_i^\text{a},t_i^\text{a};\theta) \right|^2 \right) \notag \\
&\quad + \tilde{w}_0 \left( \frac{1}{N_0} \sum_{i=1}^{N_0} 
\Big( \left|\widehat{u}(x_i^0,0;\theta) - \sin(\pi x_i^0)\right|^2 
     + \left|\widehat{u}_t(x_i^0,0;\theta)\right|^2 \Big) \right) \notag \\
&\quad + \tilde{w}_b \left( \frac{1}{N_b} \sum_{i=1}^{N_b} 
\Big( \left|\widehat{u}(0,t_i^b;\theta)\right|^2 + \left|\widehat{u}(1,t_i^b;\theta)\right|^2 
+ \left|\widehat{v}(0,t_i^b;\theta)\right|^2 + \left|\widehat{v}(1,t_i^b;\theta)\right|^2 \Big) \right).
\end{align}

\begin{figure}[H]
  \centering
  \captionsetup[sub]{justification=centering}
  {\large Physics-informed solutions}\par\vspace{0.4em}

  \begin{subfigure}{0.48\textwidth}
    \centering
    \includegraphics[width=\linewidth]{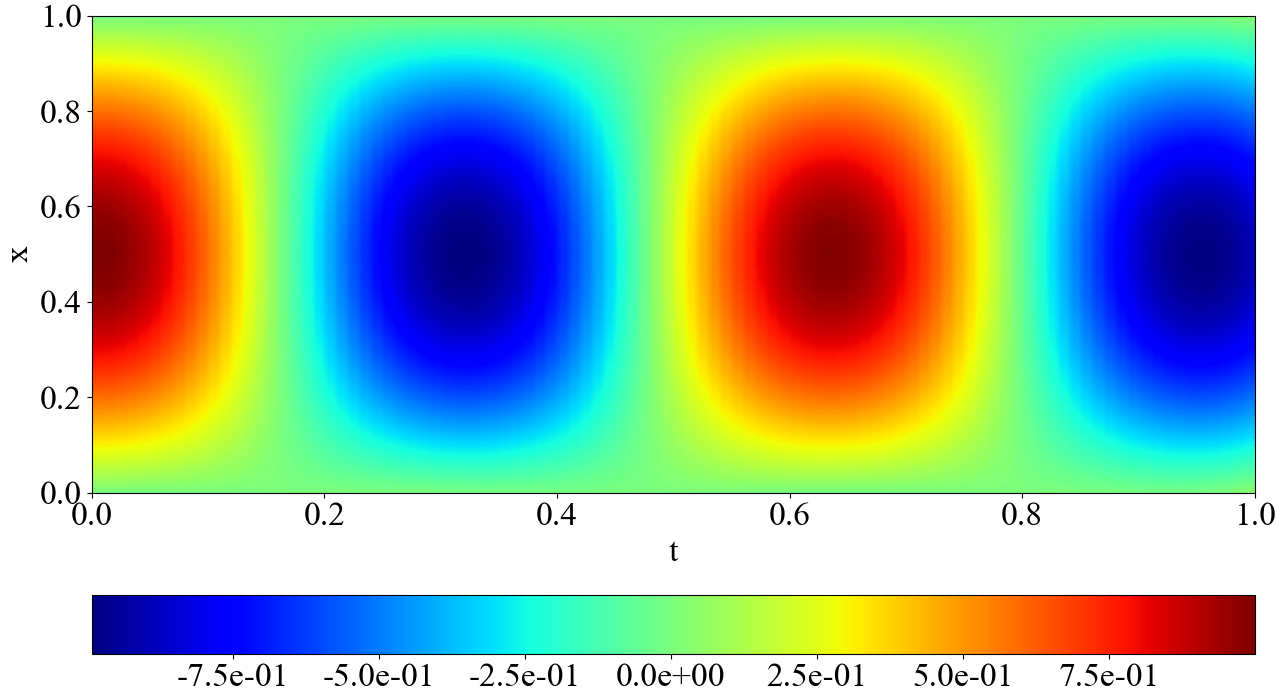}
    \caption{PINN}
  \end{subfigure}
  \hfill
  \begin{subfigure}{0.48\textwidth}
    \centering
    \includegraphics[width=\linewidth]{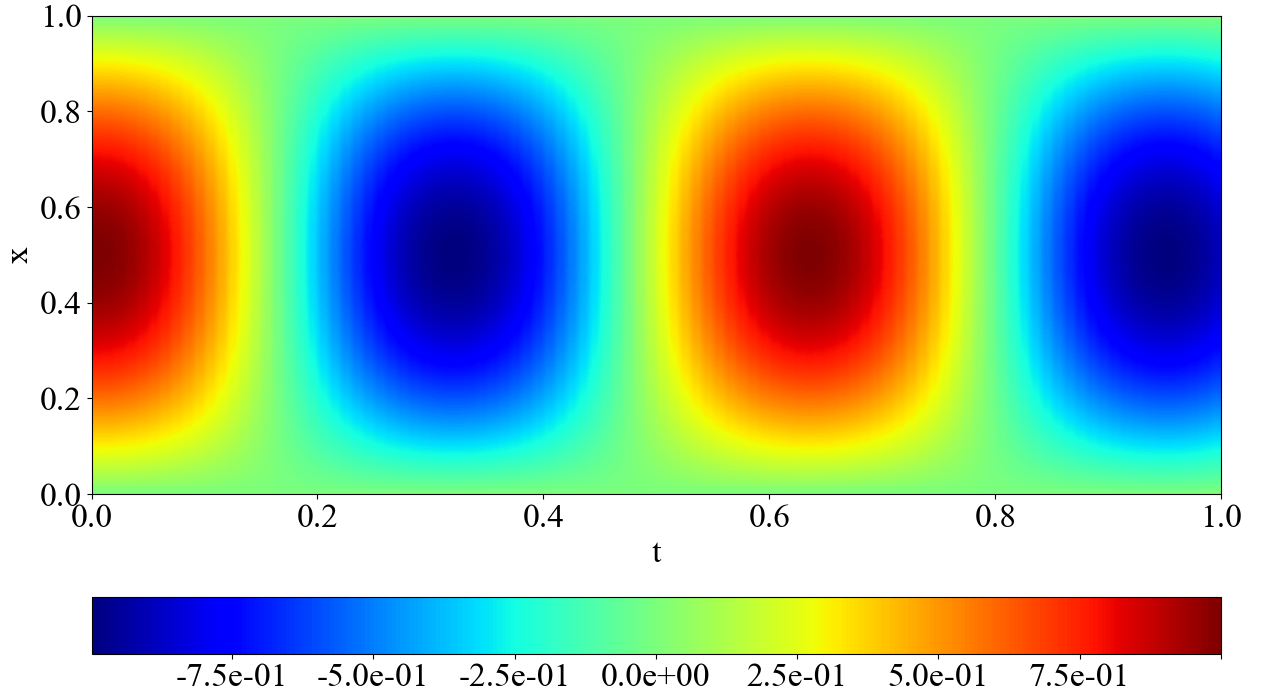}
    \caption{A-PINN (Proposed model)}
  \end{subfigure}

  \vspace{1.0em} 

{\large Baseline solutions}\par\vspace{0.4em}

  \begin{subfigure}{0.48\textwidth}
    \centering
    \includegraphics[width=\linewidth]{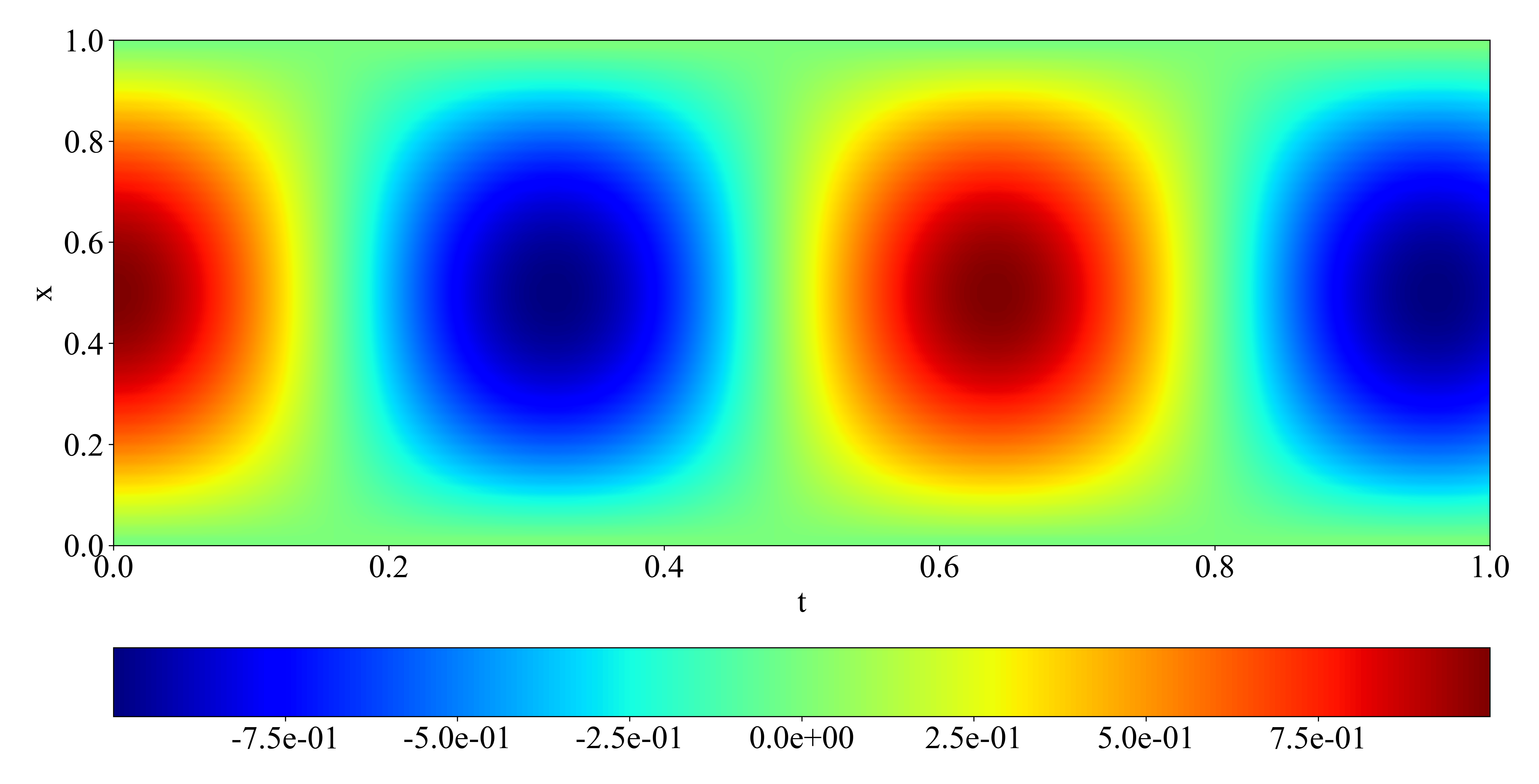}
    \caption{FDM}
  \end{subfigure}
  \hfill
  \begin{subfigure}{0.48\textwidth}
    \centering
    \includegraphics[width=\linewidth]{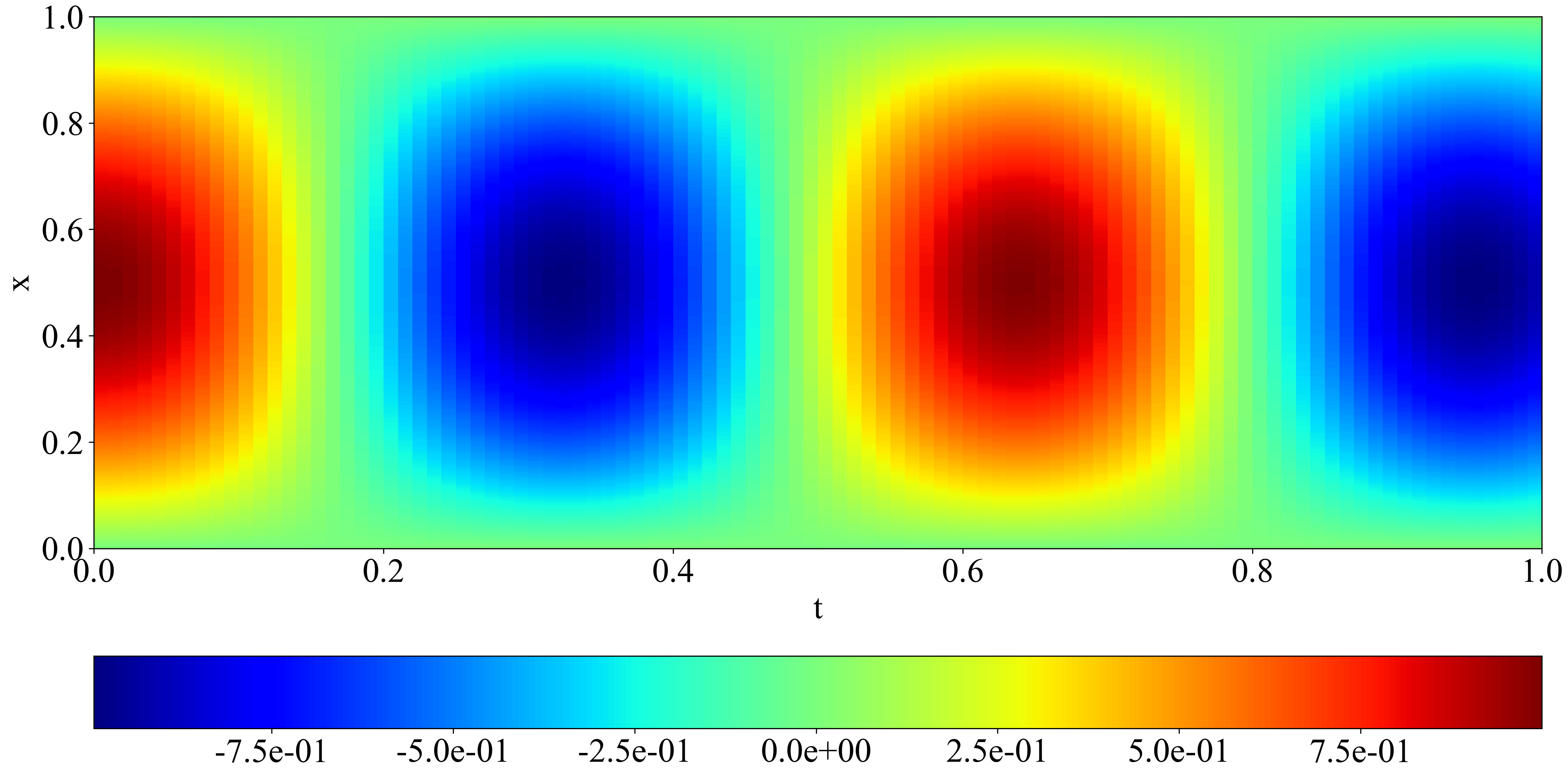}
    \caption{SANN}
  \end{subfigure}

 \vskip\baselineskip 
  \begin{subfigure}{0.48\textwidth}
    \centering
    \includegraphics[width=\linewidth]{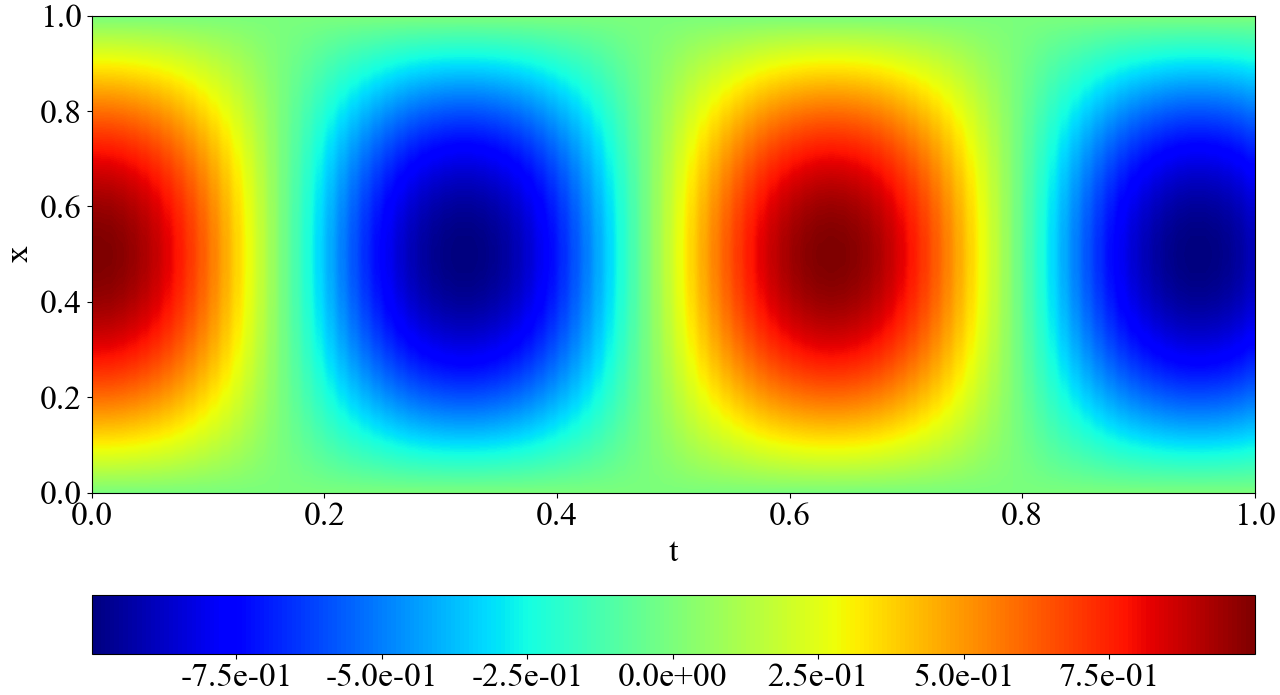}
    \caption{GT}
  \end{subfigure}

  \caption{2D comparison of physics-informed (top) and baseline (bottom) solutions for P1.}
  \label{P_1_comparison_2d_blocks}
\end{figure}

\begin{figure}[H]
  \centering
  \captionsetup[sub]{justification=centering}

  {\large Physics-informed solutions}\par\vspace{0.4em}

  \begin{subfigure}{0.48\textwidth}
    \centering
    \includegraphics[width=\linewidth]{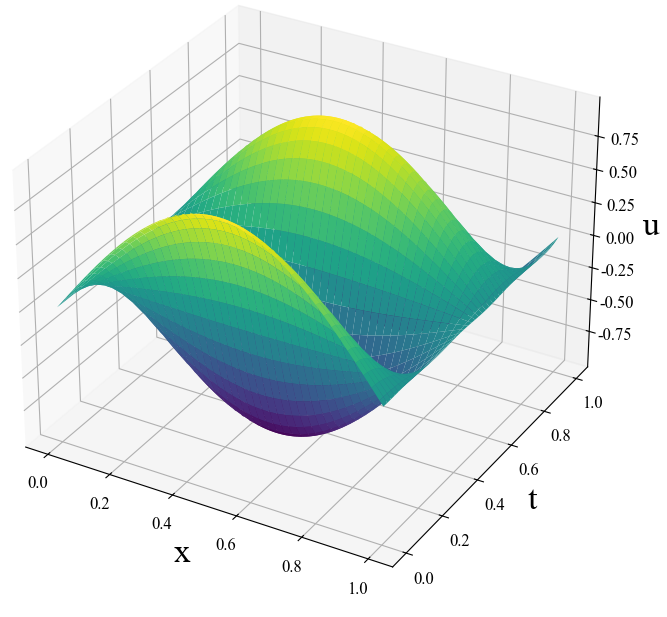}
    \caption{PINN}
  \end{subfigure}
  \hfill
  \begin{subfigure}{0.48\textwidth}
    \centering
    \includegraphics[width=\linewidth]{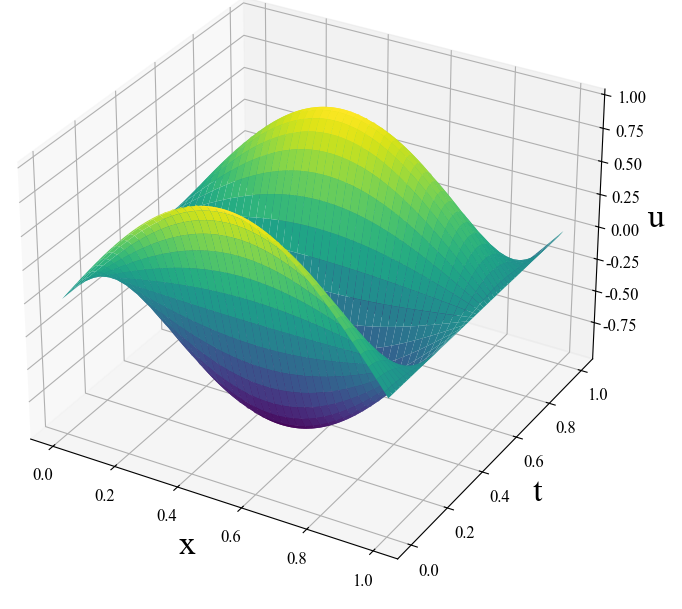}
    \caption{A-PINN (Proposed model)}
  \end{subfigure}

  \vspace{1.0em} 

  {\large Baseline solutions}\par\vspace{0.4em}

  \begin{subfigure}{0.45\textwidth}
    \centering
    \includegraphics[width=\linewidth]{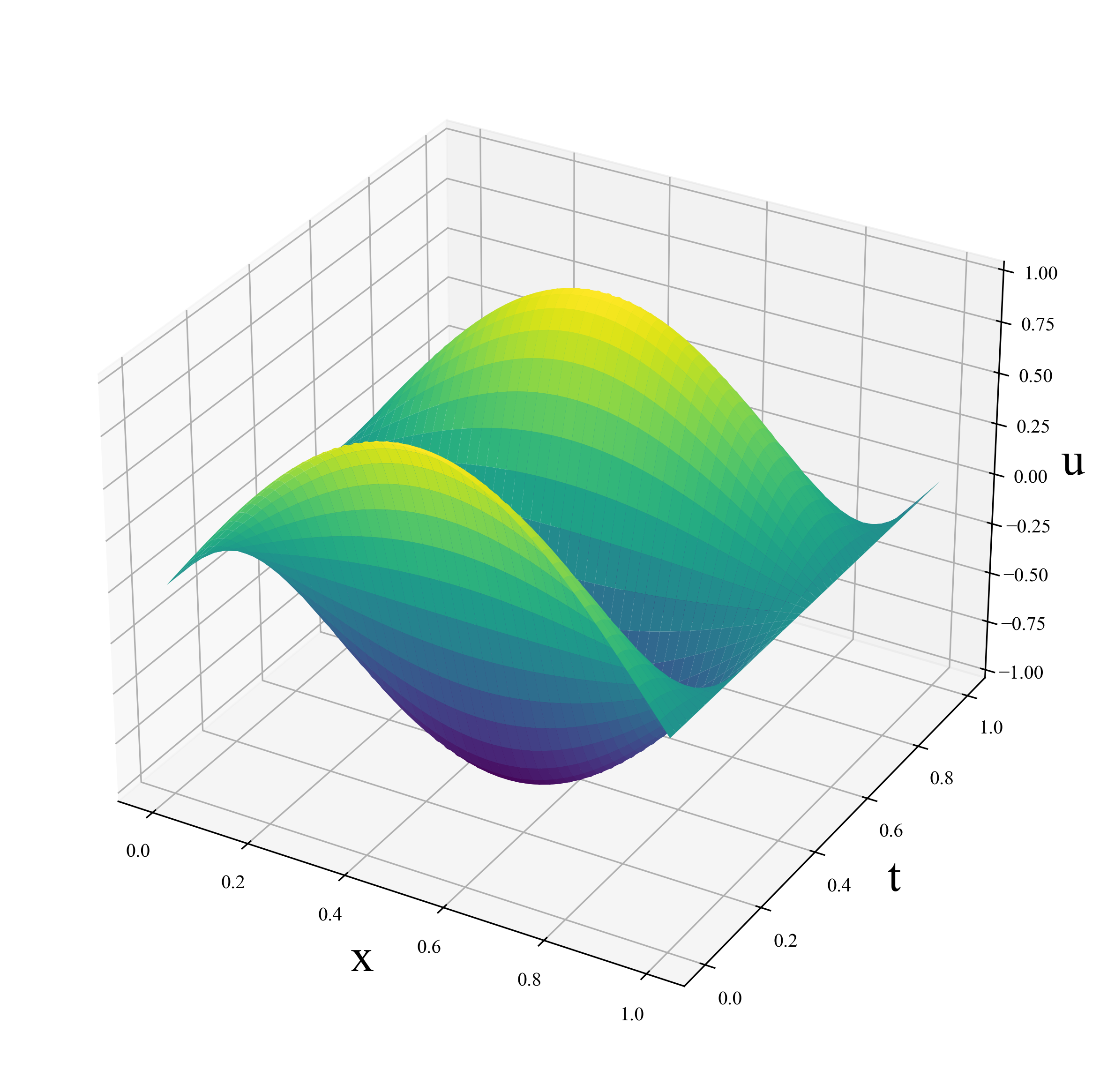}
    \caption{FDM}
  \end{subfigure}
  \hfill
  \begin{subfigure}{0.45\textwidth}
    \centering
    \includegraphics[width=\linewidth]{ 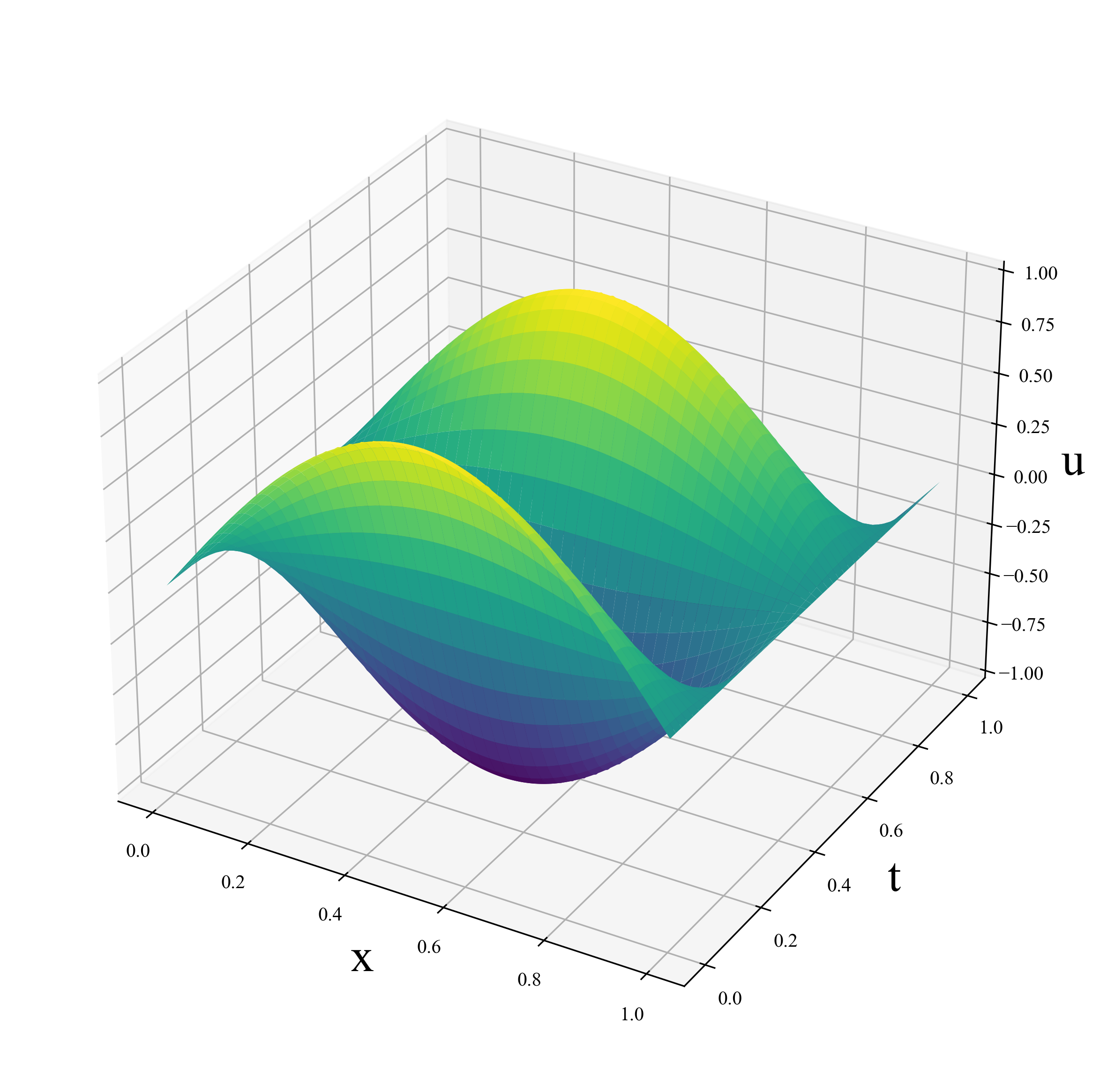}
    \caption{SANN}
  \end{subfigure}
    \vskip\baselineskip 
  \begin{subfigure}{0.45\textwidth} 
    \centering
    \includegraphics[width=\linewidth]{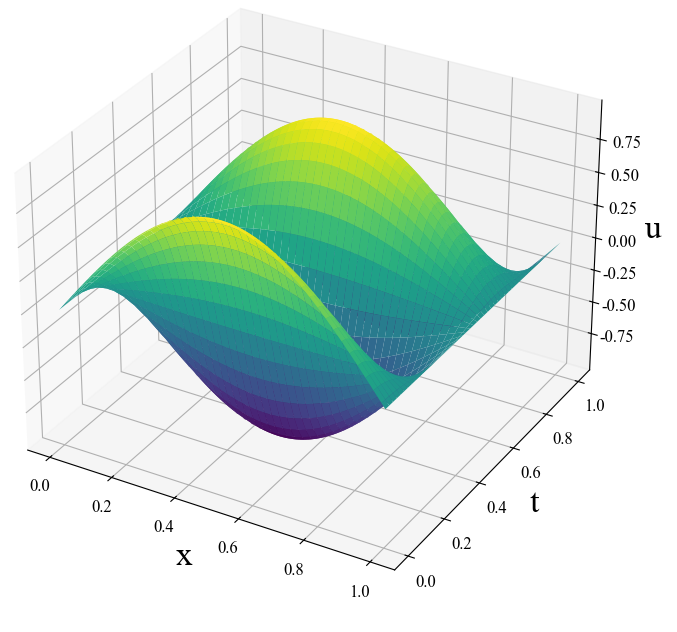}
    \caption{GT}
  \end{subfigure}

  \caption{3D comparison of physics-informed (top) and baseline (bottom) solutions for P1.}
  \label{P1_comparison_3D}
\end{figure}

Here, in Eq. \ref{eq35}, \( \widehat{u}(x,t;\theta) \) and \( \widehat{v}(x,t;\theta) \) denote the NN outputs corresponding to the displacement field and its auxiliary variable, respectively. The NN employed is a fully connected DNN, with its detailed hyperparameters summarized in Table \ref{hyperparameters}. The training dataset is constructed as follows: $N_0 = 200$ points from the initial line $t=0$, $N_b = 200$ points along each boundary $x=0$ and $x=1$, $N_f = 500$ collocation points uniformly distributed inside the spatio-temporal domain for minimizing the governing PDE residual, and $N_\text{a} = 500$ auxiliary points sampled within the domain to enforce the relation $v \approx u_{xx}$. These subsets collectively ensure that the ICs, BCs, PDE residuals, and auxiliary constraints are simultaneously satisfied during optimization. To evaluate the performance of our framework, the EBB model without forcing is considered under simply supported BCs. The exact solution \cite{alli2003solutions} is regarded as the GT and is used as the benchmark for comparison with physics-based PINN and A-PINN models, the baseline numerical technique FDM, and a well-known neural model SANN \cite{chakraverty2025artificial}. Figures~\ref{P_1_comparison_2d_blocks} and~\ref{P1_comparison_3D} illustrate the 2D and 3D displacement fields and the comparative performance of A–PINN against PINN, FDM, and SANN. It is observed that the A-PINN provides closer agreement with the GT, capturing smoother spatio-temporal variations, while the PINN exhibits discrepancies near the boundaries.

\begin{figure}[H]
    \centering
    
    \begin{subfigure}{0.45\textwidth}
        \centering
        \includegraphics[width=\linewidth]{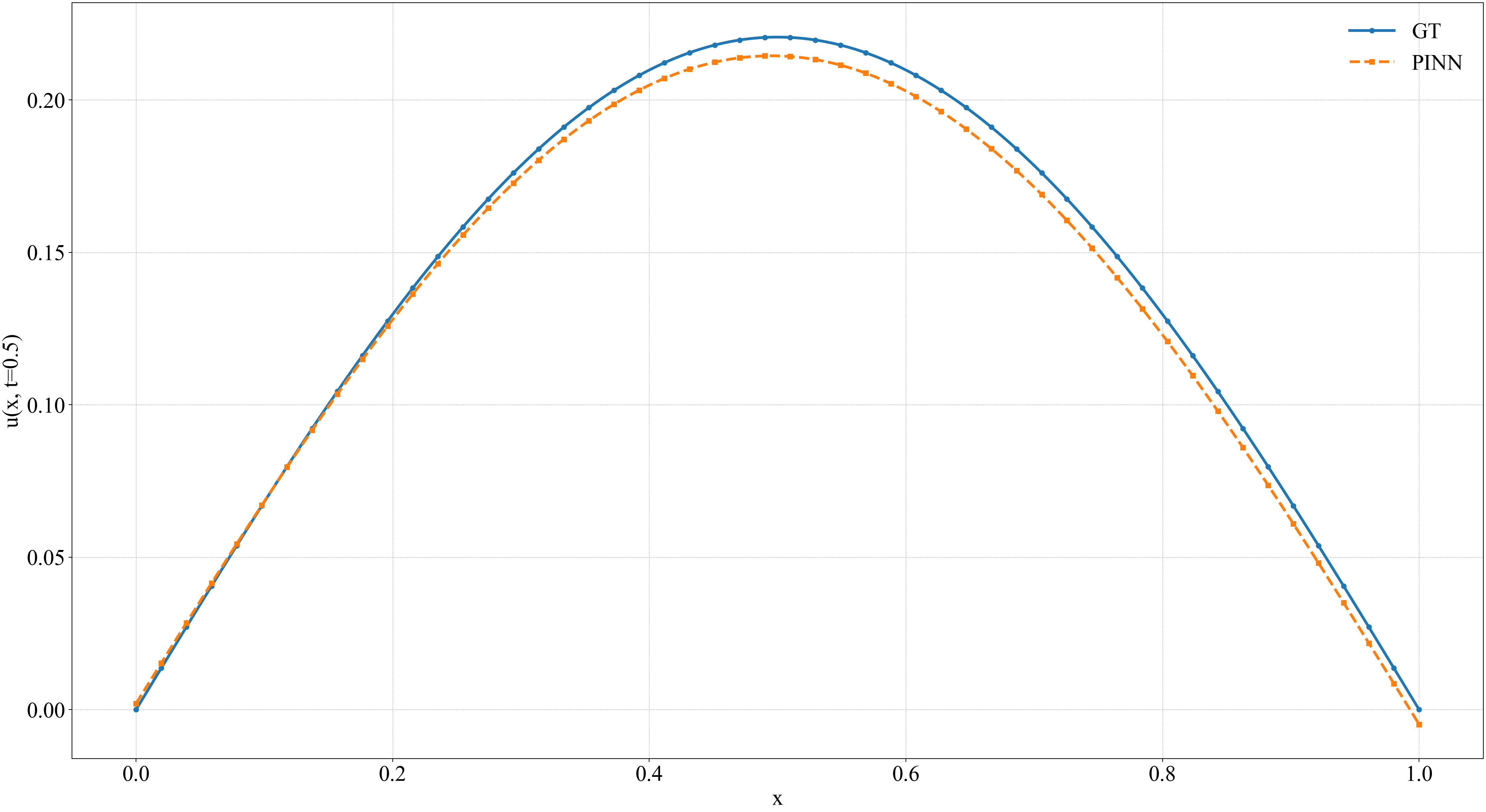}
        \caption{t = 0.5 using PINN}
    \end{subfigure}
    \hfill
    \begin{subfigure}{0.45\textwidth}
        \centering
        \includegraphics[width=\linewidth]{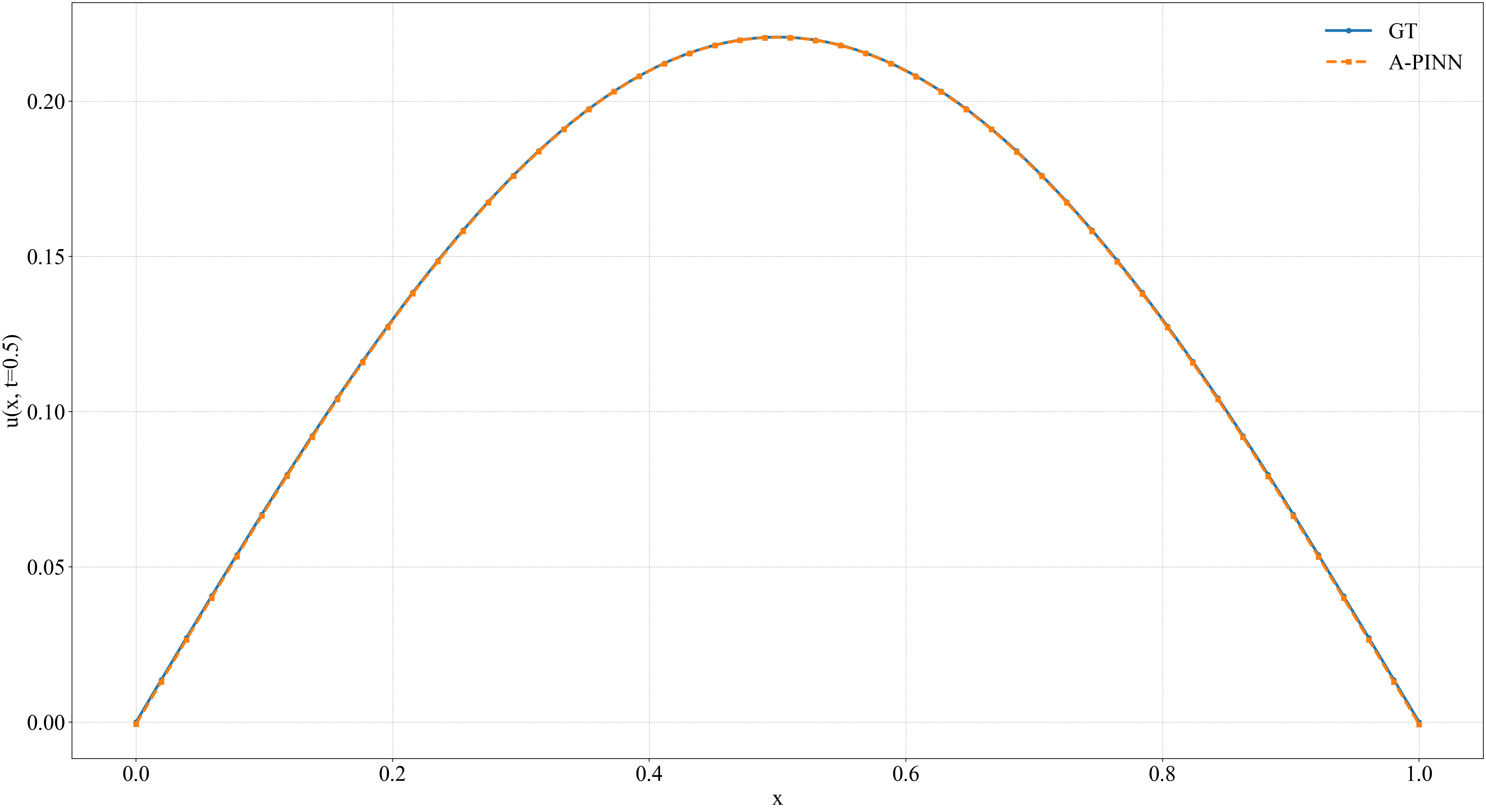}
        \caption{t = 0.5 using A-PINN}
    \end{subfigure}
    
    \vskip\baselineskip
    \begin{subfigure}{0.45\textwidth}
        \centering
        \includegraphics[width=\linewidth]{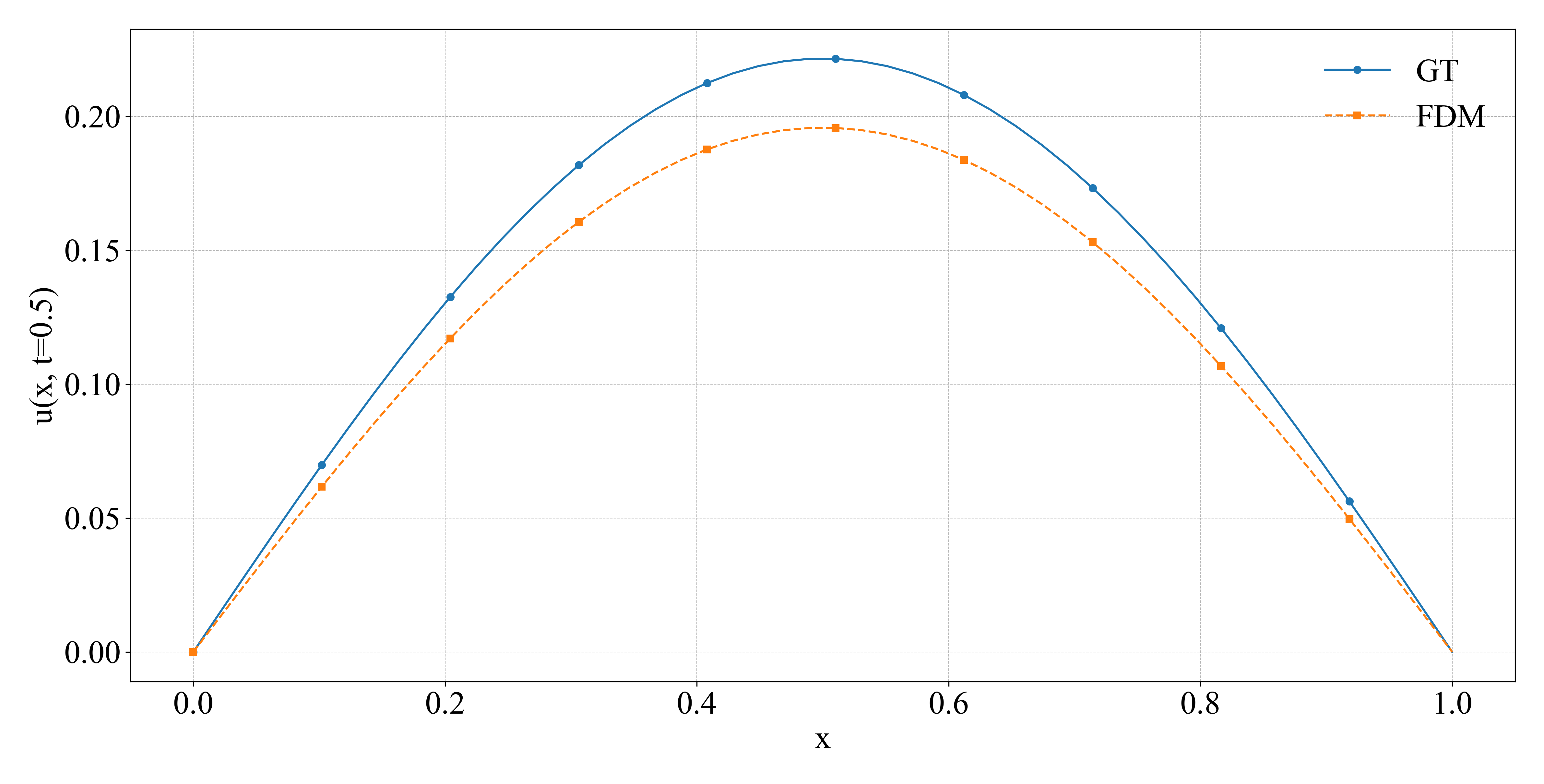}
        \caption{t = 0.5 using FDM}
    \end{subfigure}
    \hfill
    \begin{subfigure}{0.45\textwidth}
        \centering
        \includegraphics[width=\linewidth]{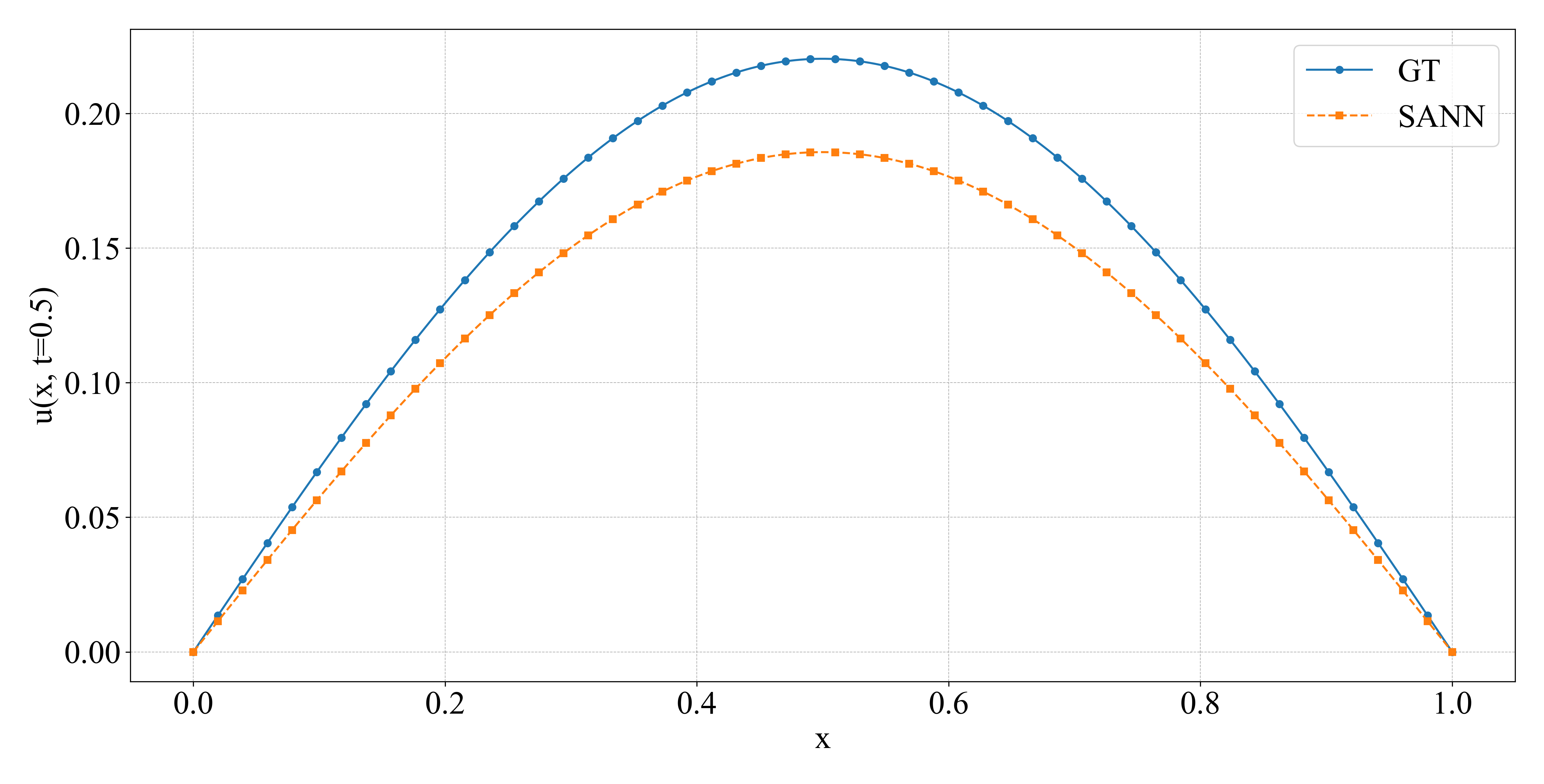}
        \caption{t = 0.5 using SANN}
    \end{subfigure}
    
    \caption{Comparison of PINN, A-PINN, FDM, and SANN with GT at $t=0.5$ for P1.}
    \label{2D_slice_t=0.5(P1)}
\end{figure}

\begin{figure}[H]
    \centering
    \begin{subfigure}{0.45\textwidth}
        \centering
        \includegraphics[width=\linewidth]{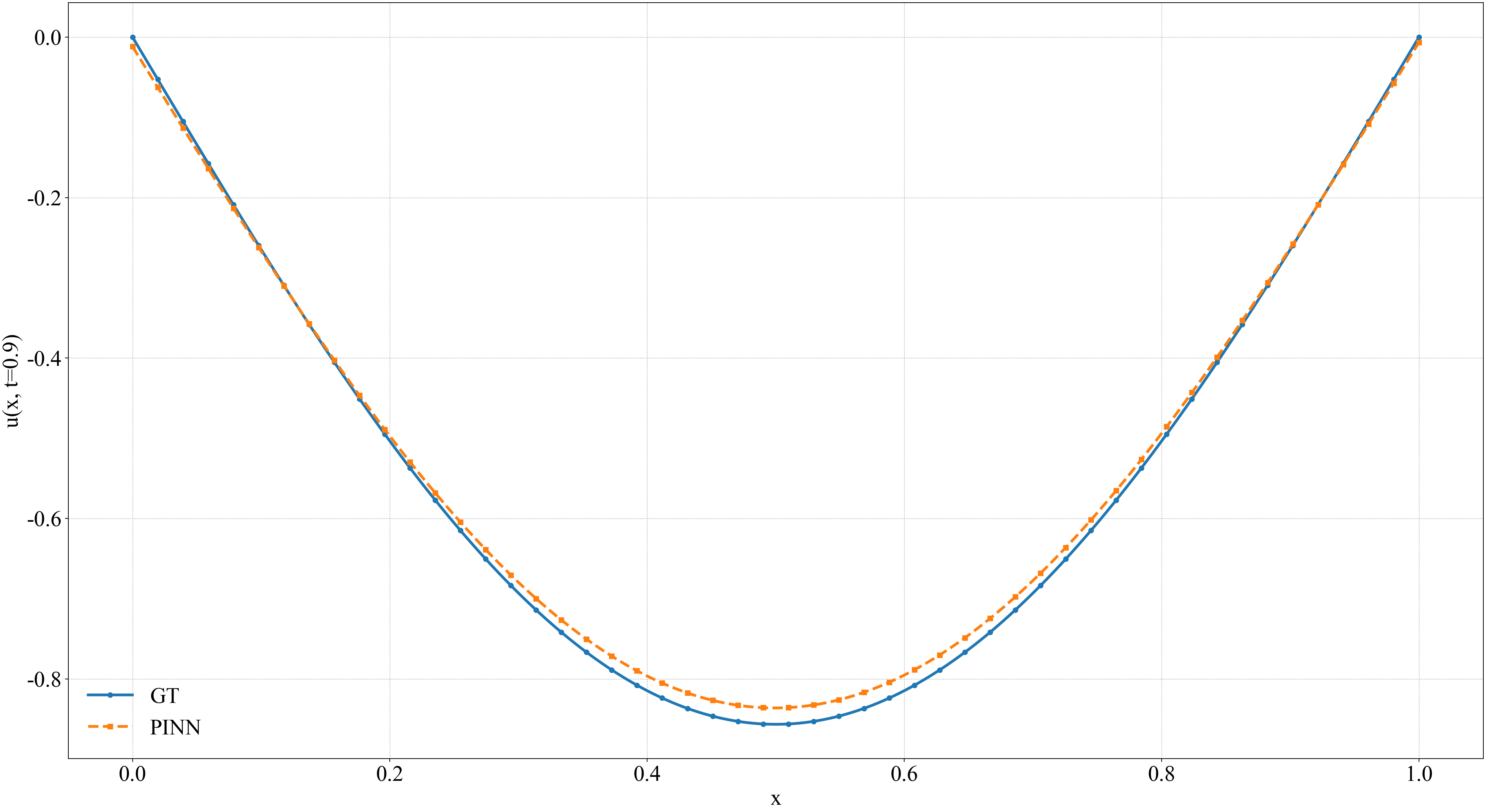}
        \caption{t=0.9 using PINN}
    \end{subfigure}
    \hfill
    \begin{subfigure}{0.45\textwidth}
        \centering
        \includegraphics[width=\linewidth]{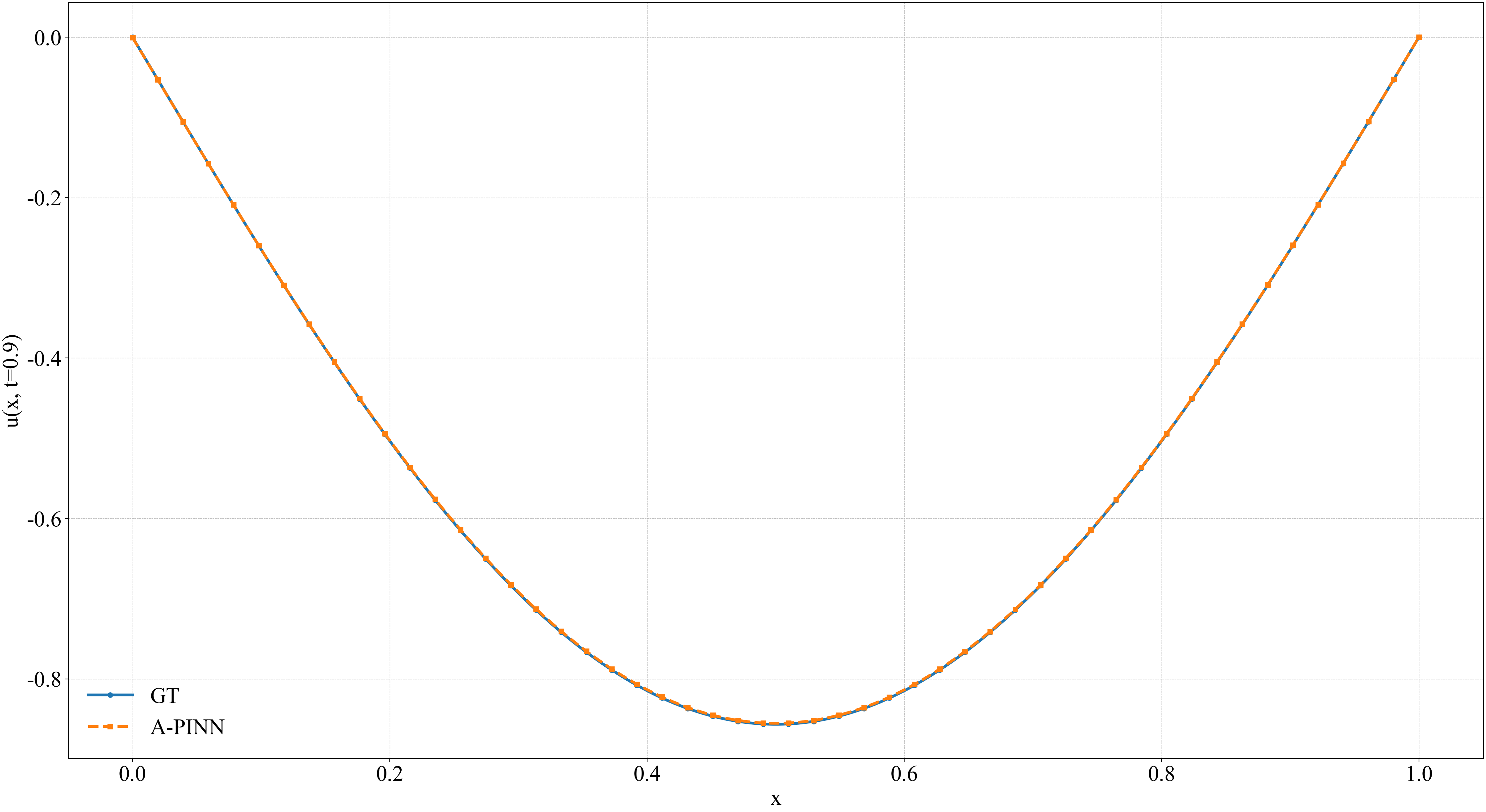}
        \caption{t=0.9 using A-PINN}
    \end{subfigure}
    
    \begin{subfigure}{0.45\textwidth}
        \centering
        \includegraphics[width=\linewidth]{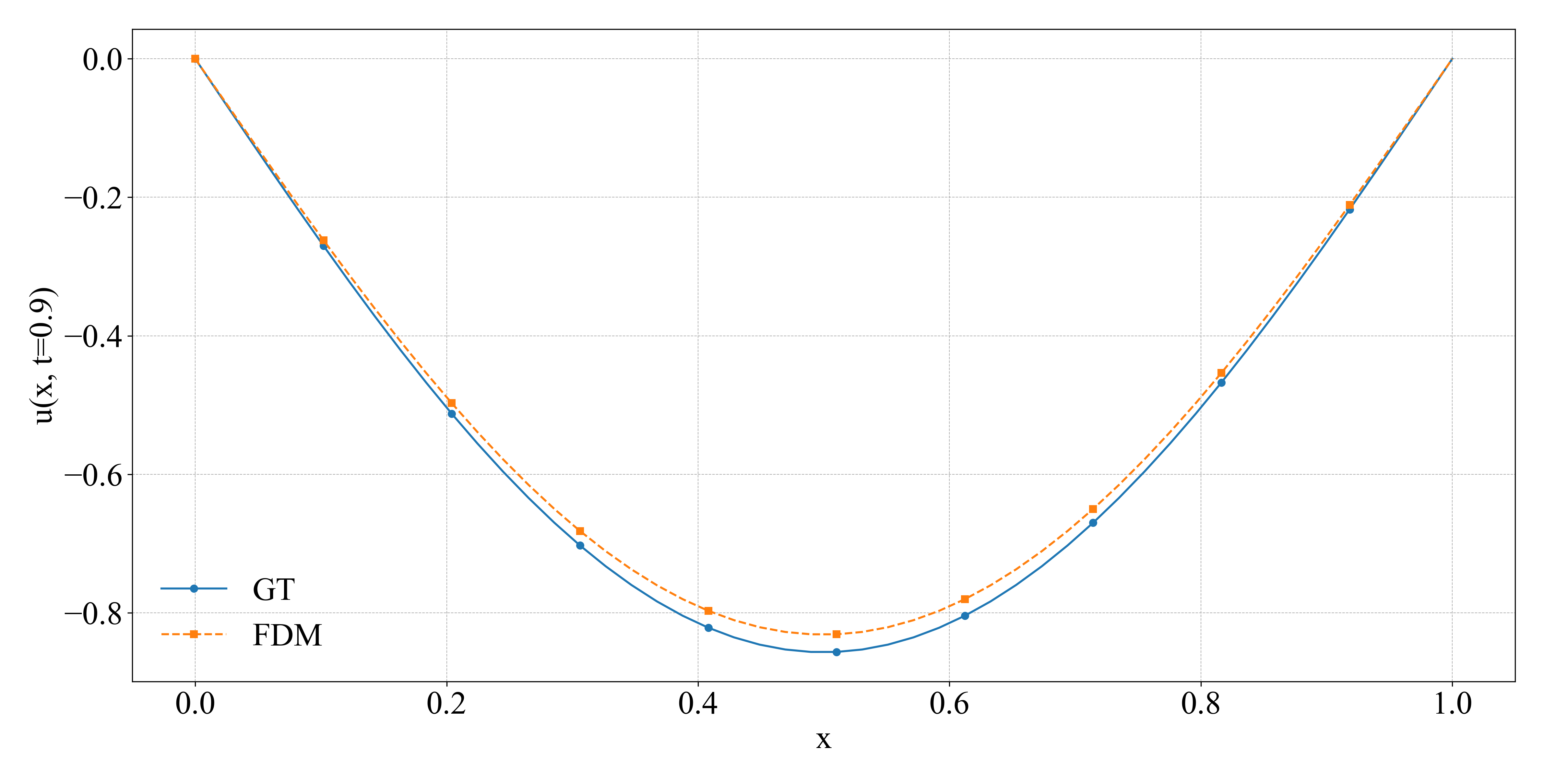}
        \caption{t=0.9 using FDM}
    \end{subfigure}
    \hfill
    \begin{subfigure}{0.45\textwidth}
        \centering
        \includegraphics[width=\linewidth]{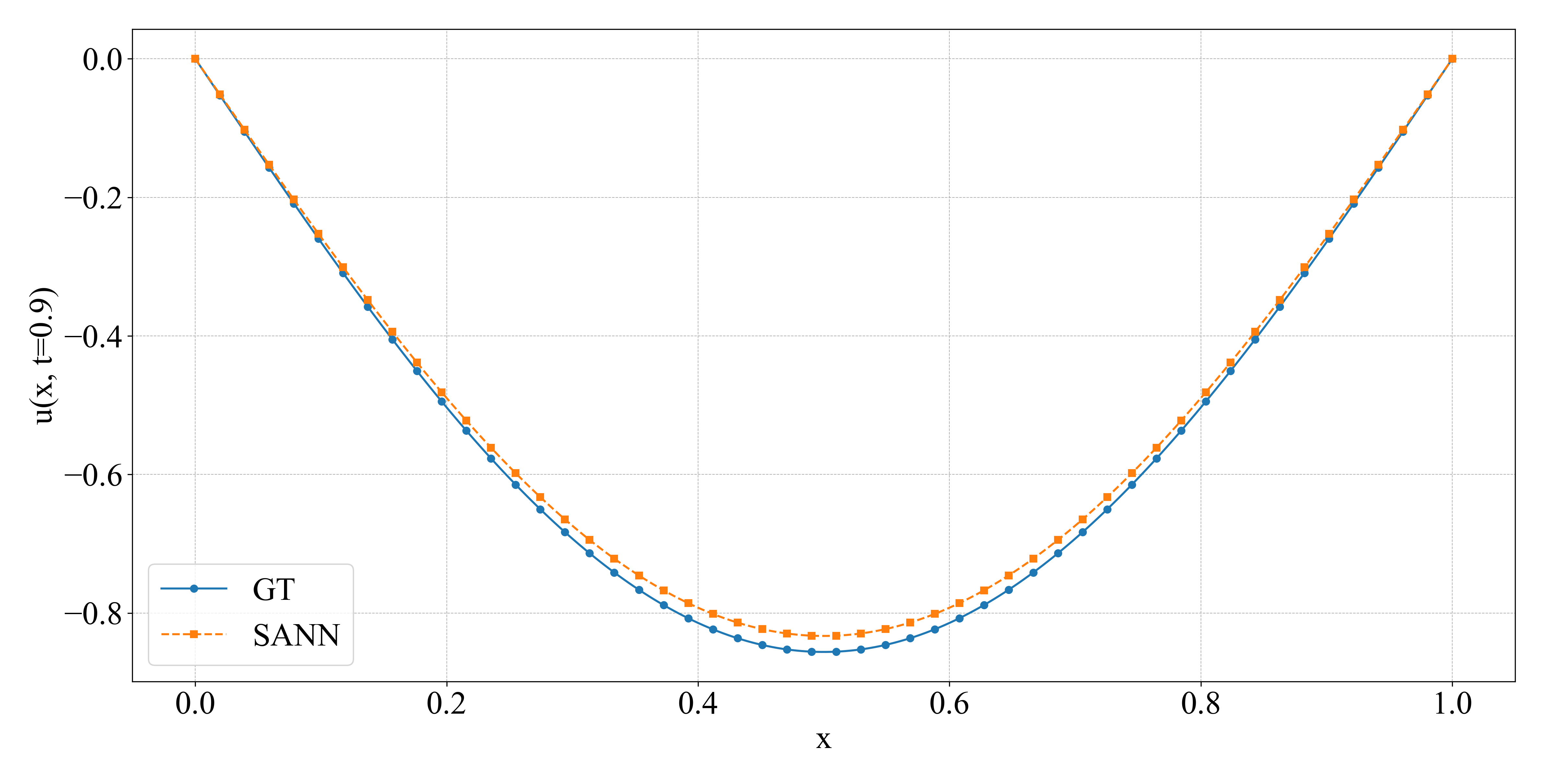}
        \caption{t=0.9 using SANN}
    \end{subfigure}
    \hfill

    \caption{Comparison of PINN, A-PINN, FDM, and SANN with GT at $t=0.9$ for P1.}
    \label{2D_Slice_t=0.9(P1)}
\end{figure}

The accuracy of the our framework is further assessed by analyzing the solutions at two representative time instances, $t=0.5$ (cf.: Figure \ref{2D_slice_t=0.5(P1)} and Table \ref{Table_t=0.5_(P1)}) and $t=0.9$ (cf.: Figure \ref{2D_Slice_t=0.9(P1)} and Table \ref{Table_t=0.9_(P1)}). At $t=0.5$, the displacement has reduced amplitude, corresponding to an intermediate oscillation, and the A-PINN prediction remains very close to the GT, whereas the PINN slightly underestimates the peak. At $t=0.9$, the profile is inverted compared to the initial condition, reflecting the opposite vibration phase. This phase reversal is captured properly by the A-PINN, while the PINN and other baselines present noticeable deviation.

The fundamental physics of the beam vibration may be connected to the variation in curve profiles at various time instances. The spatial shape $\sin(\pi x)$ remains unchanged, while the temporal factor $\cos(\pi^2 t)$ modulates the amplitude and oscillation. When $\cos(\pi^2 t)=\pm1$, the beam reaches its maximum displacement, which corresponds to maximum bending strain energy and zero kinetic energy. Conversely, when $\cos(\pi^2 t)\approx0$, the displacement is nearly zero, the velocity is maximum, which indicates dominance of kinetic energy. Thus, the differences in the displacement curves across time reflect the natural periodic exchange of energy between strain and kinetic components. Overall, our A-PINN achieves superior accuracy compared to the PINN, SANN, and FDM, demonstrating its effectiveness for high-order structural vibration problems.

\begin{table}[ht]
    \centering
    \caption{Comparison among proposed model and baseline models at $t = 0.5$, with spatial domain $x \in [0,1]$,  spatial step $\delta x= 0.1$ for P1}
    \label{Table_t=0.5_(P1)}
    \scriptsize
    \begin{tabular}{c ccc c cc c cc}
        \toprule
        $\textbf{x}$ & \multicolumn{3}{c}{Baselines Results} & \multicolumn{2}{c}{Physics-informed Results} & \multicolumn{4}{c}{$E_1$ w.r.t.\ GT} \\
        \cmidrule(lr){2-4} \cmidrule(lr){5-6} \cmidrule(lr){7-10}
            & GT & FDM & SANN & PINN & A-PINN & \multicolumn{2}{c}{PINN} & \multicolumn{2}{c}{A-PINN} \\
        \midrule
        0.00 & 0.000000 & 0.000000 & 0.000000 &  0.003881 &  -0.000709 & \multicolumn{2}{c}{3.880e-03} & \multicolumn{2}{c}{7.093e-04} \\
        0.10 & 0.068164 & 0.060478 & 0.057436 & 0.068394 & 0.066947 & \multicolumn{2}{c}{2.302e-04} & \multicolumn{2}{c}{1.217e-03} \\
        0.20 & 0.129656 & 0.115020 & 0.109249 &  0.126380 &  0.128049 & \multicolumn{2}{c}{3.276e-03} & \multicolumn{2}{c}{1.606e-03} \\
        0.30 & 0.178456 & 0.158294 & 0.150368 &  0.172049 &  0.176617 & \multicolumn{2}{c}{6.406e-03} & \multicolumn{2}{c}{1.839e-03} \\
        0.40 & 0.209788 & 0.186074 & 0.176768 &  0.200917 &  0.207823 & \multicolumn{2}{c}{8.870e-03} & \multicolumn{2}{c}{1.964e-03} \\
        0.50 & 0.220584 & 0.195646 & 0.185865 &  0.210193 &  0.218584 & \multicolumn{2}{c}{1.039e-02} & \multicolumn{2}{c}{2.000e-03} \\
        0.60 & 0.209788 & 0.186074 & 0.176768 &  0.199015 &  0.207842 & \multicolumn{2}{c}{1.077e-02} & \multicolumn{2}{c}{1.945e-03} \\
        0.70 & 0.178456 & 0.158294 & 0.150368 &  0.168502 &  0.176620 & \multicolumn{2}{c}{9.954e-03} & \multicolumn{2}{c}{1.836e-03} \\
        0.80 & 0.129656 & 0.115020 & 0.109249 &  0.121621 &  0.127967 & \multicolumn{2}{c}{8.034e-03} & \multicolumn{2}{c}{1.688e-03} \\
        0.90 & 0.068164 & 0.060478 & 0.057436 &  0.062892 &  0.066726 & \multicolumn{2}{c}{5.272e-03} & \multicolumn{2}{c}{1.438e-03} \\
        1.00 & 0.000000 & 0.000000 & 0.000000 & -0.002058 & -0.001042 & \multicolumn{2}{c}{2.057e-03} & \multicolumn{2}{c}{1.041e-03} \\
        \bottomrule
    \end{tabular}
\end{table}

\begin{table}[ht]
    \centering
    \caption{Comparison among proposed model and baseline models at $t = 0.9$, with spatial domain $x \in [0,1]$,  spatial step $\delta x= 0.1$ for P1}
    \label{Table_t=0.9_(P1)}
    \scriptsize
    \begin{tabular}{c ccc cc cc}
        \toprule
        $\textbf{x}$ & \multicolumn{3}{c}{Baselines Results} & \multicolumn{2}{c}{Physics-informed Results} & \multicolumn{2}{c}{$E_1$ w.r.t.\ GT} \\
        \cmidrule(lr){2-4} \cmidrule(lr){5-6} \cmidrule(lr){7-8}
            & GT & FDM & SANN & PINN & A-PINN & PINN & A-PINN \\
        \midrule
        0.00 &  0.000000 &  0.000000 &  0.000000 & -0.008046 & -0.002901 & 8.045862e-03 & 2.900e-03 \\
        0.10 & -0.264707 & -0.256889 & -0.257566 & -0.263632 & -0.264740 & 1.074581e-03 & 3.352e-05 \\
        0.20 & -0.503502 & -0.488560 & -0.489919 & -0.494200 & -0.500621 & 9.302905e-03 & 2.881e-03 \\
        0.30 & -0.693012 & -0.672374 & -0.674316 & -0.677241 & -0.687675 & 1.577079e-02 & 5.336e-03 \\
        0.40 & -0.814684 & -0.790373 & -0.792706 & -0.794919 & -0.807787 & 1.976471e-02 & 6.897e-03 \\
        0.50 & -0.856610 & -0.831030 & -0.833500 & -0.835738 & -0.849170 & 2.087158e-02 & 7.439e-03 \\
        0.60 & -0.814684 & -0.790373 & -0.792706 & -0.795643 & -0.807829 & 1.904115e-02 & 6.854e-03 \\
        0.70 & -0.693012 & -0.672374 & -0.674316 & -0.678435 & -0.687977 & 1.457648e-02 & 5.034e-03 \\
        0.80 & -0.503502 & -0.488560 & -0.489919 & -0.495444 & -0.501338 & 8.058452e-03 & 2.164e-03 \\
        0.90 & -0.264707 & -0.256889 & -0.257566 & -0.264490 & -0.265915 & 2.168957e-04 & 1.207e-03 \\
        1.00 &  0.000000 &  0.000000 &  0.000000 & -0.008229 & -0.004620 & 8.228837e-03 & 4.620e-03 \\
        \bottomrule
    \end{tabular}
\end{table}

  
\subsection{Undamped Force Vibration of the Euler-Bernoulli Beam}

With an external load, the EBB model accounts for inertia, bending stiffness, 
and the applied excitation, enabling the analysis of force vibration phenomena. 
The forcing term introduces additional complexity and serves as a benchmark 
for assessing NN-based solvers. It takes the form

\begin{equation}
u_{tt}(x,t) + u_{xxxx}(x,t) = f(x,t), \quad x \in [0,\pi], \; t \in [0,1],
\label{P_2}
\end{equation}
The external force is defined as  
\begin{equation}
	f(x,t) = \left(1 - 16\pi^2\right)\sin(x)\cos(4\pi t).
	\label{eq:source}
\end{equation}
The beam is assumed to be simply supported at both ends, with BCs 
\(u(0,t) = u(\pi,t) = 0\) and \(u_{xx}(0,t) = u_{xx}(\pi,t) = 0\). The initial 
displacement and velocity are prescribed as \(u(x,0) = \sin(x)\) and 
\(u_t(x,0) = 0\). An auxiliary variable \(v(x,t) \approx u_{xx}(x,t)\) is introduced to simplify the fourth-order term.

As discussed in Eq. \eqref{A_Loss}, the total loss function is defined as
\begin{align}
\mathcal{L}(\theta) 
&= \tilde{w}_f \left( \frac{1}{N_f} \sum_{i=1}^{N_f} 
\Big| \widehat{u}_{tt}(x_i^f,t_i^f;\theta) + \widehat{v}_{xx}(x_i^f,t_i^f;\theta) - f(x_i^f,t_i^f;\theta) \Big|^2 \right) \notag \\
&\quad + \tilde{w}_\text{a} \left( \frac{1}{N_\text{a}} \sum_{i=1}^{N_\text{a}} 
\Big| \widehat{v}(x_i^\text{a},t_i^\text{a};\theta) - \widehat{u}_{xx}(x_i^\text{a},t_i^\text{a};\theta) \Big|^2 \right) \notag \\
&\quad + \tilde{w}_0 \left( \frac{1}{N_0} \sum_{i=1}^{N_0} 
\Big( \big|\widehat{u}(x_i^0,0;\theta) - \sin(x_i^0)\big|^2 + \big|\widehat{u}_t(x_i^0,0;\theta)\big|^2 \Big) \right) \notag \\
&\quad + \tilde{w}_b \left( \frac{1}{N_b} \sum_{i=1}^{N_b} 
\Big( \big|\widehat{u}(0,t_i^b;\theta)\big|^2 + \big|\widehat{u}(\pi,t_i^b;\theta)\big|^2 
+ \big|\widehat{v}(0,t_i^b;\theta)\big|^2 + \big|\widehat{v}(\pi,t_i^b;\theta)\big|^2 \Big) \right),
\end{align}
where \( \widehat{u}(x,t;\theta) \) and \( \widehat{v}(x,t;\theta) \) denote the NN outputs corresponding to the displacement field and its auxiliary variable, respectively.

In this experiment, the training dataset consists of $N_0 = 400$ initial points to enforce displacement and velocity conditions, $N_b = 400$ points at each boundary $x=0,\pi$ for simply supported BCs, $N_f = 500$ collocation points for minimizing the PDE residual, and $N_\text{a} = 500$ auxiliary points to enforce the relation $v \approx u_{xx}$, thereby improving stability and accuracy. These subsets collectively ensure that ICs, BCs, PDE residuals, and auxiliary constraints are satisfied during optimization.

\begin{figure}[H]
  \centering
  \captionsetup[sub]{justification=centering}
  {\large Physics-informed solutions}\par\vspace{0.4em}

  \begin{subfigure}{0.48\textwidth}
    \centering
    \includegraphics[width=\linewidth]{ 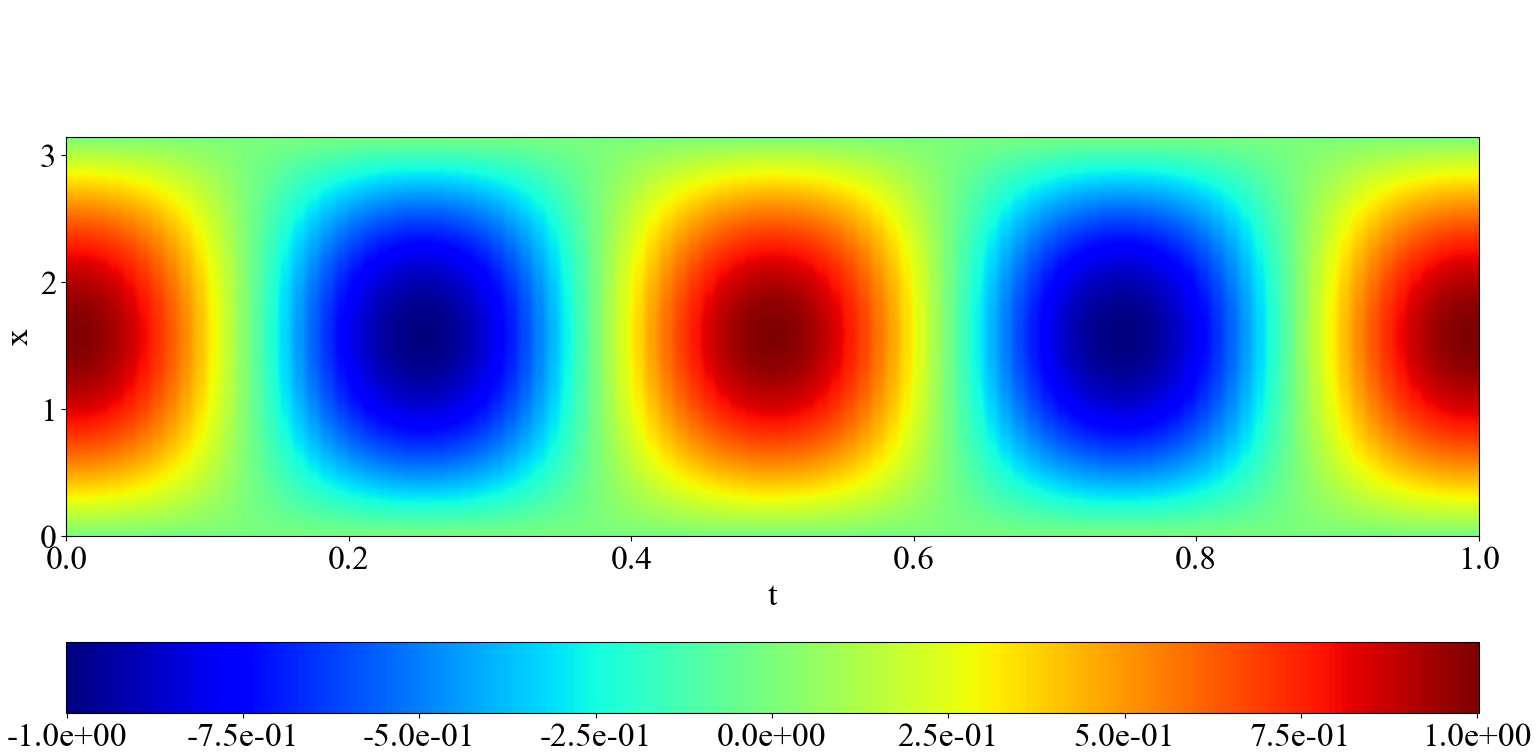}
    \caption{PINN}
  \end{subfigure}
  \hfill
  \begin{subfigure}{0.48\textwidth}
    \centering
    \includegraphics[width=\linewidth]{ 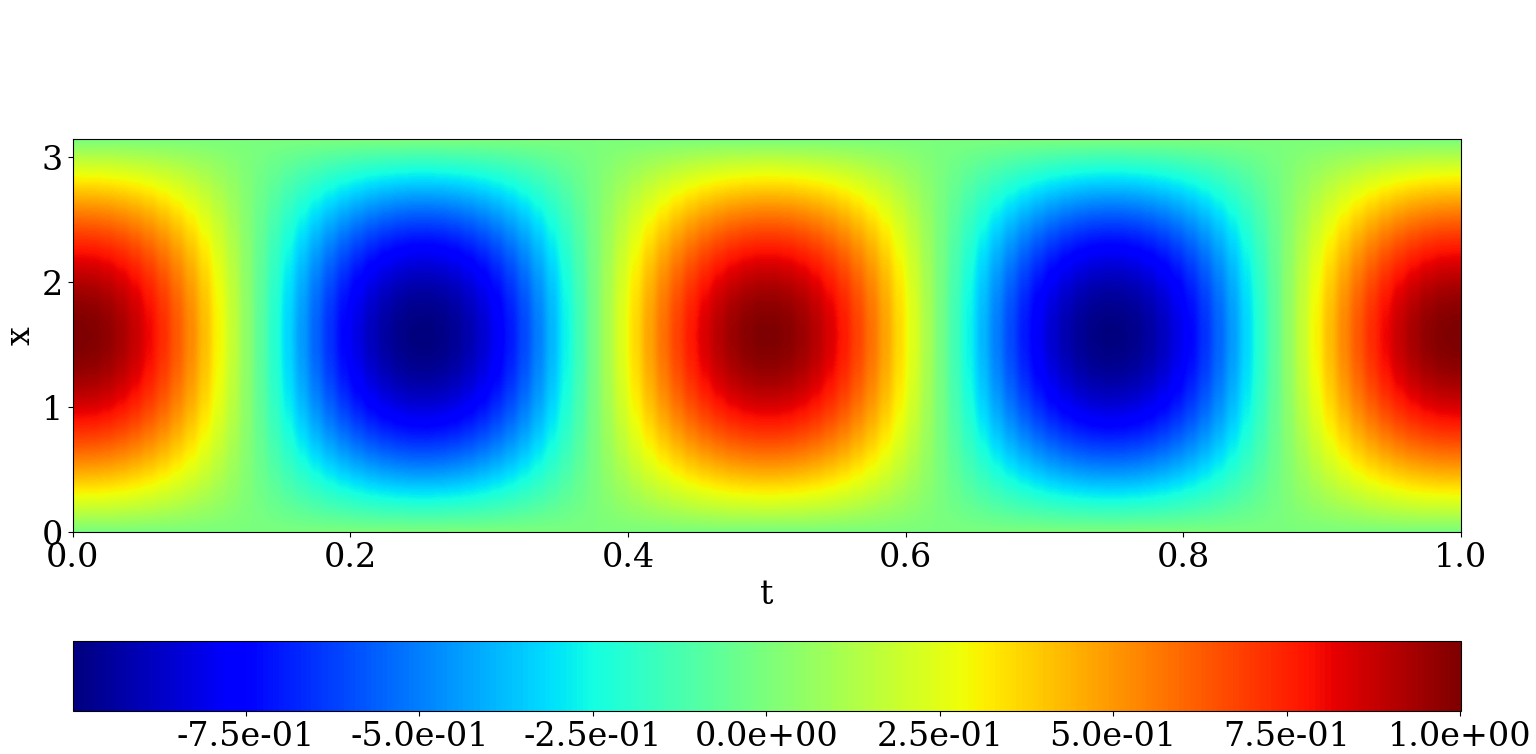}
    \caption{A-PINN (Proposed model)}
  \end{subfigure}

  \vspace{1.0em} 

{\large Baseline solutions}\par\vspace{0.4em}

  \begin{subfigure}{0.48\textwidth}
    \centering
    \includegraphics[width=\linewidth]{ 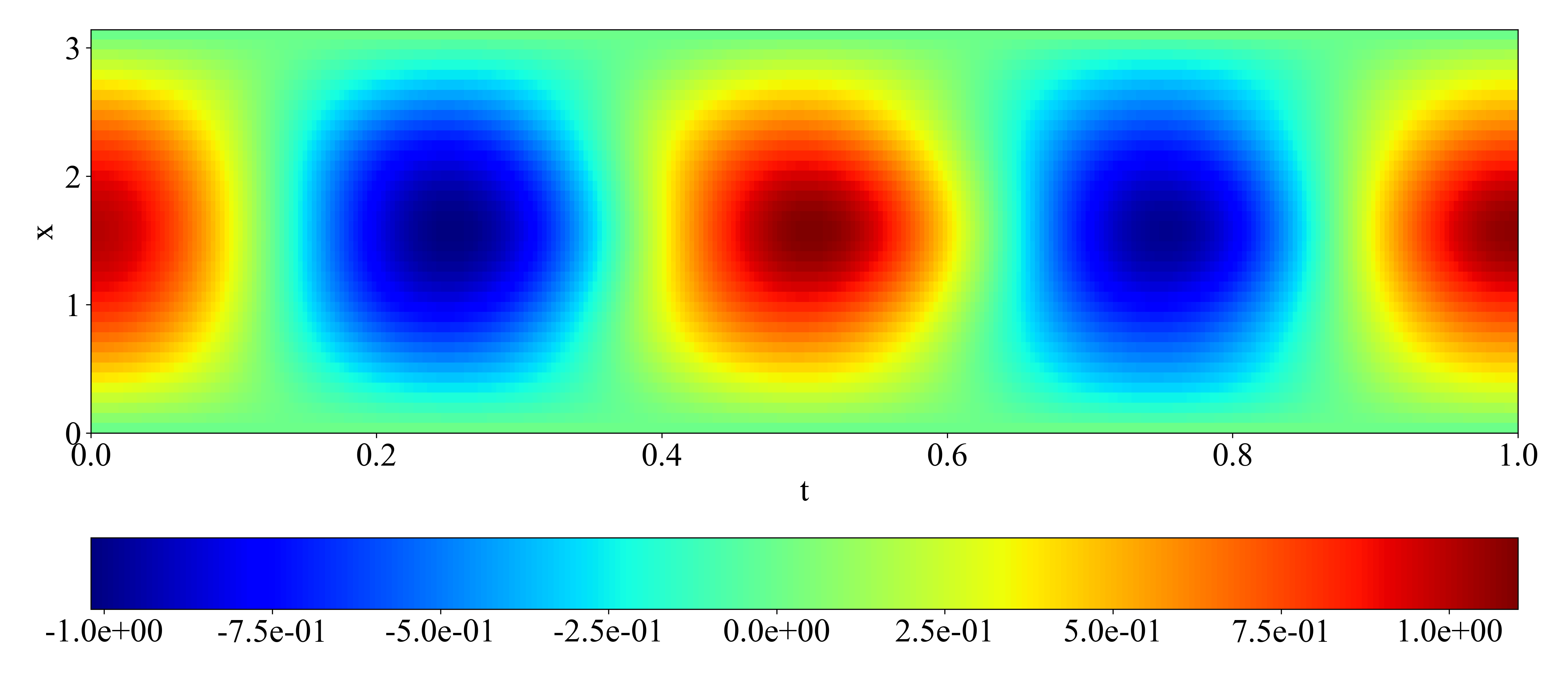}
    \caption{FDM}
  \end{subfigure}
  \hfill
  \begin{subfigure}{0.48\textwidth}
    \centering
    \includegraphics[width=\linewidth]{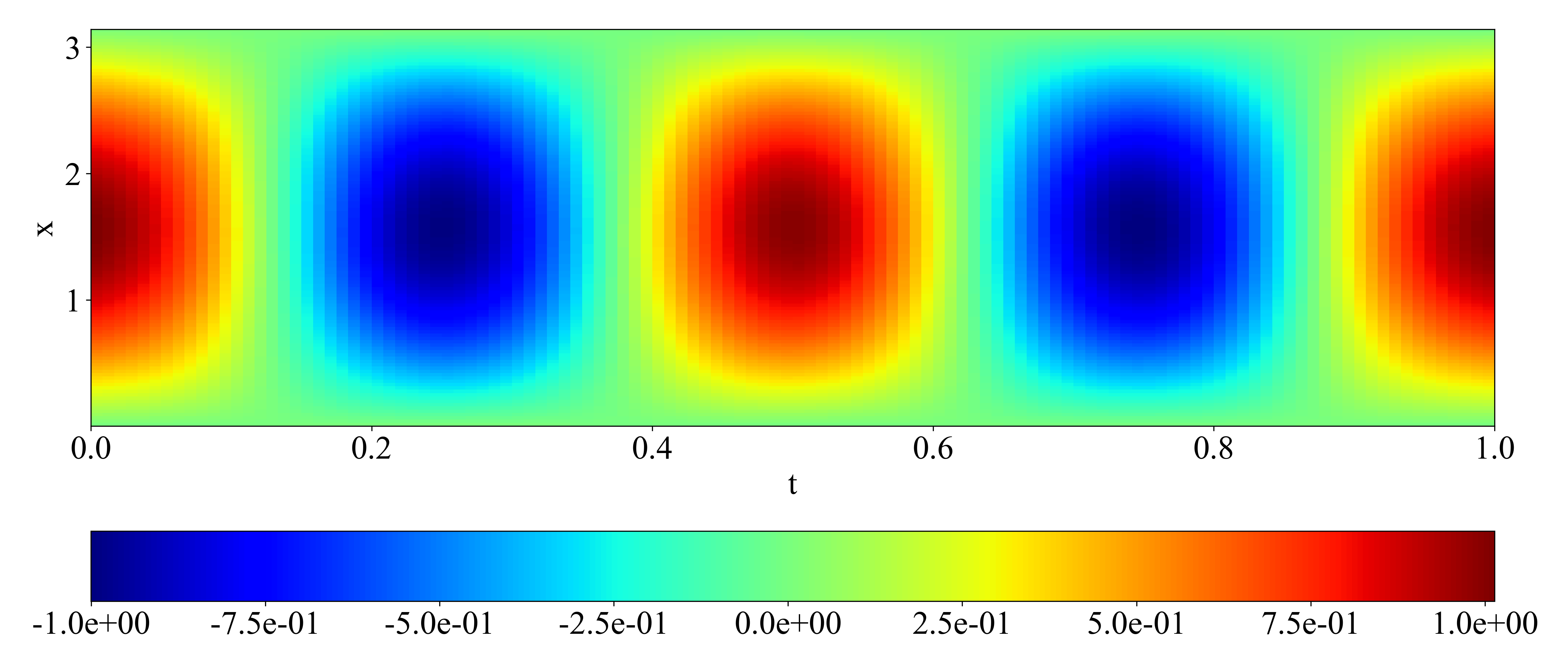}
    \caption{SANN}
  \end{subfigure}

 \vskip\baselineskip 
  \begin{subfigure}{0.48\textwidth}
    \centering
    \includegraphics[width=\linewidth]{ 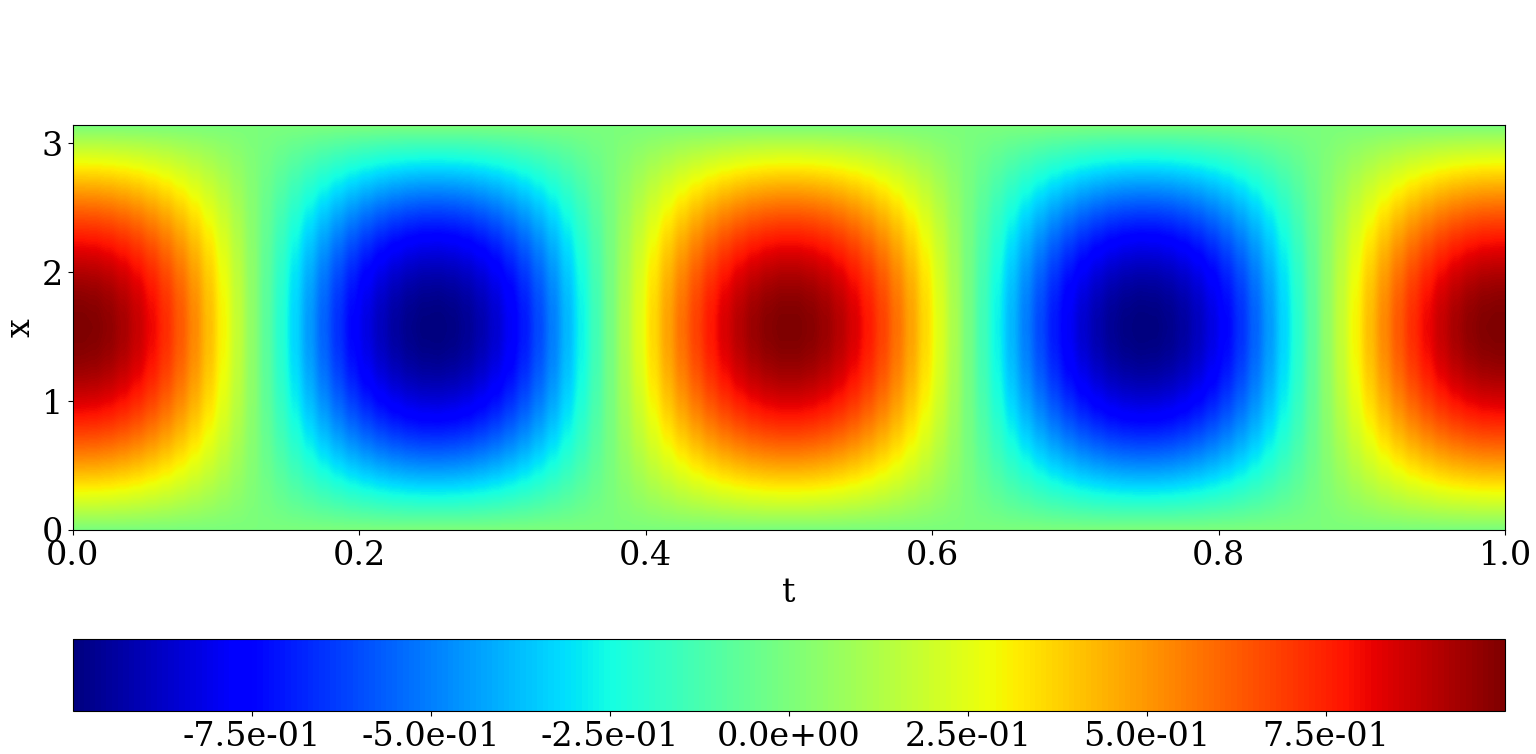}
    \caption{GT}
  \end{subfigure}

 \caption{2D comparison of physics-informed (top) and baseline (bottom) solutions for P2.}
  \label{2d_blocks(P2)}
\end{figure}

\begin{figure}[H]
  \centering
  \captionsetup[sub]{justification=centering}

  {\large Physics-informed solutions}\par\vspace{0.4em}

  \begin{subfigure}{0.48\textwidth}
    \centering
    \includegraphics[width=\linewidth]{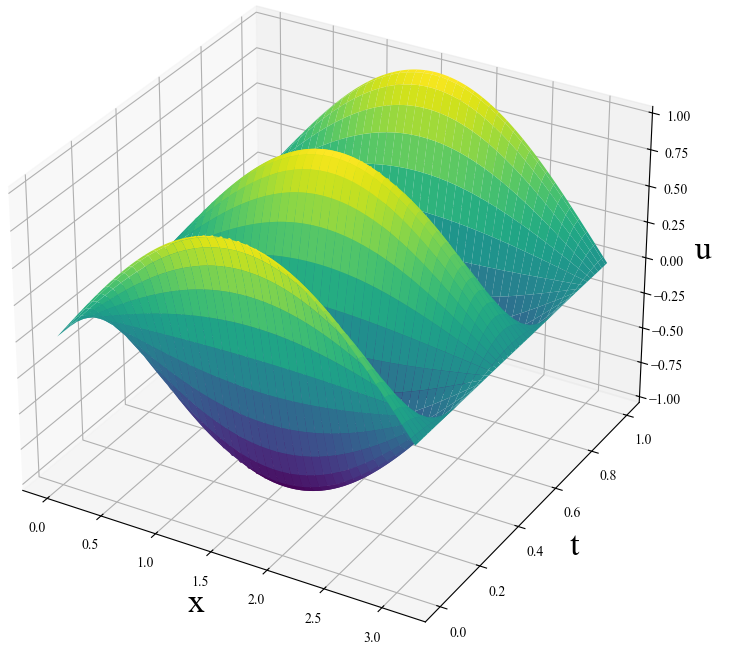}
    \caption{PINN}
  \end{subfigure}
  \hfill
  \begin{subfigure}{0.48\textwidth}
    \centering
    \includegraphics[width=\linewidth]{ 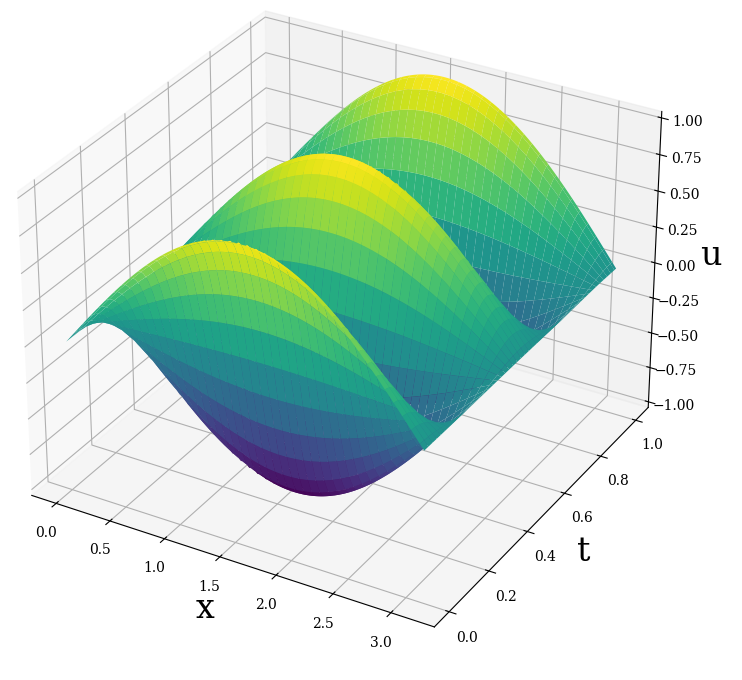}
    \caption{A-PINN (Proposed model)}
  \end{subfigure}

  \vspace{1.0em} 

  {\large Baseline solutions}\par\vspace{0.4em}

  
  \begin{subfigure}{0.45\textwidth}
    \centering
    \includegraphics[width=\linewidth]{ 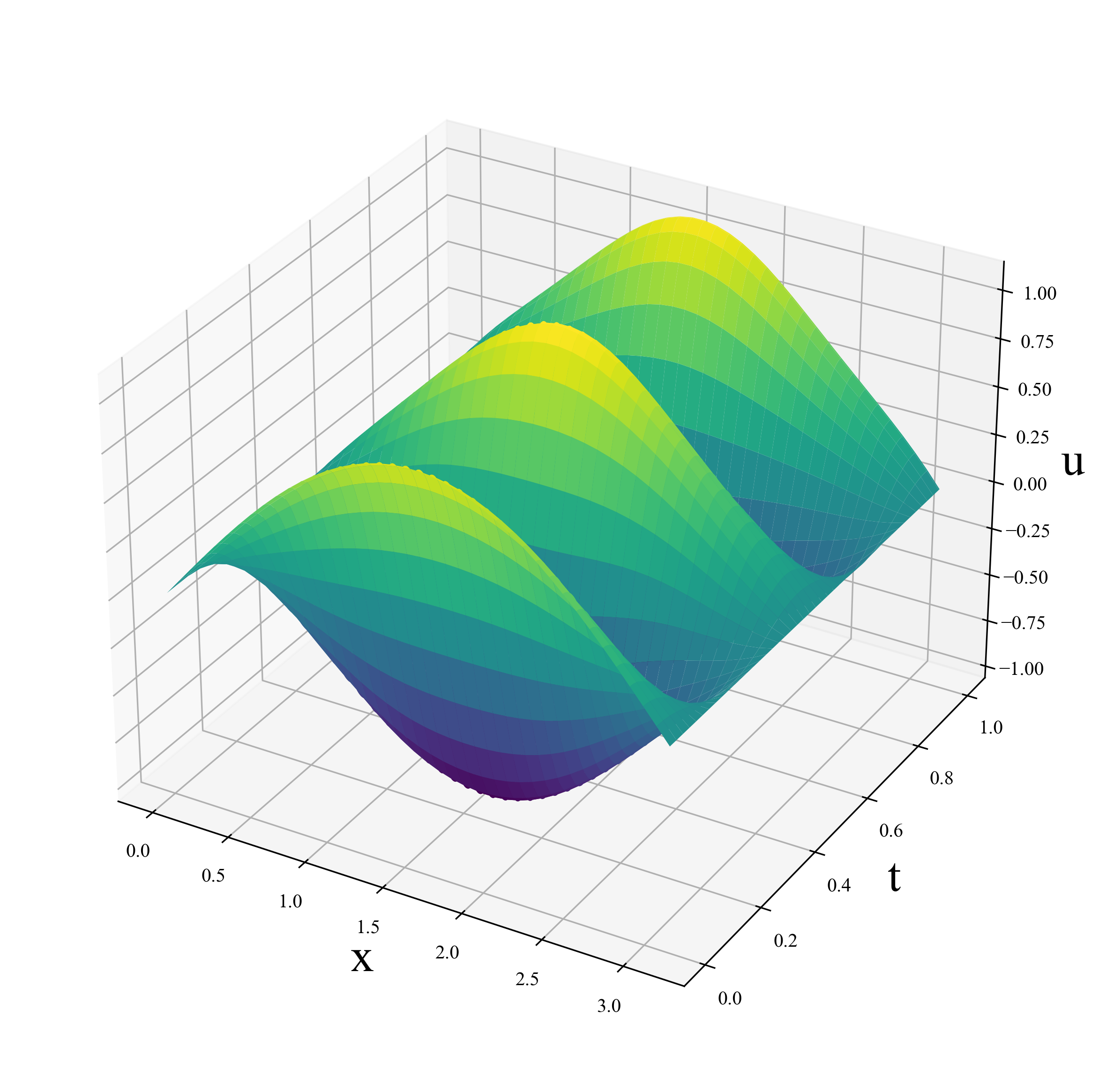}
    \caption{FDM}
  \end{subfigure}
  \hfill
  \begin{subfigure}{0.45\textwidth}
    \centering
    \includegraphics[width=\linewidth]{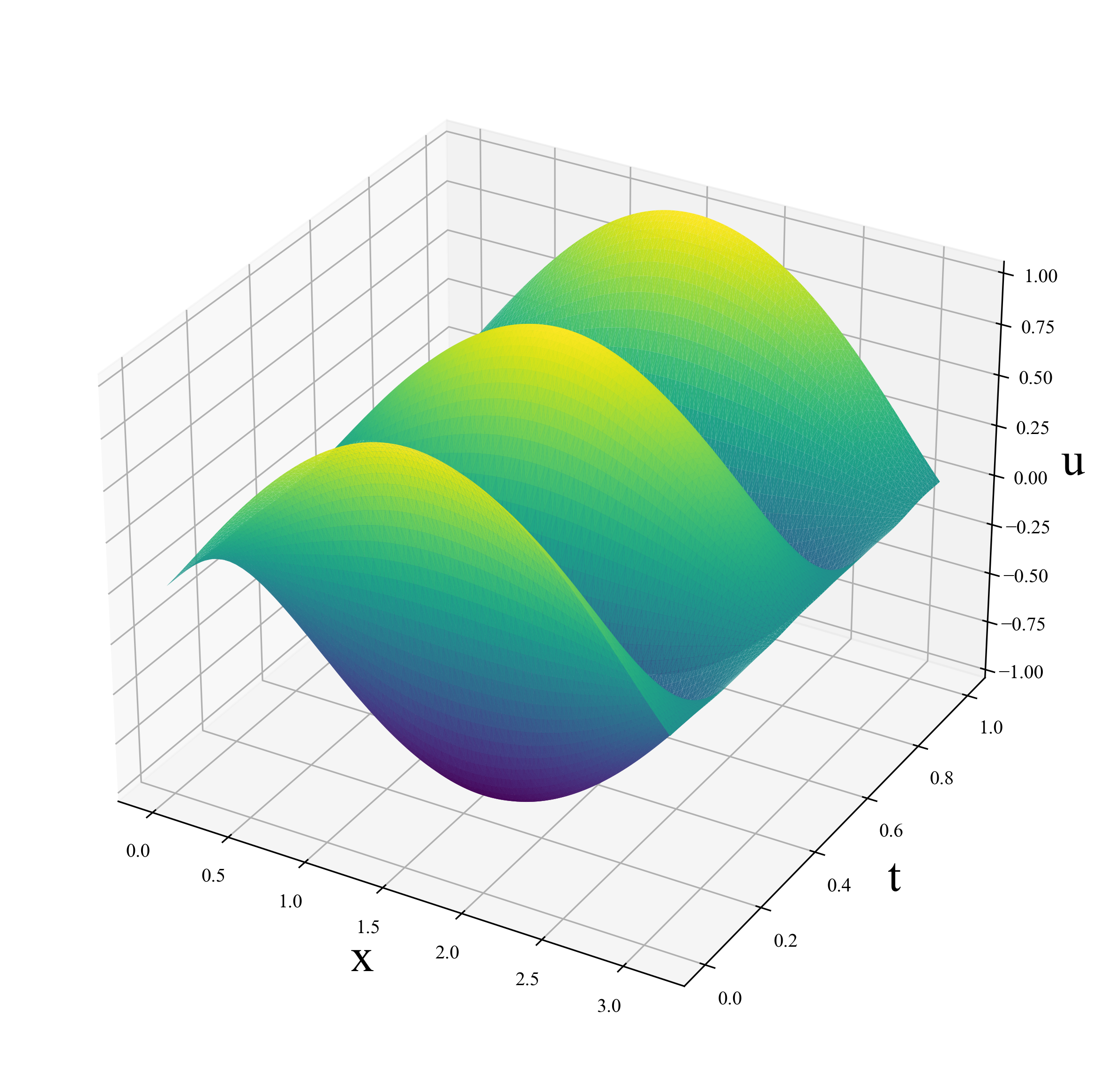}
    \caption{SANN}
  \end{subfigure}
   \vskip\baselineskip 
  \begin{subfigure}{0.45\textwidth} 
    \centering
    \includegraphics[width=\linewidth]{ 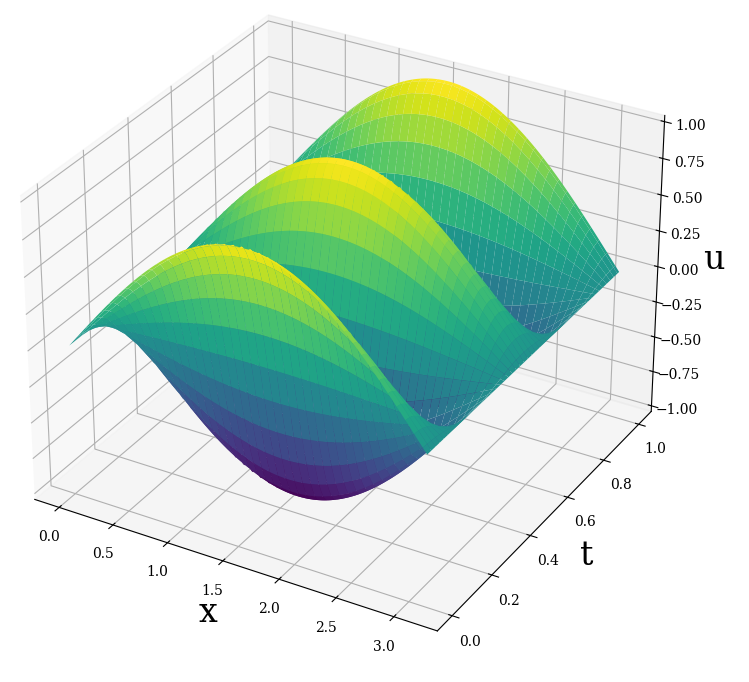}
    \caption{GT}
  \end{subfigure}
  \caption{3D comparison of physics-informed (top) and baseline (bottom) solutions for P2.}
  \label{3D_block(P2)}
\end{figure}

The effectiveness of the A-PINN is evaluated on the EBB model, considering the case of undamped free vibration under simply supported BCs. The exact solution \cite{kapoor2023physics} serves as the GT for comparison with the PINN, A-PINN, FDM, and SANN. Figures~\ref{2d_blocks(P2)} and~\ref{3D_block(P2)} demonstrate that A-PINN achieves closer agreement with the GT, capturing smooth spatio-temporal variations, while the PINN, and classical baselines SANN and FDM display larger deviations.

\begin{figure}[H]
    \centering
    \begin{subfigure}{0.45\textwidth}
        \centering
        \includegraphics[width=\linewidth]{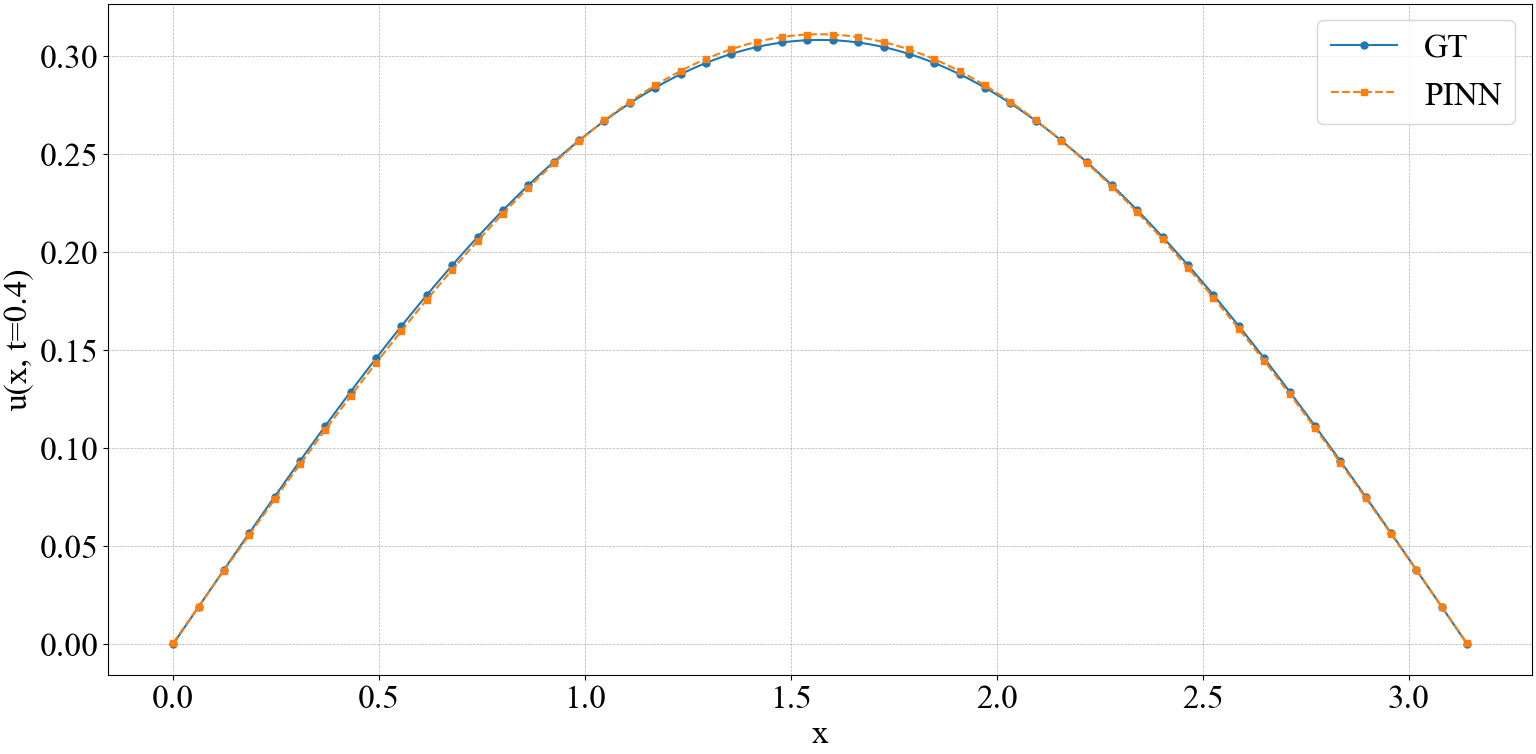}
        \caption{t=0.4 using PINN}
    \end{subfigure}
    \hfill
    \begin{subfigure}{0.45\textwidth}
        \centering
        \includegraphics[width=\linewidth]{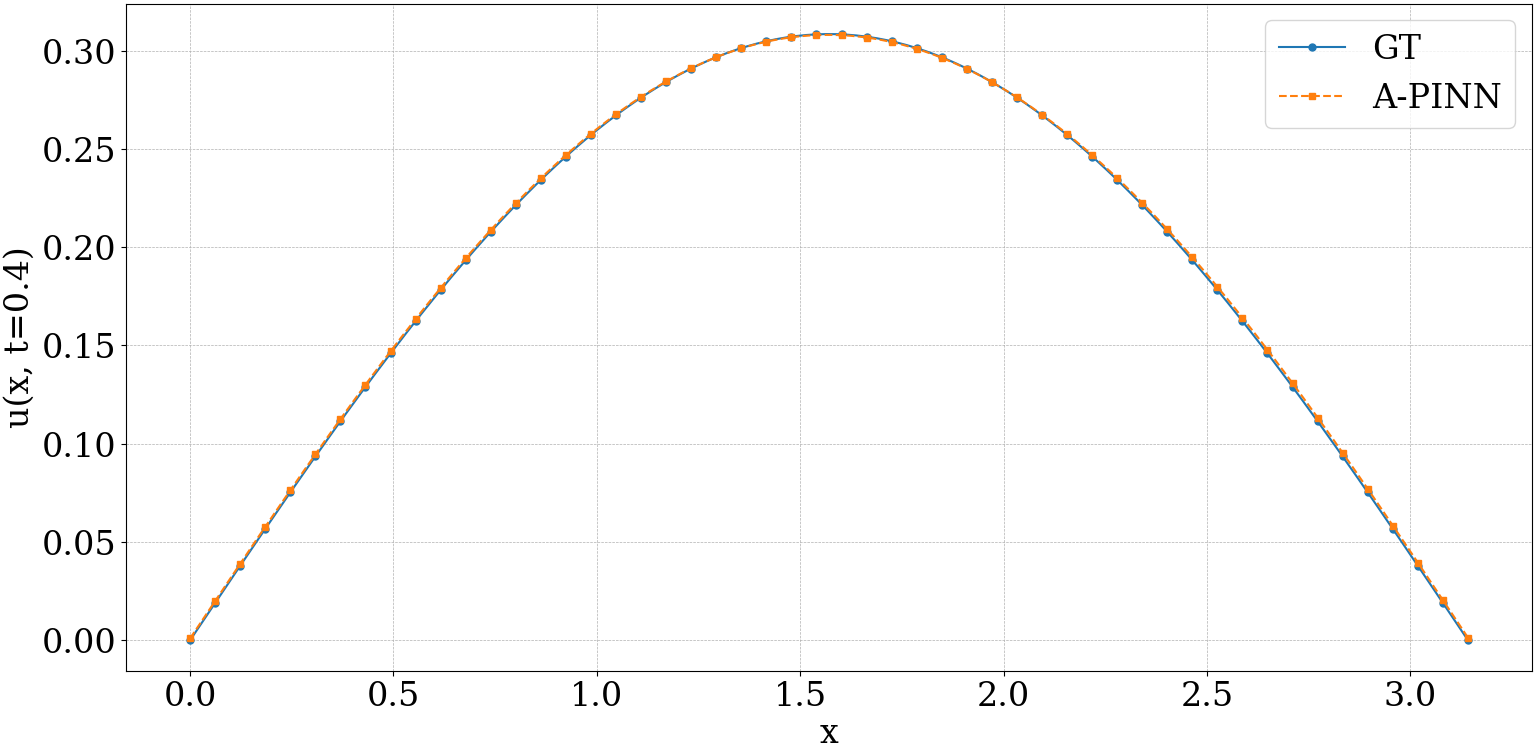}
        \caption{t=0.4 using A-PINN}
    \end{subfigure}

    \vskip\baselineskip 

    \begin{subfigure}{0.45\textwidth}
        \centering
        \includegraphics[width=\linewidth]{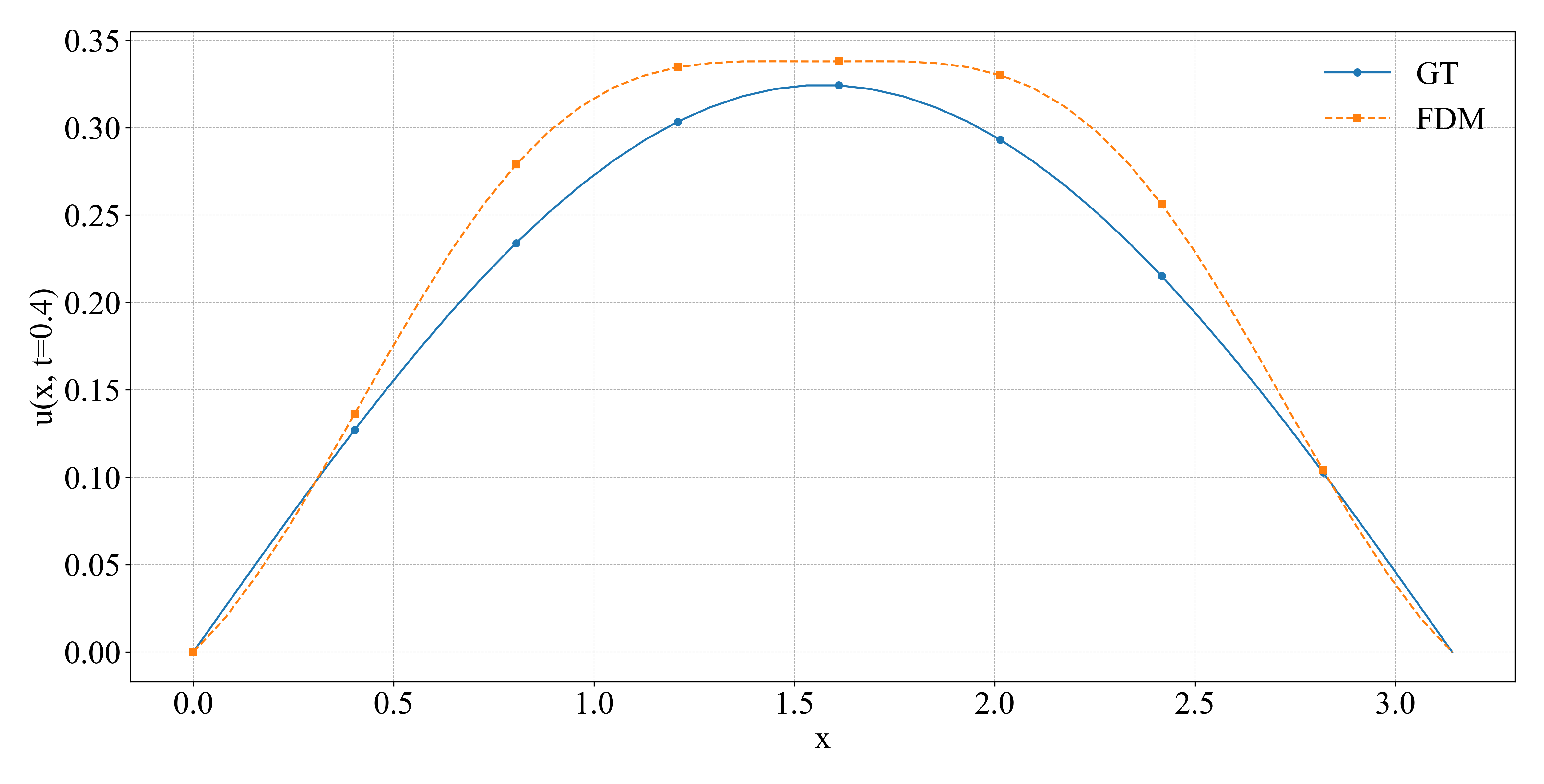}
        \caption{t=0.4 using FDM}
    \end{subfigure}
    \hfill
    \begin{subfigure}{0.45\textwidth}
        \centering
        \includegraphics[width=\linewidth]{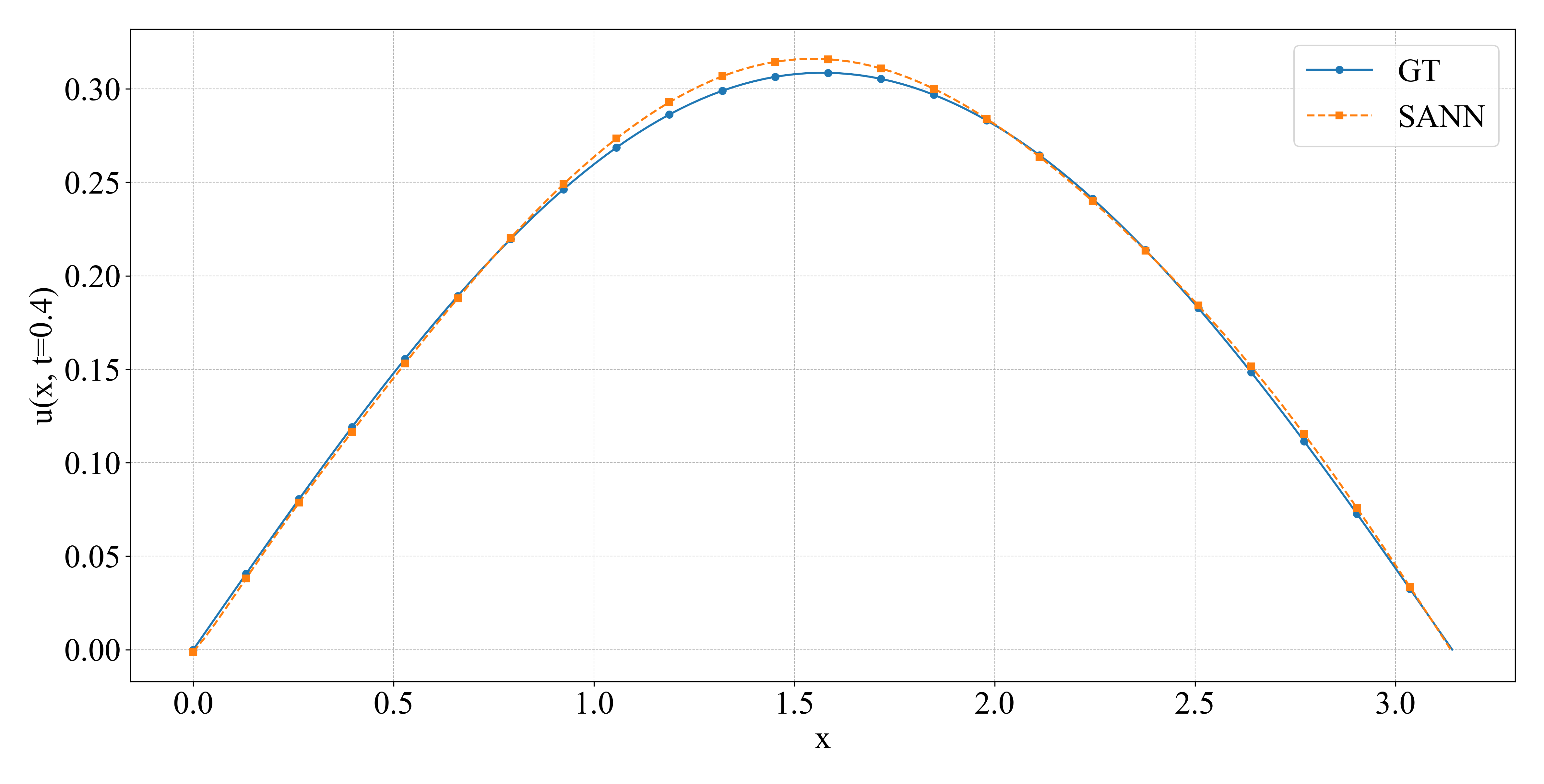}
        \caption{t=0.4 using SANN}
    \end{subfigure}

    \caption{Comparison of PINN, A-PINN, FDM, and SANN solutions with GT at $t=0.4$ for P2.}
    \label{2D_slice_t=0.4_(P2)}
\end{figure}

\begin{figure}[H]
    \centering
    \begin{subfigure}{0.45\textwidth}
        \centering
        \includegraphics[width=\linewidth]{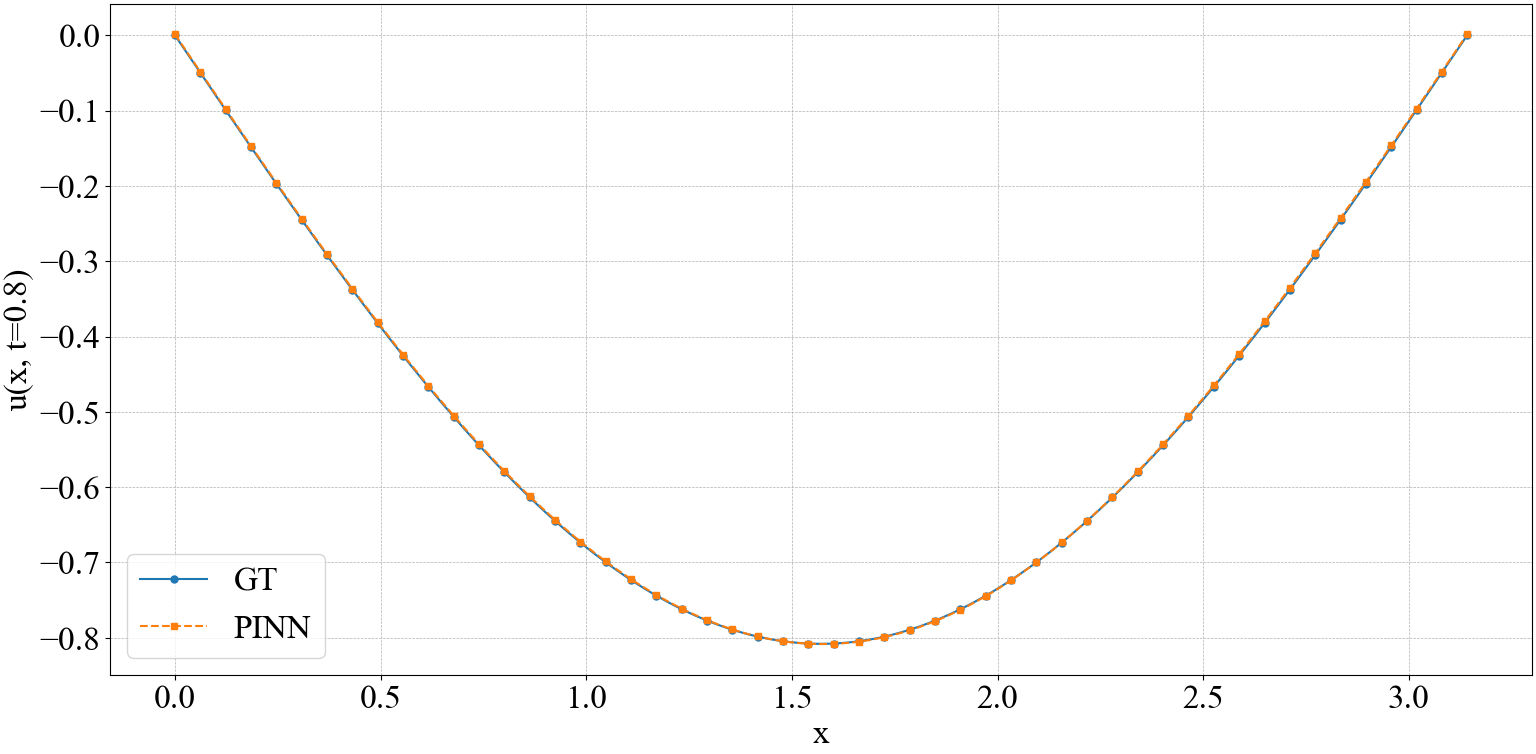}
        \caption{t=0.8 using PINN}
    \end{subfigure}
    \hfill
    \begin{subfigure}{0.45\textwidth}
        \centering
        \includegraphics[width=\linewidth]{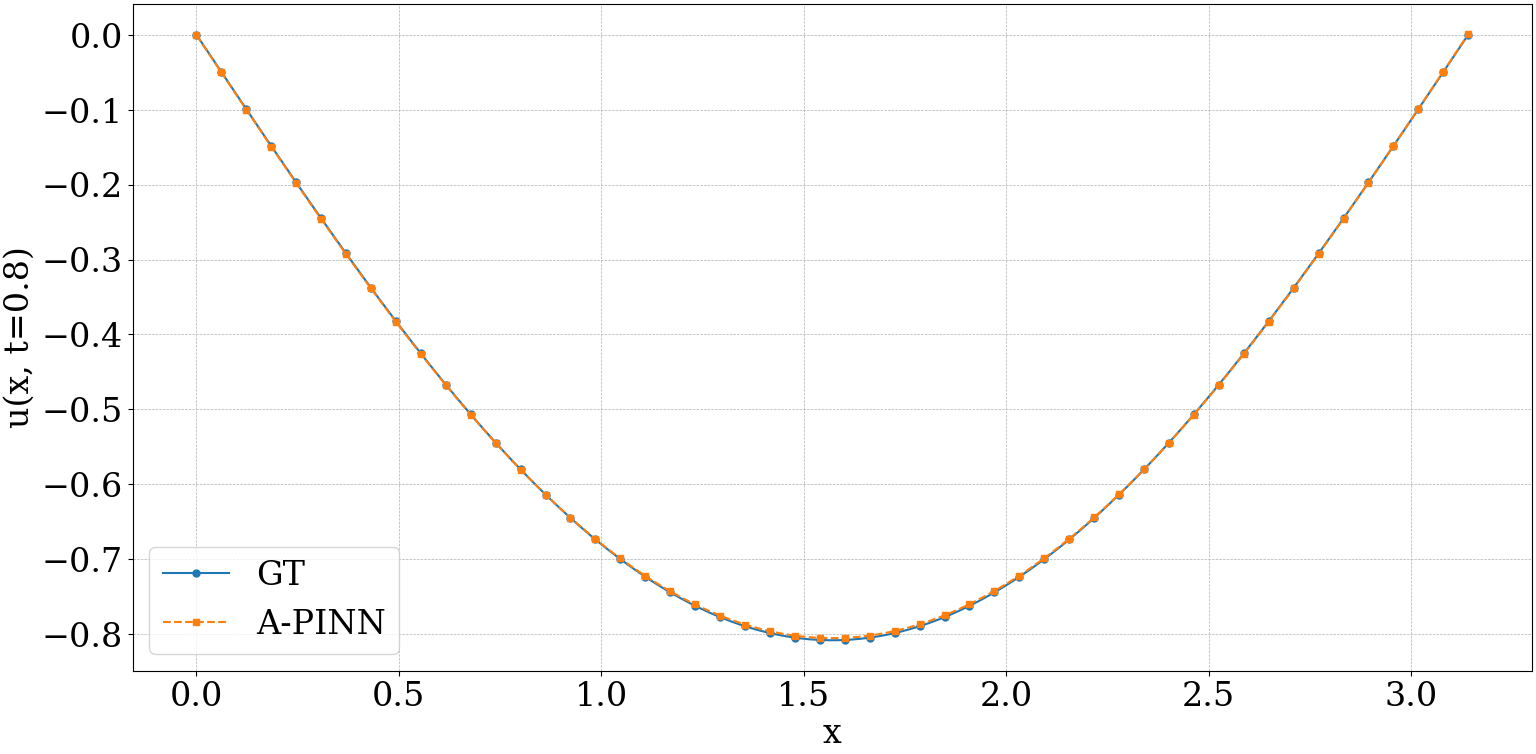}
        \caption{t=0.8 using A-PINN}
    \end{subfigure}

    \vskip\baselineskip 

    \begin{subfigure}{0.45\textwidth}
        \centering
        \includegraphics[width=\linewidth]{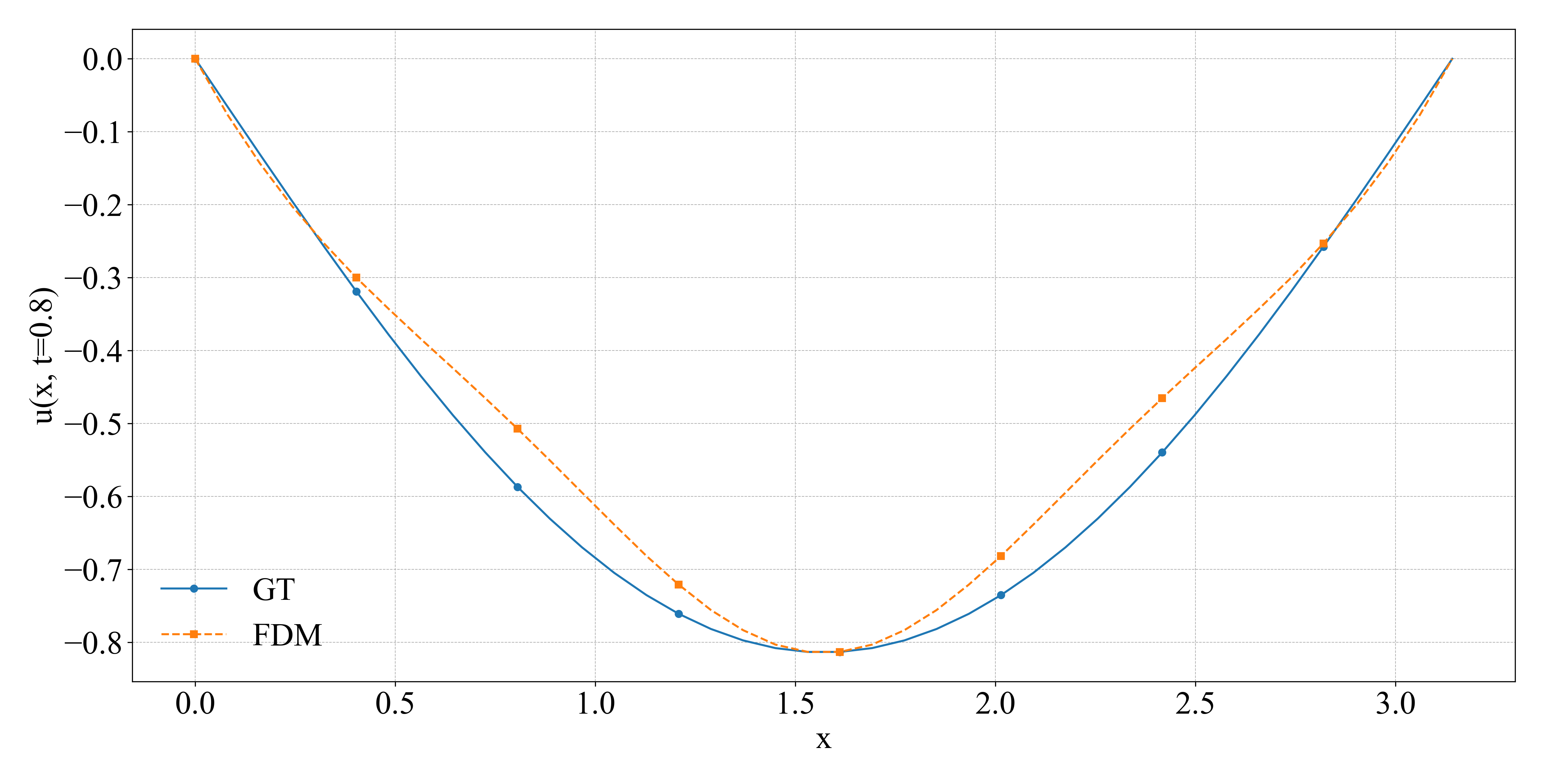}
        \caption{t=0.8 using FDM}
    \end{subfigure}
    \hfill
    \begin{subfigure}{0.45\textwidth}
        \centering
        \includegraphics[width=\linewidth]{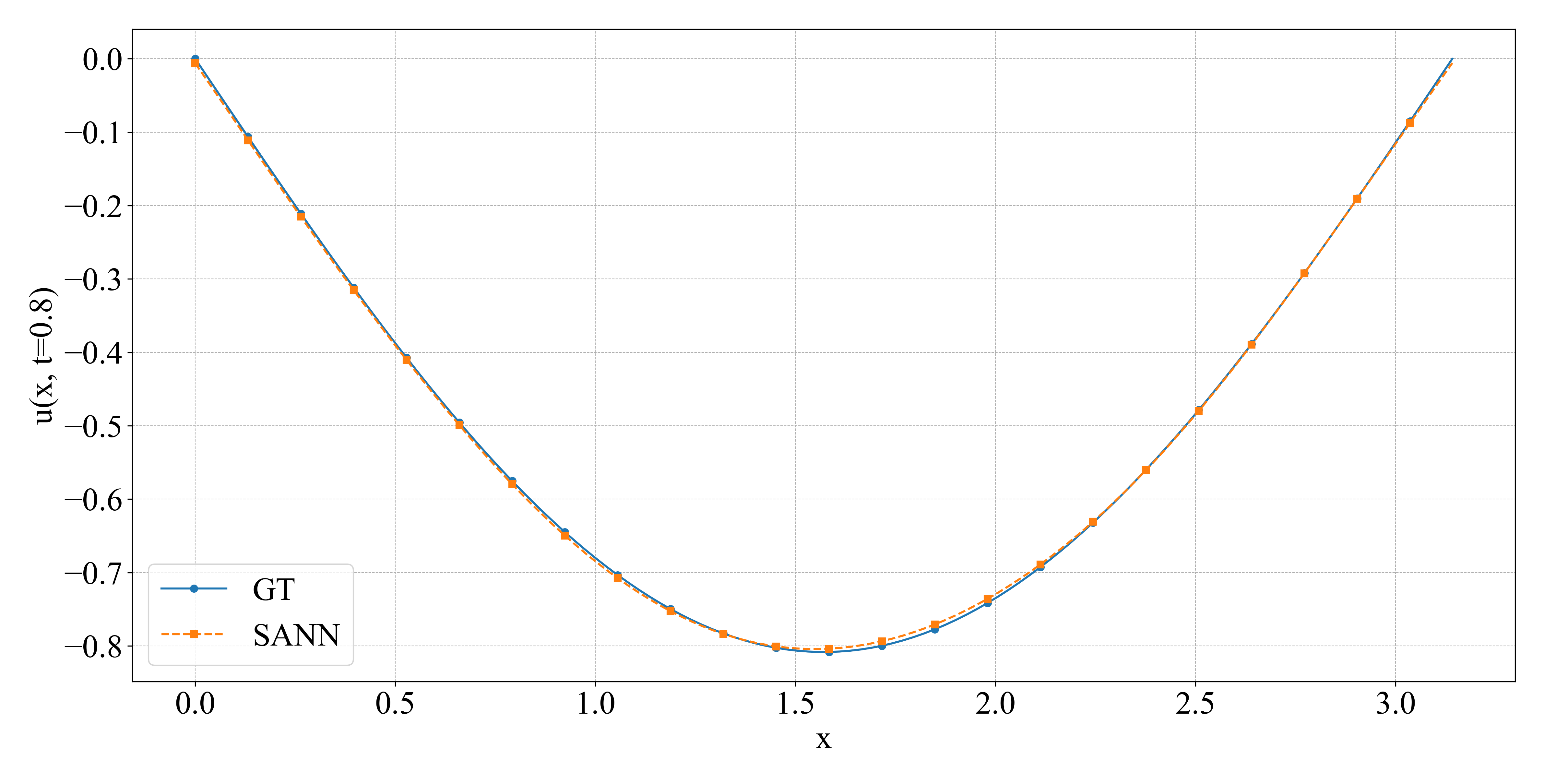}
        \caption{t=0.8 using SANN}
    \end{subfigure}

    \caption{Comparison of PINN, A-PINN, FDM, and SANN solutions with GT at $t=0.8$ for P2.}
    \label{2D_slice_t=0.8_(P2)}
\end{figure}

To gain clearer insight into the dynamic behavior, comparisons are made at two representative time instances, $t=0.4$ (cf.: Figure~\ref{2D_slice_t=0.4_(P2)} and  Table~\ref{Table_t=0.4_(P2)}) and $t=0.8$ (cf.: Figure~\ref{2D_slice_t=0.8_(P2)} and Table~\ref{Table_t=0.8_(P2)}). At $t=0.4$, the displacement profile exhibits reduced amplitude, reflecting an intermediate oscillation stage. Here, the A-PINN prediction remains nearly indistinguishable from the GT, while the PINN slightly underestimates the peak response. At $t=0.8$, the displacement profile is inverted relative to the initial state, corresponding to a phase reversal of the vibration. Unlike the PINN and the baselines FDM and SANN, which fail to capture the phase reversal precisely, the A-PINN reproduces it in close agreement with the GT.

The temporal variation in the displacement profiles arises naturally from the vibration dynamics of the beam. The spatial mode $\sin(x)$ remains fixed due to the simply supported BCs, while the temporal factor $\cos(4\pi t)$ controls the oscillatory amplitude and phase. Maximum displacements occur when $\cos(4\pi t)=\pm 1$, corresponding to peak strain energy and zero kinetic energy, while near $\cos(4\pi t)=0$ the displacement vanishes, and kinetic energy dominates. This periodic exchange between strain and kinetic energy characterizes the natural vibration of the beam. Overall, the A-PINN consistently achieves closer agreement with the GT, outperforming the PINN and traditional numerical baselines, thereby validating its robustness for high-order structural vibration problems.

\begin{table}[ht]
    \centering
    \caption{Comparison among proposed model and baseline models at $t = 0.4$, with spatial domain $x \in [0,\pi]$, spatial step $\Delta x = 0.31$ for P2}
    \label{Table_t=0.4_(P2)}
    \scriptsize
    \begin{tabular}{c ccc cc cc}
        \toprule
        $\textbf{x}$ & \multicolumn{3}{c}{Baselines Results} & \multicolumn{2}{c}{Physics-informed Results} & \multicolumn{2}{c}{$E_1$ w.r.t.\ GT} \\
        \cmidrule(lr){2-4} \cmidrule(lr){5-6} \cmidrule(lr){7-8}
            & GT & FDM & SANN & PINN & A-PINN & PINN & A-PINN \\
        \midrule
        0.00 & 0.000000 & 0.000000 & 0.000405 & 0.000622 & 0.00088 & 6.224e-04 & 8.800e-04 \\
        0.31 & 0.095492 & 0.101028 & 0.094749 & 0.093756 & 0.09488 & 1.735e-03 & 1.087e-03 \\
        0.63 & 0.181636 & 0.224066 & 0.180100 & 0.179274 & 0.18255 & 2.361e-03 & 5.000e-04 \\
        0.94 & 0.250000 & 0.307731 & 0.251820 & 0.249352 & 0.25021 & 6.478e-04 & 6.600e-04 \\
        1.26 & 0.293893 & 0.336002 & 0.298816 & 0.295753 & 0.29495 & 1.861e-03 & 7.300e-04 \\
        1.57 & 0.309017 & 0.337926 & 0.313155 & 0.311911 & 0.30945 & 2.894e-03 & 4.300e-04 \\
        1.88 & 0.293893 & 0.336002 & 0.292584 & 0.295597 & 0.29442 & 1.705e-03 & 6.000e-05 \\
        2.20 & 0.250000 & 0.307731 & 0.244910 & 0.249643 & 0.24977 & 3.570e-04 & 7.000e-05 \\
        2.51 & 0.181636 & 0.224066 & 0.180143 & 0.180232 & 0.18292 & 1.404e-03 & 4.700e-04 \\
        2.83 & 0.095492 & 0.101028 & 0.097272 & 0.094596 & 0.09630 & 8.960e-04 & 1.560e-03 \\
        3.14 & 0.000000 & 0.000000 & -0.002172 & 0.000295 & 0.00213 & 2.946e-04 & 1.640e-03 \\
        \bottomrule
    \end{tabular}
\end{table}

\begin{table}[ht]
    \centering
    \caption{Comparison among proposed model and baseline models at $t = 0.8$, with spatial domain $x \in [0,\pi]$, spatial step $\Delta x = 0.31$ for P2}
    \label{Table_t=0.8_(P2)}
    \scriptsize
    \begin{tabular}{c ccc cc cc}
        \toprule
        $\textbf{x}$ & \multicolumn{3}{c}{Baselines Results} & \multicolumn{2}{c}{Physics-informed Results} & \multicolumn{2}{c}{$E_1$ w.r.t.\ GT} \\
        \cmidrule(lr){2-4} \cmidrule(lr){5-6} \cmidrule(lr){7-8}
            & GT & FDM & SANN & PINN & A-PINN & PINN & A-PINN \\
        \midrule
        0.00 &  0.000000 &  0.000000 & -0.004137 &  0.001258 & -0.000010 & 1.257e-03 & 1.000e-05 \\
        0.31 & -0.250000 & -0.248050 & -0.251666 & -0.248608 & -0.247410 & 1.391e-03 & 6.100e-04 \\
        0.63 & -0.475528 & -0.416310 & -0.476758 & -0.474050 & -0.477100 & 1.478e-03 & 4.700e-04 \\
        0.94 & -0.654508 & -0.581410 & -0.657566 & -0.653094 & -0.652940 & 1.413e-03 & 3.900e-04 \\
        1.26 & -0.769421 & -0.741750 & -0.770151 & -0.768687 & -0.768000` & 7.338e-04 & 2.260e-03 \\
        1.57 & -0.809017 & -0.813061 & -0.804124 & -0.809348 & -0.805740 & 3.313e-04 & 3.280e-03 \\
        1.88 & -0.769421 & -0.741750 & -0.762598 & -0.770121 & -0.768320 & 7.005e-04 & 2.330e-03 \\
        2.20 & -0.654508 & -0.581410 & -0.652402 & -0.654245 & -0.653490 & 2.635e-04 & 6.000e-04 \\
        2.51 & -0.475528 & -0.416310 & -0.476721 & -0.473735 & -0.477940 & 1.793e-03 & 2.700e-04 \\
        2.83 & -0.250000 & -0.248050 & -0.249760 & -0.247646 & -0.248410 & 2.353e-03 & 3.900e-04 \\
        3.14 &  0.000000 &  0.000000 & -0.004832 &  0.001304 &  -0.000810 & 1.303e-03 & 4.800e-04 \\
        \bottomrule
    \end{tabular}
\end{table}

\newpage

\subsection{Undamped Force Vibration of an Euler-Bernoulli Beam on a Winkler Foundation}

The EBB on a Winkler foundation can be modeled as a fourth-order PDE that includes both the bending stiffness and the elastic restoring effect of the foundation. The governing equation is defined as  
\begin{equation}
u_{tt}(x,t) + u_{xxxx}(x,t) + k\,u(x,t) = f(x,t), 
\quad x \in [0,3\pi], \; t \in [0,1],
\label{P_3}
\end{equation}
where \(u(x,t)\) denotes the transverse displacement of the beam and \(k\,u(x,t)\) models the linear reaction of the Winkler foundation with stiffness parameter \(k=1\). The external forcing is defined as  
\begin{equation}
	f(x,t) = (2 - \pi^2)\sin(x)\cos(\pi t).
	\label{eq:winkler_forcing}
\end{equation}

\begin{figure}[H]
  \centering
  \captionsetup[sub]{justification=centering}
  {\large Physics-informed solutions}\par\vspace{0.4em}

\begin{subfigure}{0.48\textwidth}
    \centering
    \includegraphics[width=\linewidth]{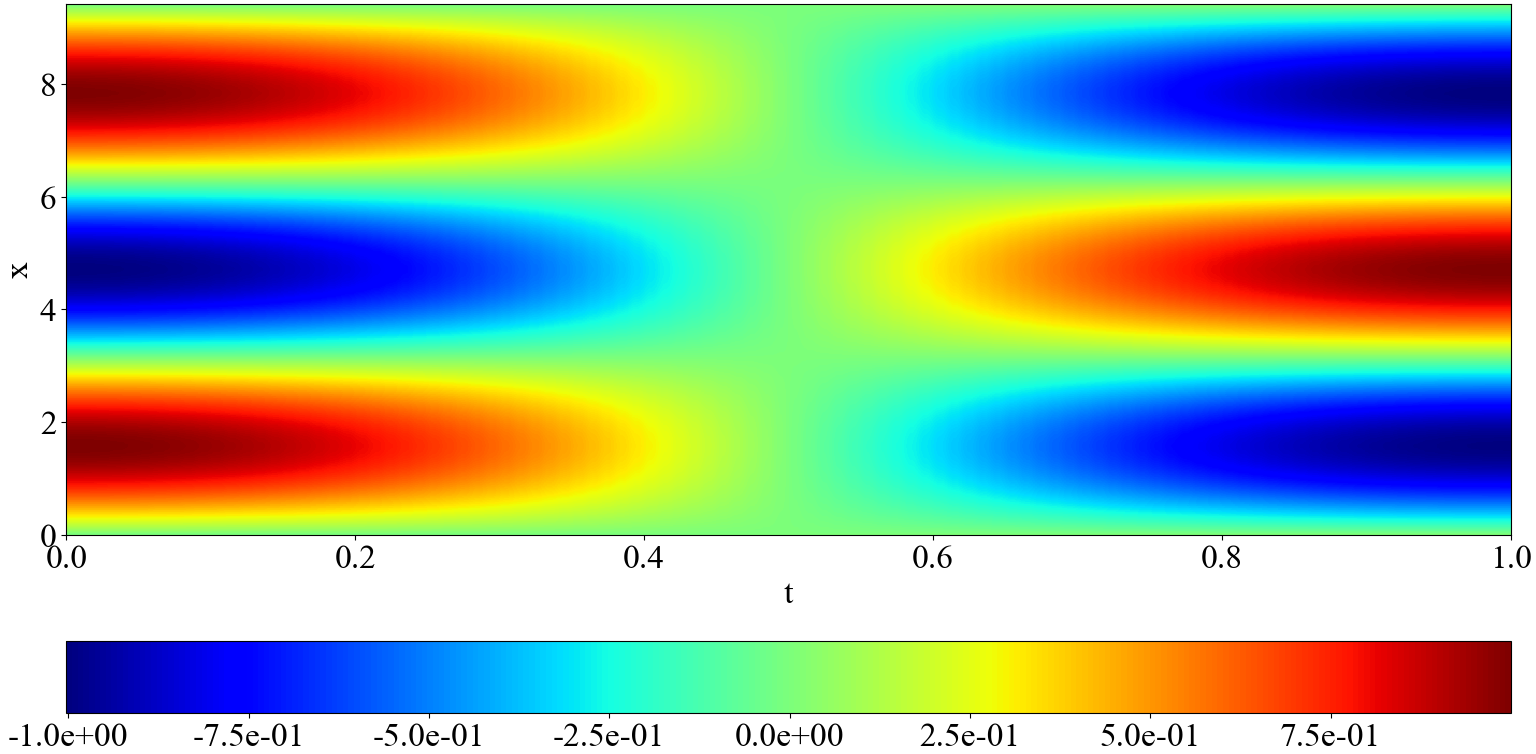}
    \caption{PINN}
  \end{subfigure}
  \hfill
  \begin{subfigure}{0.48\textwidth}
    \centering
    \includegraphics[width=\linewidth]{ 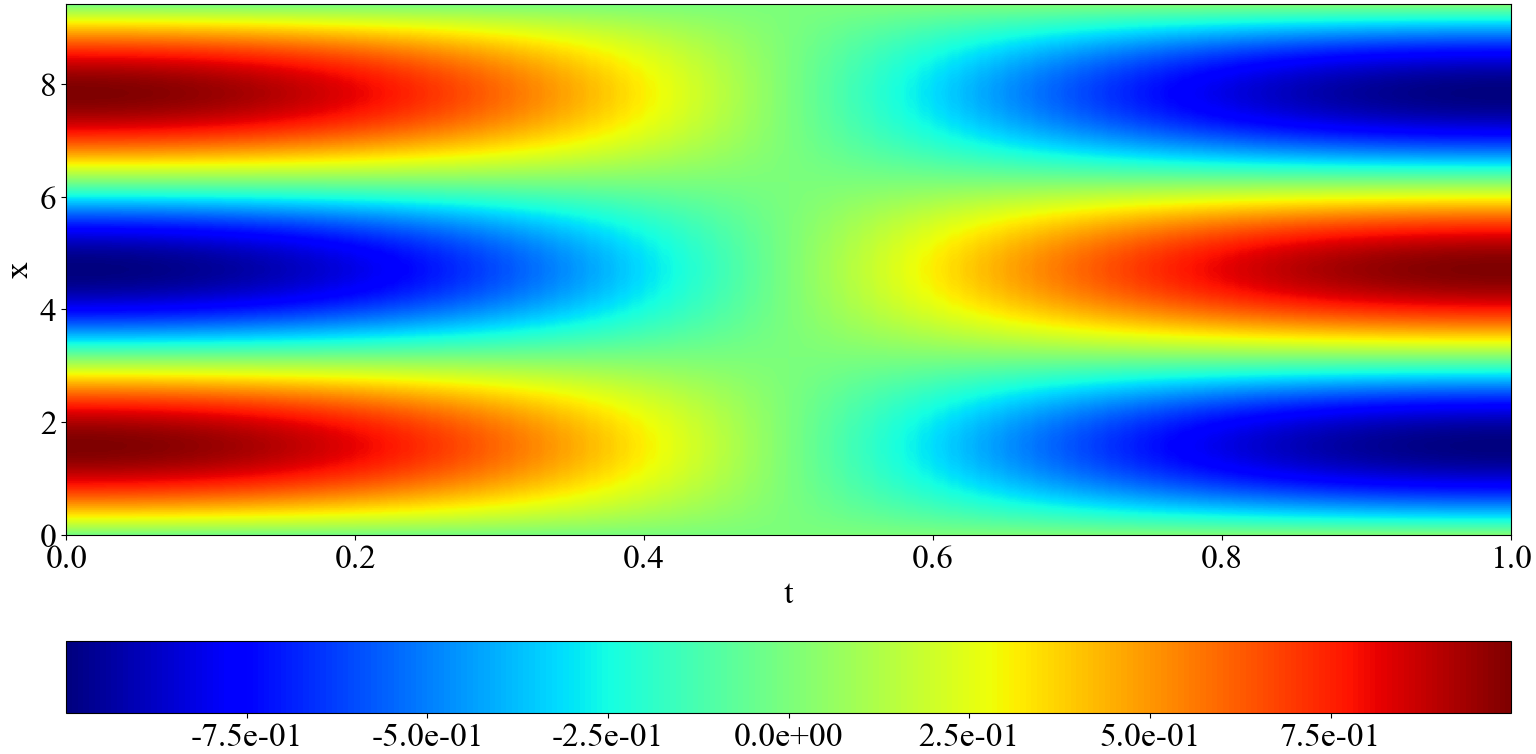}
    \caption{A-PINN (Proposed model)}
  \end{subfigure}

  \vspace{1.0em} 

  {\large Baseline solutions}\par\vspace{0.4em}

  \begin{subfigure}{0.45\textwidth}
    \centering
    \includegraphics[width=\linewidth]{ 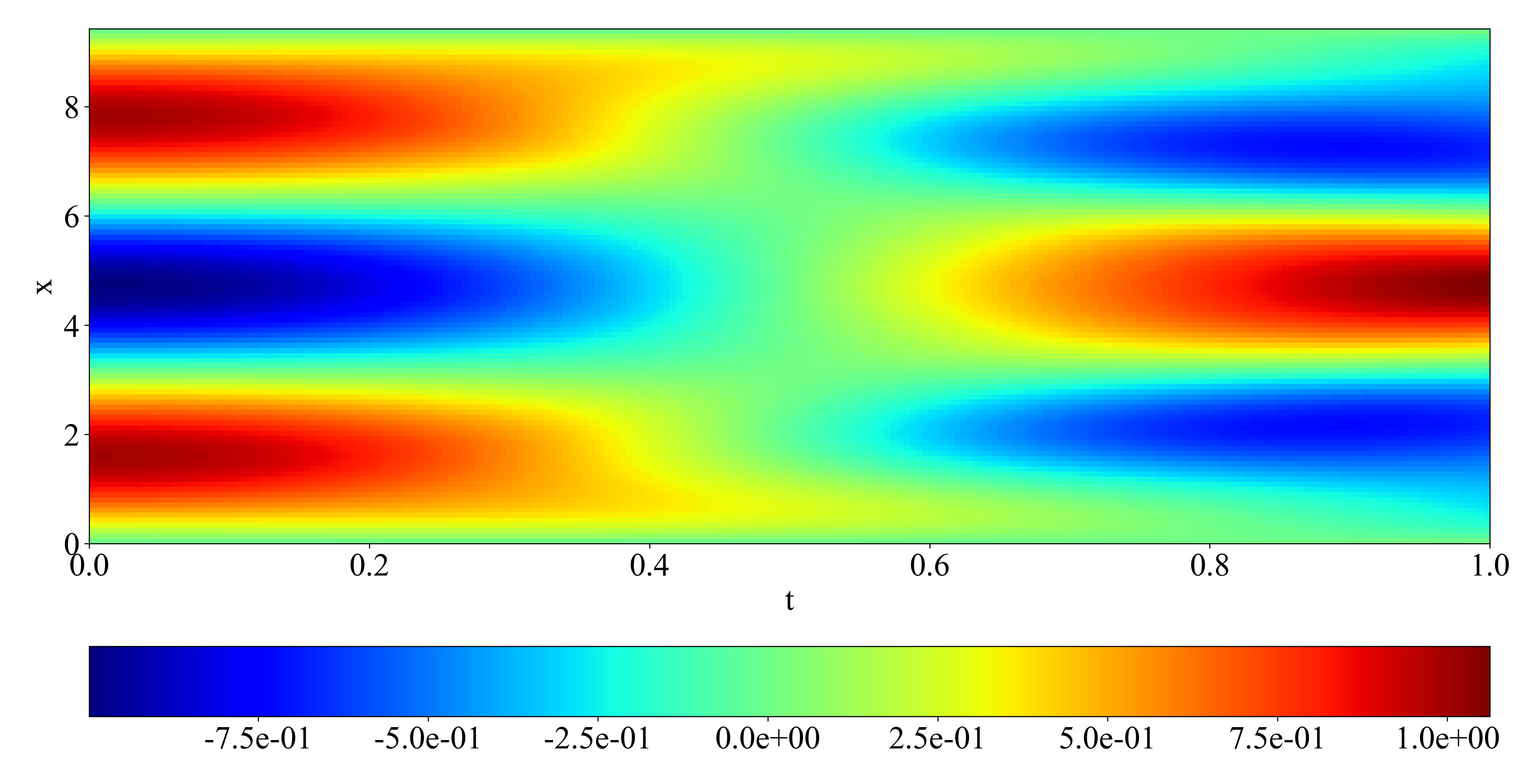}
    \caption{FDM}
  \end{subfigure}
  \hfill
  \begin{subfigure}{0.45\textwidth}
    \centering
    \includegraphics[width=\linewidth]{ 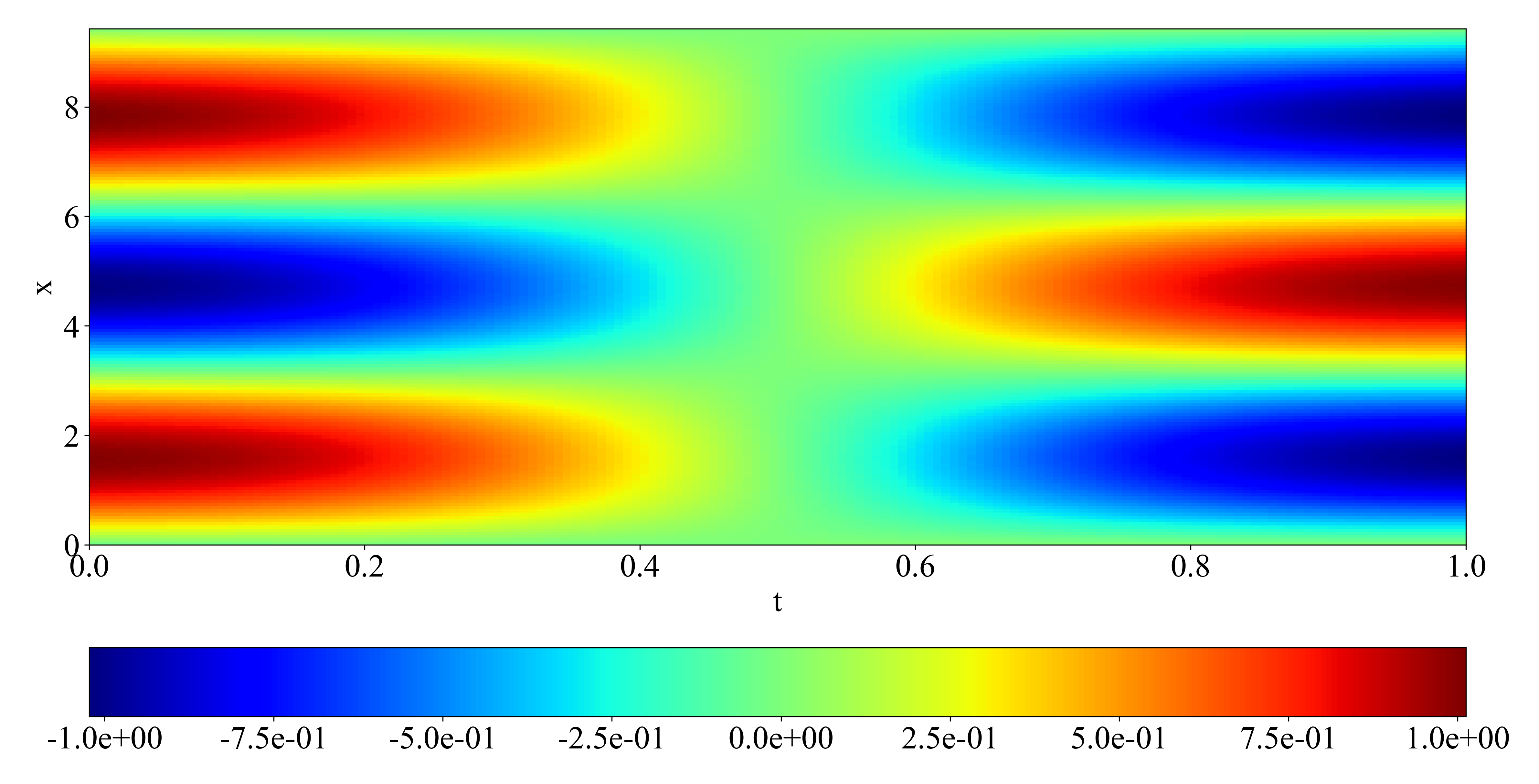}
    \caption{SANN}
  \end{subfigure}
   \vspace{0.8em}
  \begin{subfigure}{0.48\textwidth}
    \centering
    \includegraphics[width=\linewidth]{ 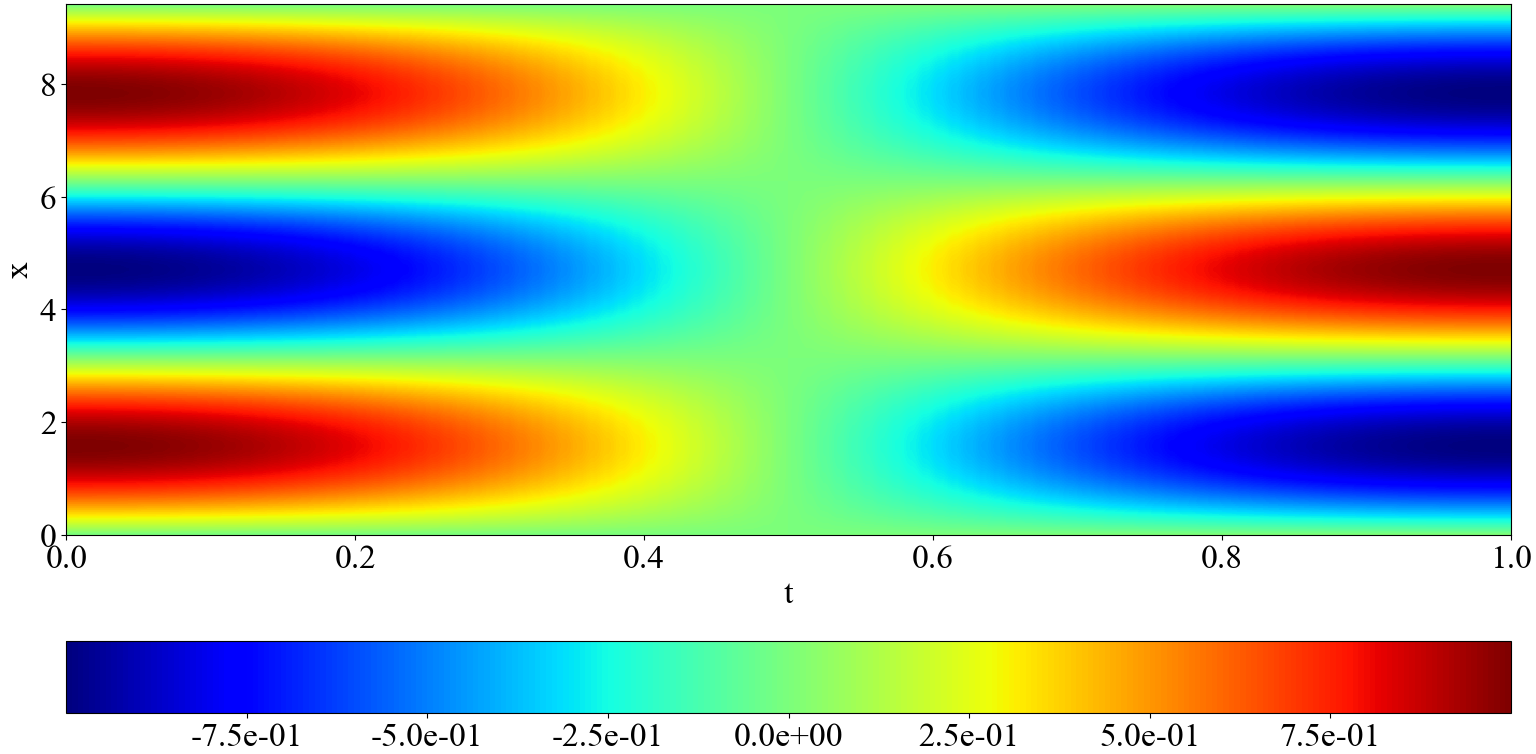}
    \caption{GT}
  \end{subfigure}

  \vspace{0.8em}
 \caption{2D comparison of physics-informed (top) and baseline (bottom) solutions for P3.}
  \label{P3_comparison_2d_blocks}
\end{figure}

\begin{figure}[H]
  \centering
  \captionsetup[sub]{justification=centering}
  {\large Physics-informed solutions}\par\vspace{0.4em}

\begin{subfigure}{0.48\textwidth}
    \centering
    \includegraphics[width=\linewidth]{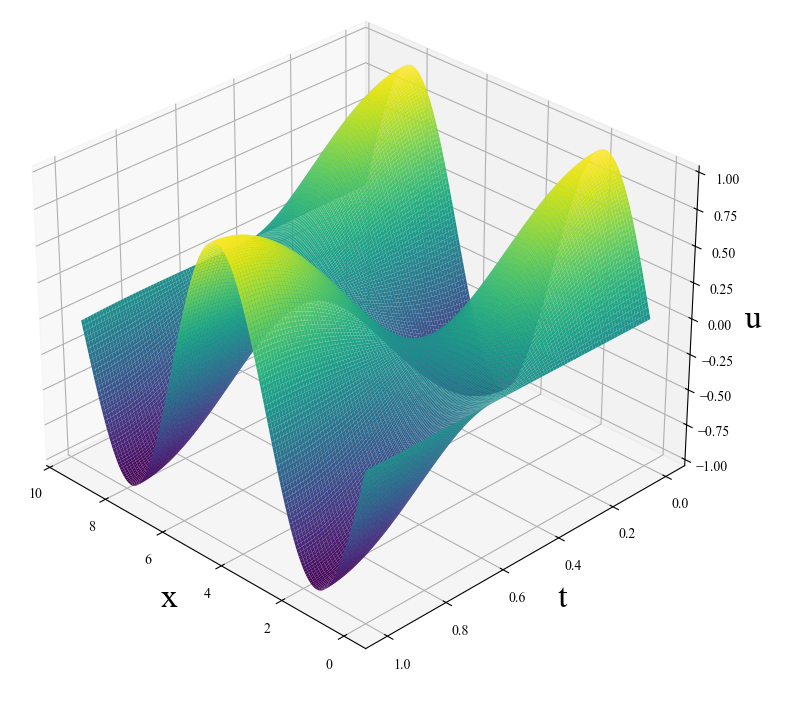}
    \caption{PINN}
  \end{subfigure}
  \hfill
  \begin{subfigure}{0.48\textwidth}
    \centering
    \includegraphics[width=\linewidth]{ 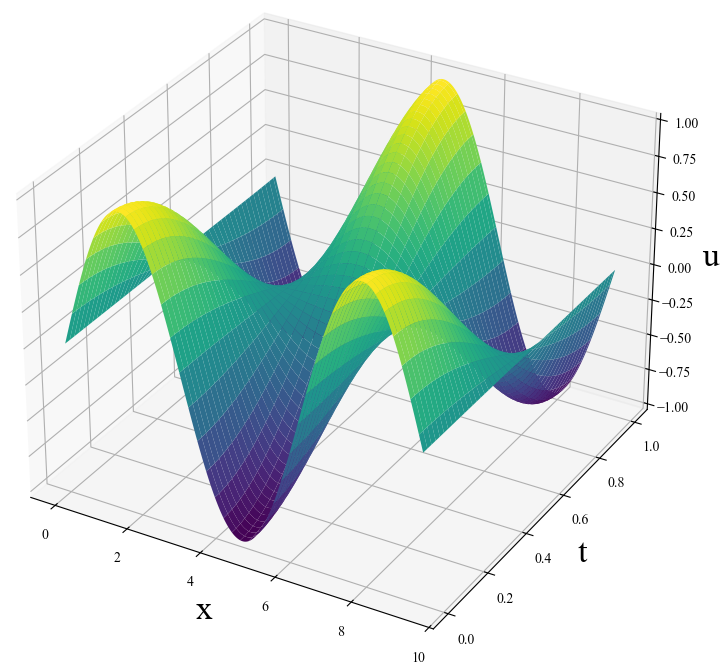}
    \caption{A-PINN (Proposed model)}
  \end{subfigure}

  \vspace{1.0em} 

{\large Baseline solutions}\par\vspace{0.4em}

\begin{subfigure}{0.45\textwidth}
  \centering
  \includegraphics[width=\linewidth]{ 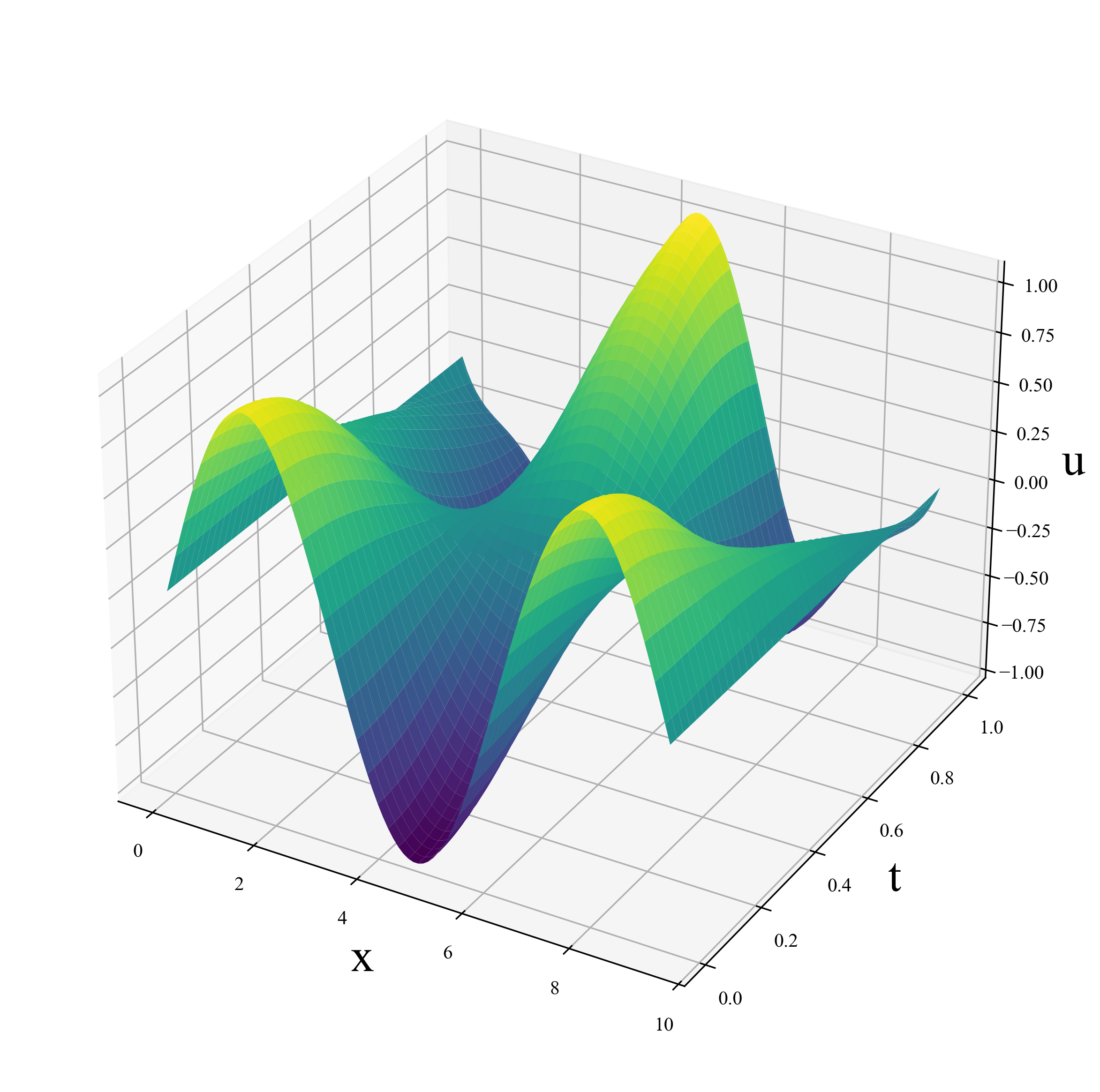}
  \caption{FDM}
\end{subfigure}
\hfill
\begin{subfigure}{0.45\textwidth}
  \centering
  \includegraphics[width=\linewidth]{ 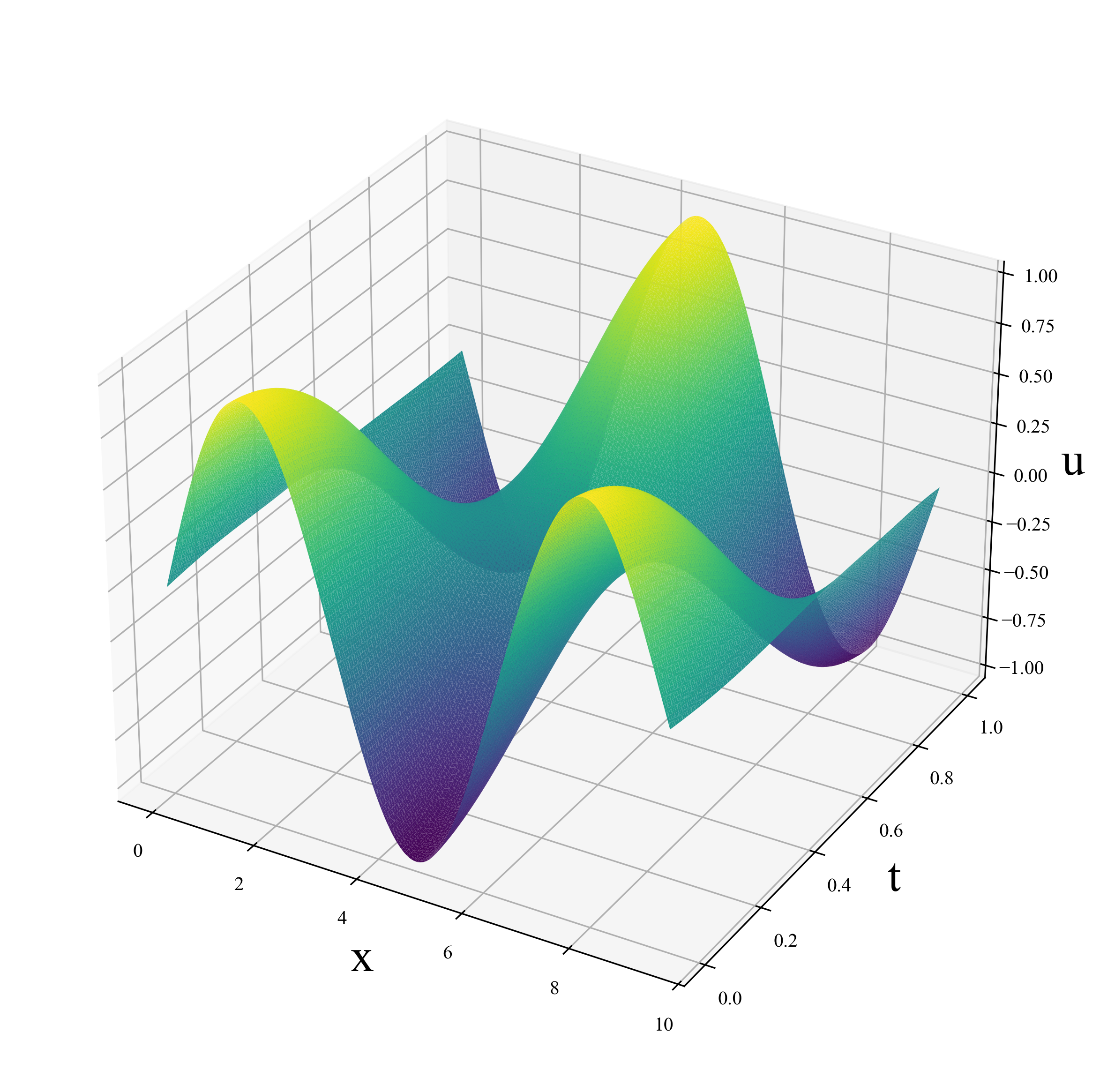}
  \caption{SANN}
\end{subfigure}
\vspace{0.8em} 
\begin{subfigure}{0.45\textwidth}
  \centering
  \includegraphics[width=\linewidth]{ 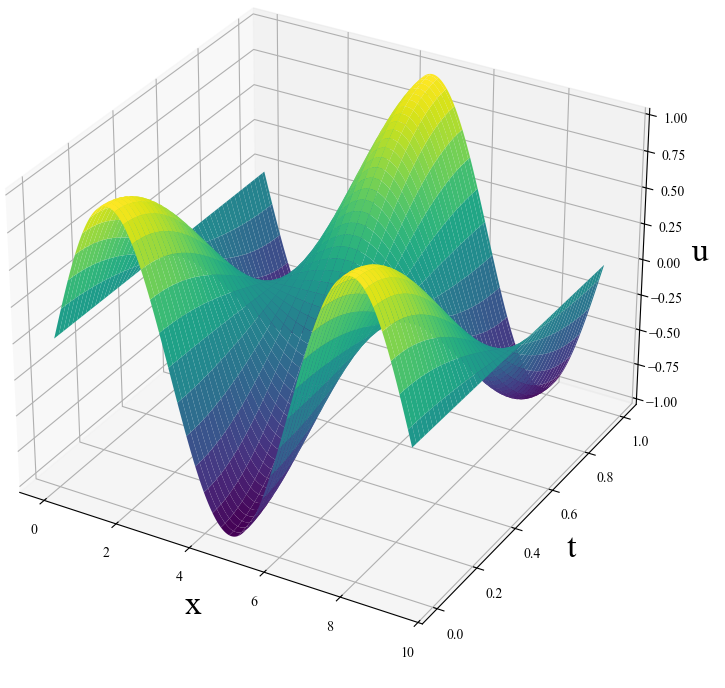}
  \caption{GT}
\end{subfigure}

\caption{3D comparison of physics-informed (top) and baseline (bottom) solutions for P3.}
\label{P3_comparison_3d_blocks}
\end{figure}

The beam is assumed to be simply supported at both ends, BCs 
\(u(0,t) = u(3\pi,t) = 0\) and \(u_{xx}(0,t) = u_{xx}(3\pi,t) = 0\). 
The initial displacement and velocity are prescribed as 
\(u(x,0) = \sin(x)\) and \(u_t(x,0) = 0\). To simplify the fourth-order term, an auxiliary variable 
\(v(x,t)\approx u_{xx}(x,t)\) is introduced.
As shown in Eq.~\eqref{A_Loss}, the total A-PINN loss is obtained as:

\begin{align} \label{eq:42}
\mathcal{L}(\theta) 
&= \tilde{w}_f \left( \frac{1}{N_f} \sum_{i=1}^{N_f} 
\Big| \widehat{u}_{tt}(x_i^f,t_i^f;\theta) + \widehat{v}_{xx}(x_i^f,t_i^f;\theta) + \widehat{u}(x_i^f,t_i^f;\theta) - f(x_i^f,t_i^f) \Big|^2 \right) \notag \\
&\quad + \tilde{w}_\text{a} \left( \frac{1}{N_\text{a}} \sum_{i=1}^{N_\text{a}} 
\Big| \widehat{v}(x_i^\text{a},t_i^\text{a};\theta) - \widehat{u}_{xx}(x_i^\text{a},t_i^\text{a};\theta) \Big|^2 \right) \notag \\
&\quad + \tilde{w}_0 \left( \frac{1}{N_0} \sum_{i=1}^{N_0} 
\Big( \big|\widehat{u}(x_i^0,0;\theta) - \sin(x_i^0)\big|^2 + \big|\widehat{u}_t(x_i^0,0;\theta)\big|^2 \Big) \right) \notag \\
&\quad + \tilde{w}_b \left( \frac{1}{N_b} \sum_{i=1}^{N_b} 
\Big( \big|\widehat{u}(0,t_i^b;\theta)\big|^2 + \big|\widehat{u}(3\pi,t_i^b;\theta)\big|^2 
+ \big|\widehat{v}(0,t_i^b;\theta)\big|^2 + \big|\widehat{v}(3\pi,t_i^b;\theta)\big|^2 \Big) \right).
\end{align}
Here, in Eq. \ref{eq:42} \( \widehat{u}(x,t;\theta) \) and \( \widehat{v}(x,t;\theta) \) denote the NN outputs corresponding to the displacement field and its auxiliary variable, respectively.
In this simulation, the training dataset is designed to include $N_0 = 500$ initial points, $N_b = 500$ boundary points at $x=0$, and $\,3\pi$, $N_f = 1000$ collocation points distributed within the spatio-temporal domain, and $N_\text{a} = 500$ auxiliary points to enforce the relation $v \approx u_{xx}$. Together, these subsets ensure that ICs, BCs, PDE residuals, and auxiliary constraints are simultaneously satisfied during optimization. The EBB is considered under undamped, simply supported conditions, with the exact solution \cite{kapoor2024transfer} serving as the GT. Training is carried out using the L-BFGS optimizer, chosen for its stability in physics-informed learning. (cf.: Figures~\ref{P3_comparison_2d_blocks} and~\ref{P3_comparison_3d_blocks}) demonstrate that the A-PINN matches the GT much more closely than the PINN and classical baselines FDM and SANN, especially in reproducing the temporal evolution across space with high fidelity.

\begin{figure}[H]
    \centering
    
    \begin{subfigure}{0.45\textwidth}
        \centering
        \includegraphics[width=\linewidth]{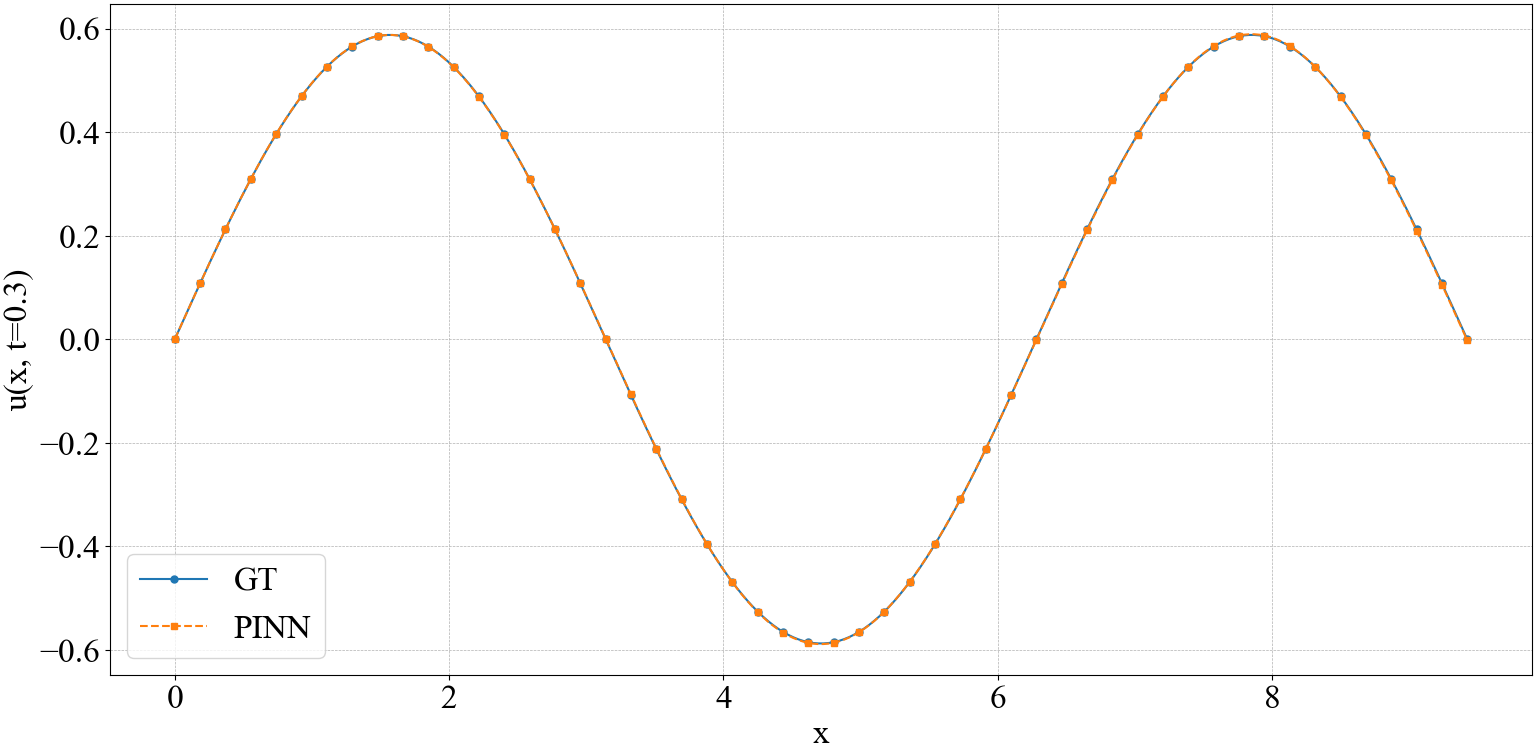}
        \caption{t=0.3 using PINN}
    \end{subfigure}
    \hfill
    \begin{subfigure}{0.45\textwidth}
        \centering
        \includegraphics[width=\linewidth]{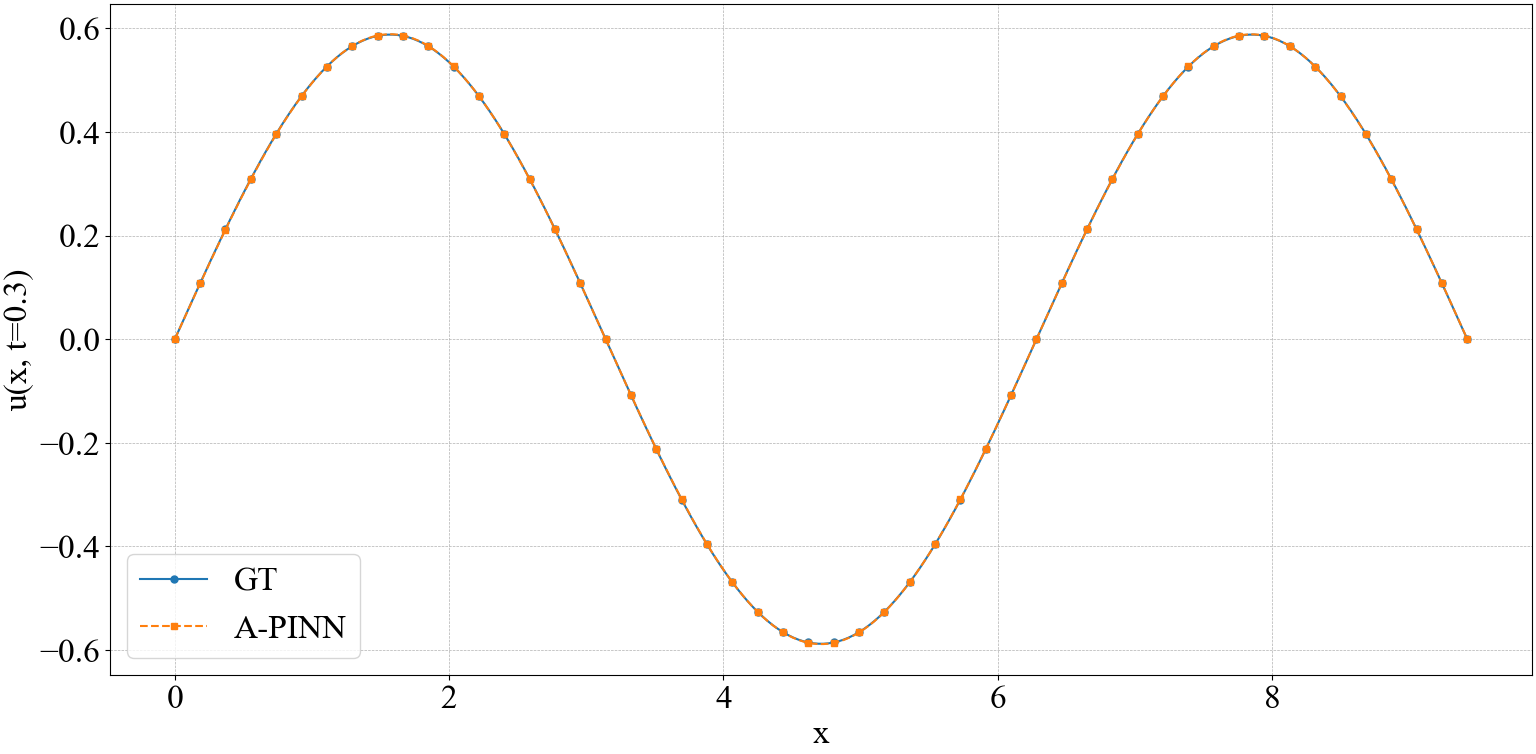}
        \caption{t=0.3 using A-PINN}
    \end{subfigure}

    \vskip\baselineskip 

    \begin{subfigure}{0.45\textwidth}
        \centering
        \includegraphics[width=\linewidth]{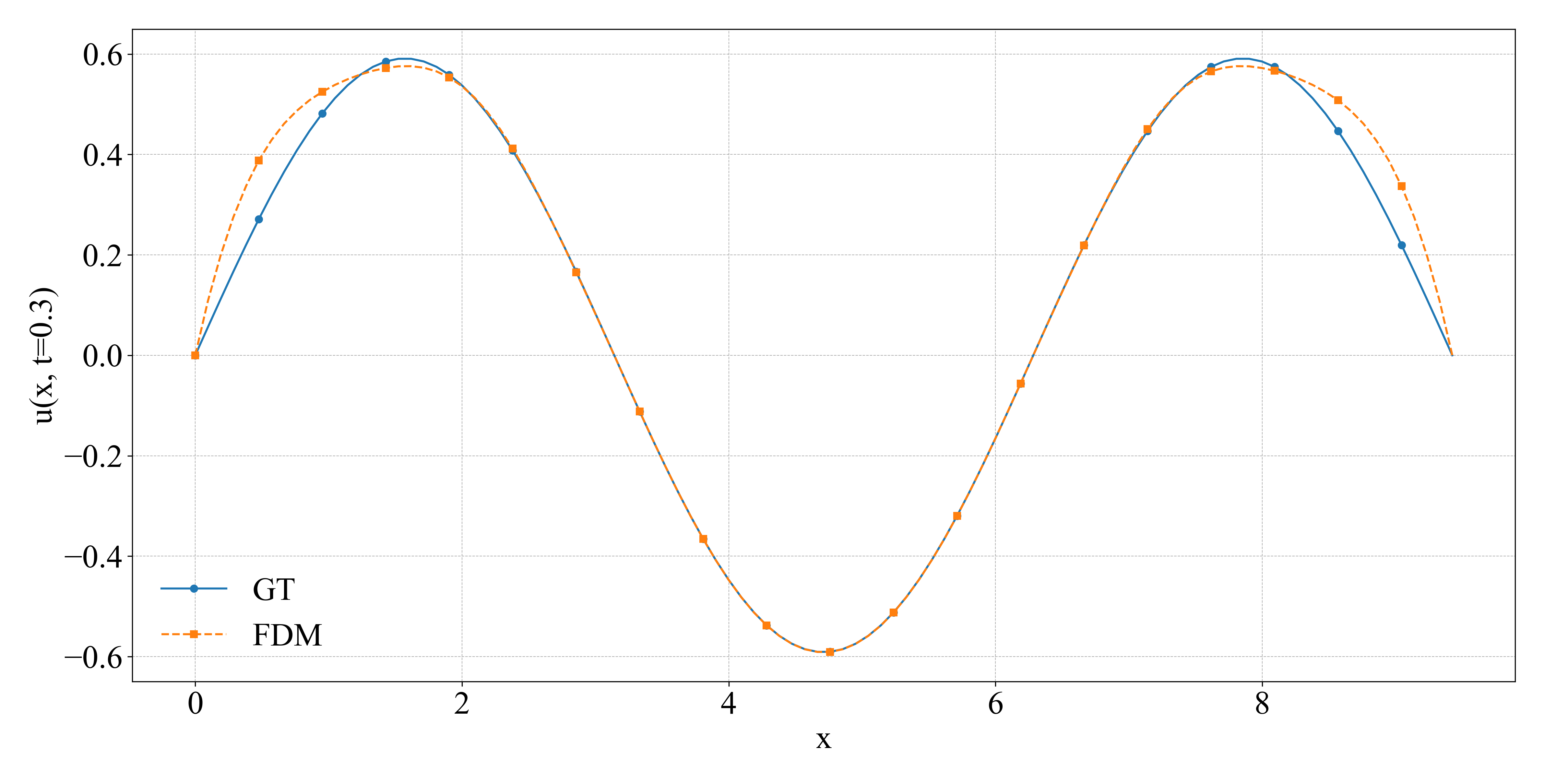}
        \caption{t=0.3 using FDM}
    \end{subfigure}
    \hfill
    \begin{subfigure}{0.45\textwidth}
        \centering
        \includegraphics[width=\linewidth]{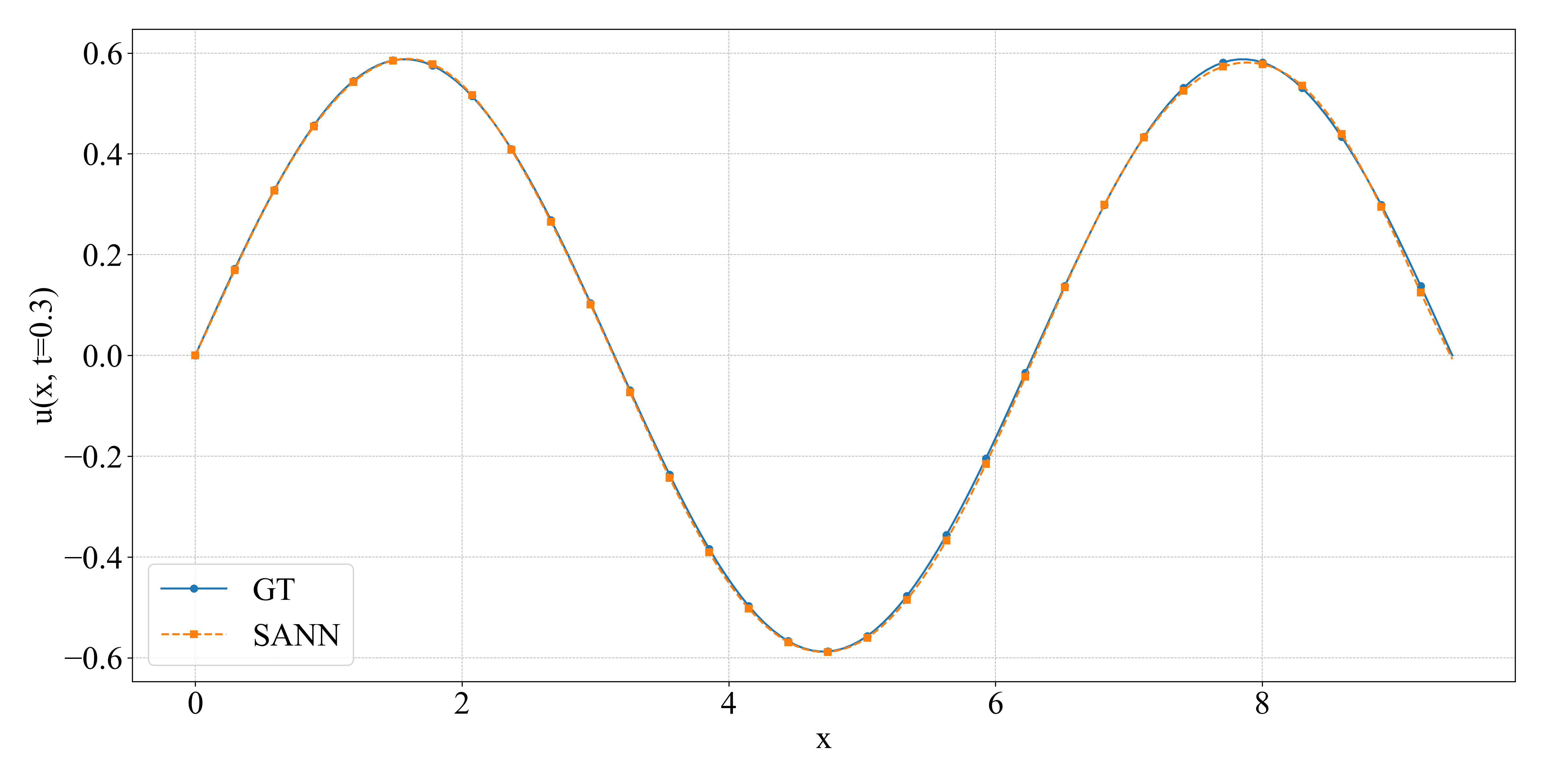}
        \caption{t=0.3 using SANN}
    \end{subfigure}

    \caption{Comparison of PINN, A-PINN, FDM, and SANN solutions with GT at $t=0.3$ for P3.}
    \label{2D_slice_t=0.3_P3}
\end{figure}

\begin{figure}[H]
    \centering
    
    \begin{subfigure}{0.45\textwidth}
        \centering
        \includegraphics[width=\linewidth]{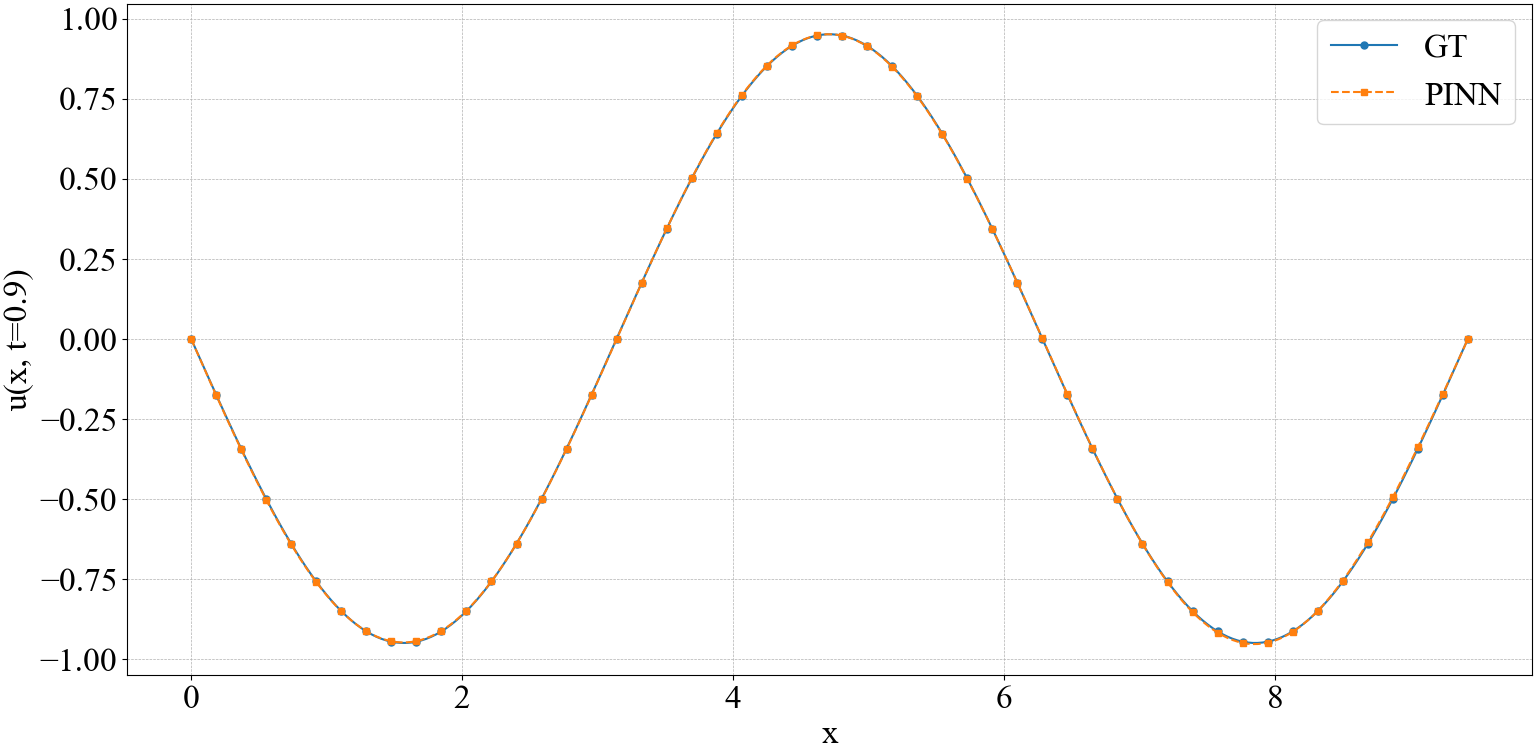}
        \caption{t=0.9 using PINN}
    \end{subfigure}
    \hfill
    \begin{subfigure}{0.45\textwidth}
        \centering
        \includegraphics[width=\linewidth]{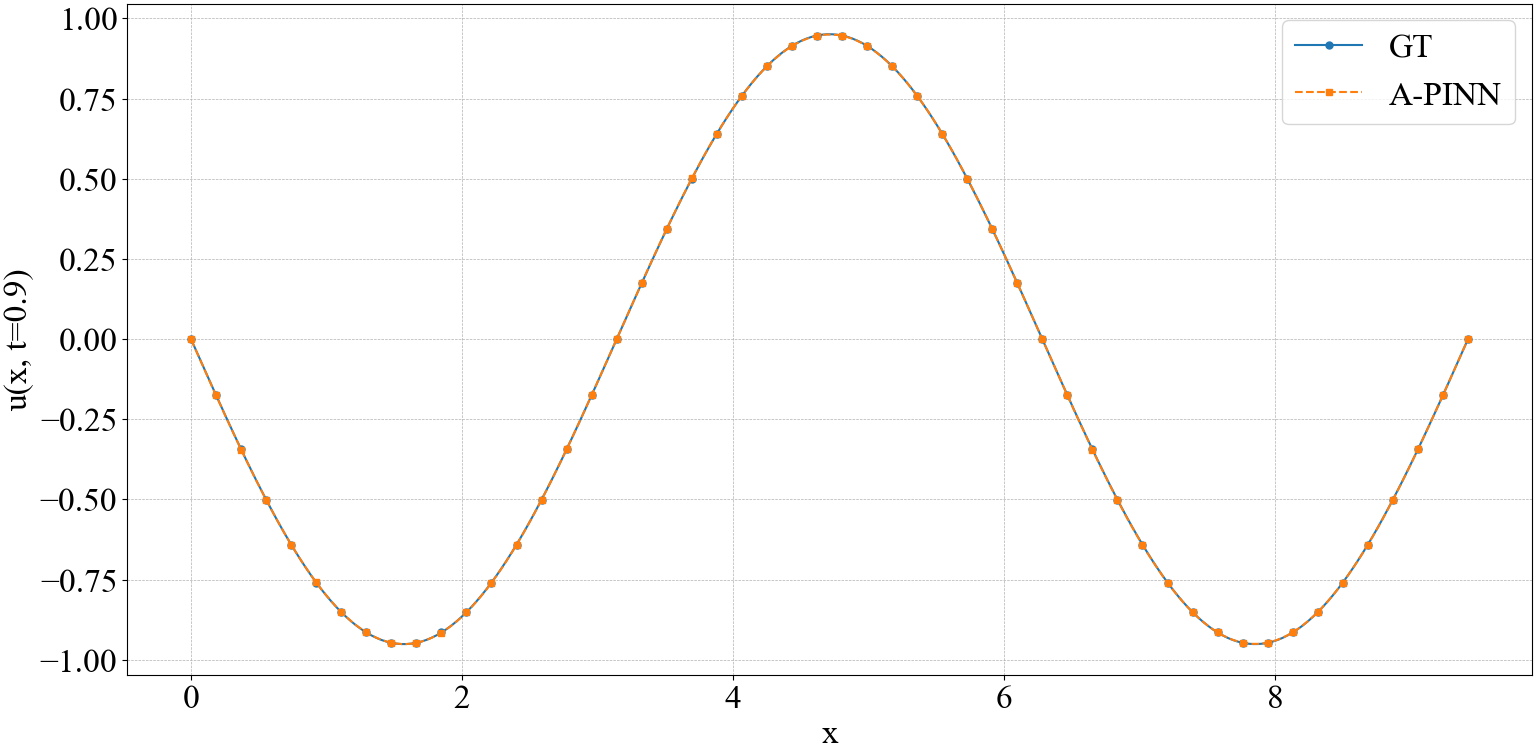}
        \caption{t=0.9 using A-PINN}
    \end{subfigure}

    \vskip\baselineskip 

    \begin{subfigure}{0.45\textwidth}
        \centering
        \includegraphics[width=\linewidth]{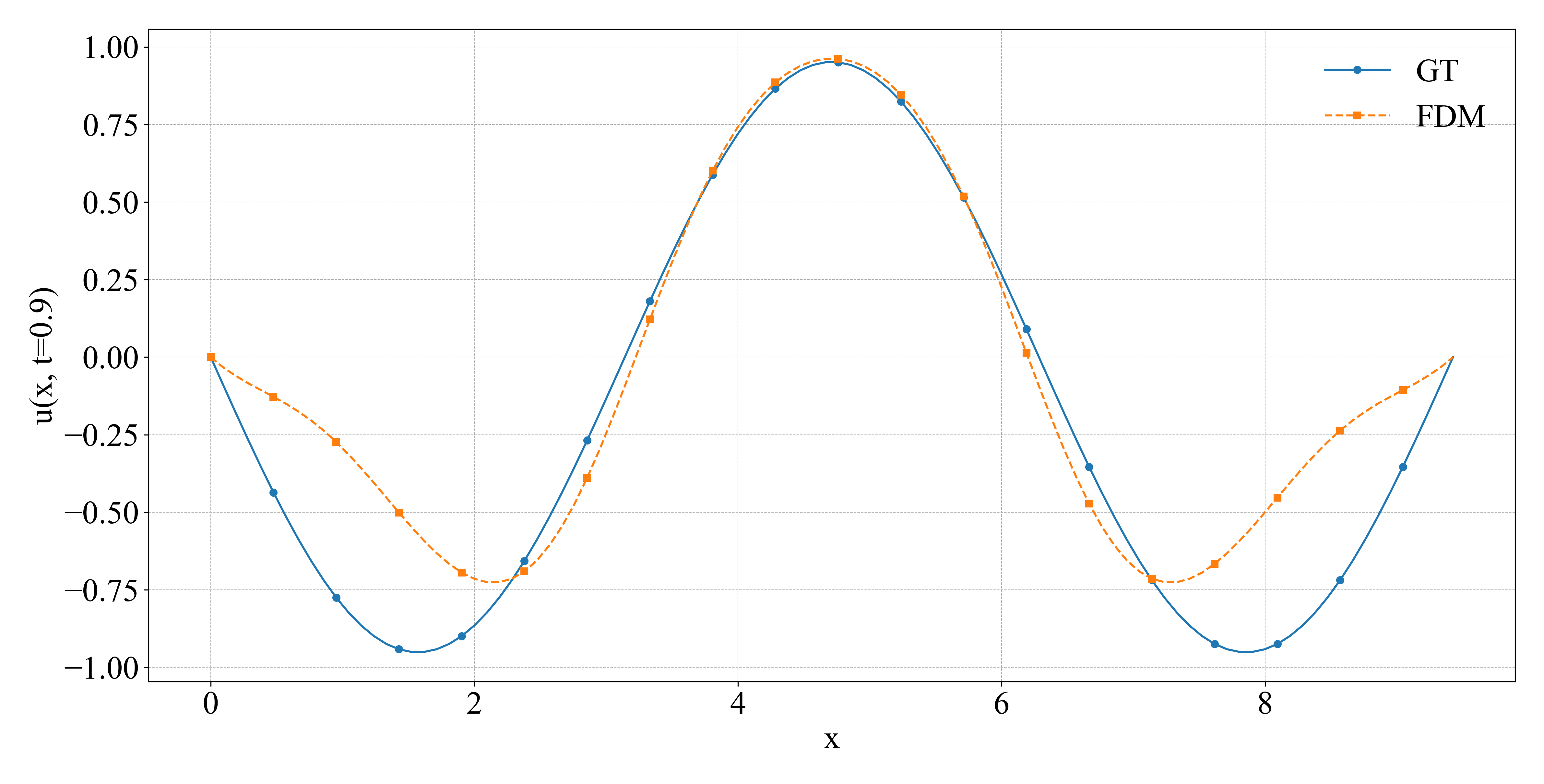}
        \caption{t=0.9 using FDM}
    \end{subfigure}
    \hfill
    \begin{subfigure}{0.45\textwidth}
        \centering
        \includegraphics[width=\linewidth]{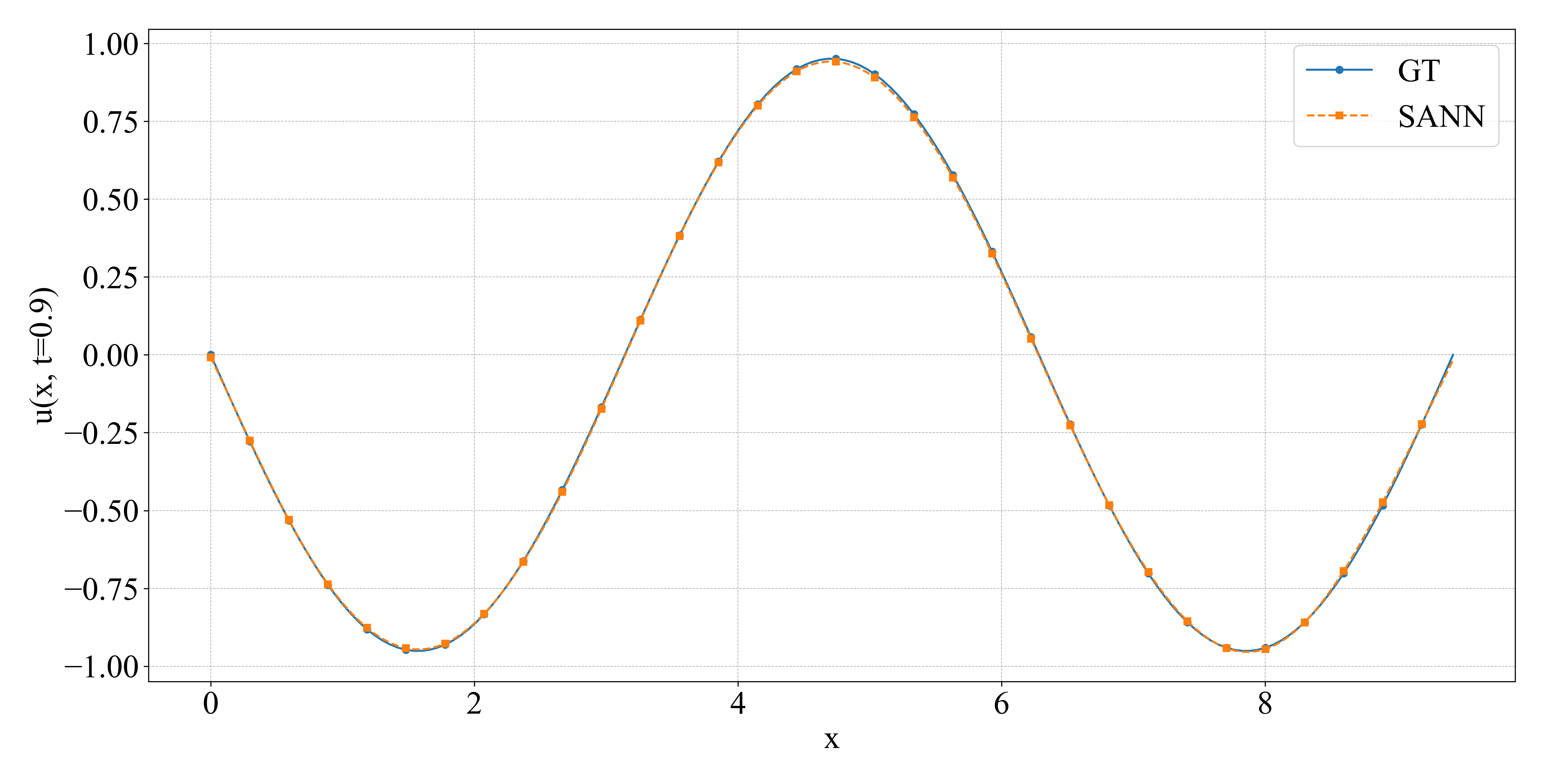}
        \caption{t=0.9 using SANN}
    \end{subfigure}

    \caption{Comparison of PINN, A-PINN, FDM, and SANN solutions with GT at $t=0.9$ for P3.}
    \label{2D_slice_t=0.9_P3}
\end{figure}

   Representative comparisons are presented at $t=0.3$ and $t=0.9$ 
(cf.: Figures~\ref{2D_slice_t=0.3_P3} and \ref{2D_slice_t=0.9_P3}, 
Tables~\ref{Table_t=0.3_(P3)} and \ref{Table_t=0.9_(P3)}). 
At $t=0.5$, the beam exhibits a reduced deflection amplitude, whereas at $t=0.9$ 
The deformation pattern is reversed relative to the initial state, reflecting 
the opposite oscillation phase. In both situations, the A-PINN solution 
remains nearly to the GT, while the PINN and baseline 
approaches, FDM and SANN, show clear deviations. The spatial dependence $\sin(x)$ is dictated by the simply supported BCs, and the temporal factor $\cos(\pi t)$ governs amplitude modulation and phase reversal. The Winkler foundation contributes an additional elastic restoring effect, which shifts the natural frequency and enhances the stability of the oscillations. Overall, the A-PINN delivers more accurate predictions than the PINN, FDM, and SANN.

  \begin{table}[ht]
    \centering
    \caption{Comparison among the proposed model and baselines at $t = 0.3$, with spatial domain $x \in [0,3\pi]$ and spatial step $\Delta x = 0.94$ for P3.}
    \label{Table_t=0.3_(P3)}
    \scriptsize
    \begin{tabular}{c ccc cc cc}
        \toprule
        $\textbf{x}$ & \multicolumn{3}{c}{Baselines Results} & \multicolumn{2}{c}{Physics-informed Results} & \multicolumn{2}{c}{$E_1$ w.r.t.\ GT} \\
        \cmidrule(lr){2-4} \cmidrule(lr){5-6} \cmidrule(lr){7-8}
            & GT & FDM & SANN & PINN & A-PINN & PINN & A-PINN \\
        \midrule
        0.00 & 0.000000 & 0.000000 &  0.000373 & -0.000020 & -0.000335 & 2.034e-05 & 3.345e-04 \\
        0.94 & 0.475528 & 0.522990 &  0.473433 &  0.475933 &  0.474592 & 4.044e-04 & 7.867e-05 \\
        1.88 & 0.559017 & 0.555856 &  0.562127 &  0.558488 &  0.560275 & 5.291e-04 & 3.651e-04 \\
        2.83 & 0.181636 & 0.181550 &  0.178362 &  0.181885 &  0.179670 & 2.498e-04 & 5.303e-04 \\
        3.77 &-0.345492 &-0.347335 & -0.352500 & -0.344927 & -0.344705 & 5.649e-04 & 8.288e-04 \\
        4.71 &-0.587785 &-0.590800 & -0.588651 & -0.589019 & -0.588463 & 1.233e-03 & 6.794e-04 \\
        5.65 &-0.345492 &-0.347335 & -0.356520 & -0.344911 & -0.347409 & 5.804e-04 & 3.930e-04 \\
        6.60 & 0.181636 & 0.181550 &  0.180111 &  0.179907 &  0.183495 & 1.728e-03 & 3.758e-04 \\
        7.54 & 0.559017 & 0.555856 &  0.551495 &  0.560003 &  0.558473 & 9.863e-04 & 5.763e-04 \\
        8.48 & 0.475528 & 0.522990 &  0.482275 &  0.474883 &  0.476748 & 6.455e-04 & 4.265e-04 \\
        9.42 & 0.000000 & 0.000000 & -0.007299 & -0.002152 &  0.002860 & 2.152e-03 & 5.171e-05 \\
        \bottomrule
    \end{tabular}
\end{table}

  \begin{table}[ht]
    \centering
    \caption{Comparison among the proposed model and baselines at $t = 0.9$, with spatial domain $x \in [0,3\pi]$ and spatial step $\Delta x = 0.94$ for P3.}
    \label{Table_t=0.9_(P3)}
    \scriptsize
    \begin{tabular}{c ccc cc cc}
        \toprule
        $\textbf{x}$ & \multicolumn{3}{c}{Baselines Results} & \multicolumn{2}{c}{Physics-informed Results} & \multicolumn{2}{c}{$E_1$ w.r.t.\ GT} \\
        \cmidrule(lr){2-4} \cmidrule(lr){5-6} \cmidrule(lr){7-8}
            & GT & FDM & SANN & PINN & A-PINN & PINN & A-PINN \\
        \midrule
        0.00 &  0.000000 &  0.000000 & -0.007751 & -0.000434 & -0.000378 & 4.340e-04 & 3.780e-04 \\
        0.94 & -0.769421 & -0.269922 & -0.766781 & -0.770993 & -0.767894 & 1.572e-03 &  1.391e-04 \\
        1.88 & -0.904508 & -0.689098 & -0.901414 & -0.902686 & -0.905780 & 1.822e-03 & 1.741e-04 \\
        2.83 & -0.293893 & -0.414003 & -0.299753 & -0.295005 & -0.290394 & 1.112e-03 & 1.175e-03 \\
        3.77 &  0.559017 &  0.567619 &  0.556569 &  0.560535 &  0.557912 & 1.518e-03 & 1.173e-03 \\
        4.71 &  0.951057 &  0.961611 &  0.941888 &  0.951099 &  0.950872 & 4.272e-05 & 1.813e-04 \\
        5.65 &  0.559017 &  0.567619 &  0.551058 &  0.557657 &  0.563826 & 1.360e-03 & 1.070e-03 \\
        6.60 & -0.293893 & -0.414003 & -0.295819 & -0.292742 & -0.296697 & 1.150e-03 & 4.040e-04 \\
        7.54 & -0.904508 & -0.689098 & -0.902940 & -0.908210 & -0.903463 & 3.701e-03 & 1.097e-03 \\
        8.48 & -0.769421 & -0.269922 & -0.765265 & -0.766701 & -0.772193 & 2.719e-03 & 1.488e-03 \\
        9.42 &  0.000000 &  0.000000 & -0.017068 & -0.000805 &  0.004474 & 8.048e-04 & 7.012e-05  \\
        \bottomrule
    \end{tabular}
\end{table}

\section{Discussion}\label{sec6}
\justifying
A comprehensive discussion of the obtained numerical and graphical results is crucial to interpret the physical relevance and computational behavior of the A-PINN framework. The key objective of this section is to analyze the performance of the A-PINN model in solving EBB problems and to validate its superiority over traditional numerical and NN approaches. The conversation focuses on how the incorporation of auxiliary variables and adaptive optimization techniques improves accuracy and convergence in all vibration settings. The results obtained from the A-PINN model are evaluated against the PINN, SANN an established NN approach, FDM a classical numerical scheme, and the GT. The comparative analysis highlights the superior accuracy and convergence rate of the A-PINN across all three benchmark problems, along with its percentage improvements over the physics-informed model and the baseline SANN and FDM methods, as summarized in (cf.: Tables~\ref{all_comparison_errors} and~\ref{loss_epochs}).
One may observe that the mean of $E_1$ values lies around $10^{-4}$ for our experiments. Furthermore, the training loss remains in the range of $10^{-5}$ to $10^{-4}$ for each problem, as reported in Table~\ref{loss_epochs}. 

\begin{table*}[h]
\centering
\caption{Quantitative comparison of physics-informed and baseline models across three benchmark vibration problems using error metrics $E_2$, $E_3$, and $E_4$, along with percentage improvement of A-PINN.}
\label{all_comparison_errors}
\scriptsize
\renewcommand{\arraystretch}{1.35}
\resizebox{\textwidth}{!}{%
\begin{tabular}{llccccc}
\toprule
\textbf{Problem} & \textbf{Model Type} & \textbf{Method} & \textbf{$E_2$} & \textbf{$E_3$} & \textbf{$E_4$} & \textbf{Improvement in (\%)} \\
\midrule

\multirow{4}{*}{\textbf{P1}} 
& \multirow{2}{*}{Physics-informed} 
& A-PINN & \textbf{7.8306$\times$10$^{-7}$} & \textbf{1.7330$\times$10$^{-3}$} & \textbf{7.9130$\times$10$^{-3}$} & -- \\
&  & PINN  & 1.4670$\times$10$^{-5}$ & 7.5015$\times$10$^{-3}$ & 3.3967$\times$10$^{-2}$ & \textbf{95} \\
\cmidrule(lr){2-7}\addlinespace[2pt]
& \multirow{2}{*}{Baselines} 
& SANN  & 2.9434$\times$10$^{-4}$ & 3.3600$\times$10$^{-2}$ & 3.6730$\times$10$^{-2}$ & \textbf{99} \\
&  & FDM   & 1.9390$\times$10$^{-4}$ & 2.8016$\times$10$^{-2}$ & 4.2482$\times$10$^{-2}$ & \textbf{99} \\
\midrule

\multirow{4}{*}{\textbf{P2}} 
& \multirow{2}{*}{Physics-informed} 
& A-PINN & \textbf{1.0062$\times$10$^{-6}$} & \textbf{2.0001$\times$10$^{-3}$} & \textbf{3.6902$\times$10$^{-3}$} & -- \\
&  & PINN  & 1.6856$\times$10$^{-6}$ & 2.5888$\times$10$^{-3}$ & 3.1474$\times$10$^{-3}$ & \textbf{40} \\
\cmidrule(lr){2-7}\addlinespace[2pt]
& \multirow{2}{*}{Baselines} 
& SANN  & 1.5504$\times$10$^{-5}$ & 7.8155$\times$10$^{-3}$ & 1.7886$\times$10$^{-2}$ & \textbf{94} \\
&  & FDM   & 1.3216$\times$10$^{-3}$ & 7.3516$\times$10$^{-2}$ & 1.2633$\times$10$^{-1}$ & \textbf{99} \\
\midrule

\multirow{4}{*}{\shortstack{\textbf{P3}}} 
& \multirow{2}{*}{Physics-informed} 
& A-PINN & \textbf{1.9486$\times$10$^{-7}$} & \textbf{8.8019$\times$10$^{-4}$} & \textbf{1.9091$\times$10$^{-3}$} & -- \\
&  & PINN  & 2.7020$\times$10$^{-6}$ & 3.2776$\times$10$^{-3}$ & 1.1415$\times$10$^{-2}$ & \textbf{93} \\
\cmidrule(lr){2-7}\addlinespace[2pt]
& \multirow{2}{*}{Baselines} 
& SANN  & 4.3340$\times$10$^{-5}$ & 1.3166$\times$10$^{-2}$ & 3.7576$\times$10$^{-2}$ & \textbf{99} \\
&  & FDM   & 2.2444$\times$10$^{-2}$ & 3.0046$\times$10$^{-1}$ & 5.2873$\times$10$^{-1}$ & \textbf{99} \\
\bottomrule
\end{tabular}}
\end{table*}


For P1, the A-PINN exhibits outstanding precision where the error magnitudes are of the order of $10^{-7}$ for $E_2$, $10^{-3}$ for $E_3$, and $10^{-3}$ for $E_4$. The final training loss remains of the order of $10^{-5}$ after 20{,}000 epochs, whereas the PINN reaches a similar level only after nearly twice as many iterations. Such rapid and steady convergence confirms the effectiveness of the adaptive learning process. This behavior is with the theoretical understanding that the auxiliary variables enhance the network’s capacity to represent higher-order spatial derivatives and to ensure physically consistent displacement.  

In P2, the A-PINN  maintains the lower error magnitudes of the order of $10^{-6}$, $10^{-3}$, and $10^{-3}$ for $E_2$, $E_3$, and $E_4$. We get the faster convergence in the 10{,}000 epochs compared to 1.5 times for the PINN. The adaptive loss balancing enables stable training and accurate amplitude-frequency responses consistent with the governing physics.

 For the P3, where numerical approaches frequently encounter coupling behavior and stiffness-induced complexity, the improvement becomes much more substantial. These consistent and substantial enhancements confirm that the incorporation of auxiliary variables and the adaptive Adam to L-BFGS optimization strategy collectively improve the convergence and accuracy across all three EBB problems. In this scenario, the A-PINN yields error magnitudes of the order of $10^{-7}$, $10^{-4}$, and $10^{-3}$ for $E_2$, $E_3$, and $E_4$, respectively, demonstrating the robustness and efficiency of A-PINN.
The final loss value remains around $10^{-6}$ after 10{,}000 epochs, signifying a well-stabilized and efficient optimization process. This strong performance supports the theoretical premise that the auxiliary formulation decomposes the higher-order beam operator into simpler, lower-order components, facilitating smooth and stable learning even under strong stiffness effects. Consequently, the displacement field predicted by the A-PINN remains continuous and physically meaningful without oscillations near the supports or within the foundation region.  

\begin{figure}[H]
\centering
\captionsetup[subfigure]{justification=centering}

\begin{subfigure}[t]{0.48\textwidth}
  \centering
  \includegraphics[width=0.95\linewidth,height=4cm]{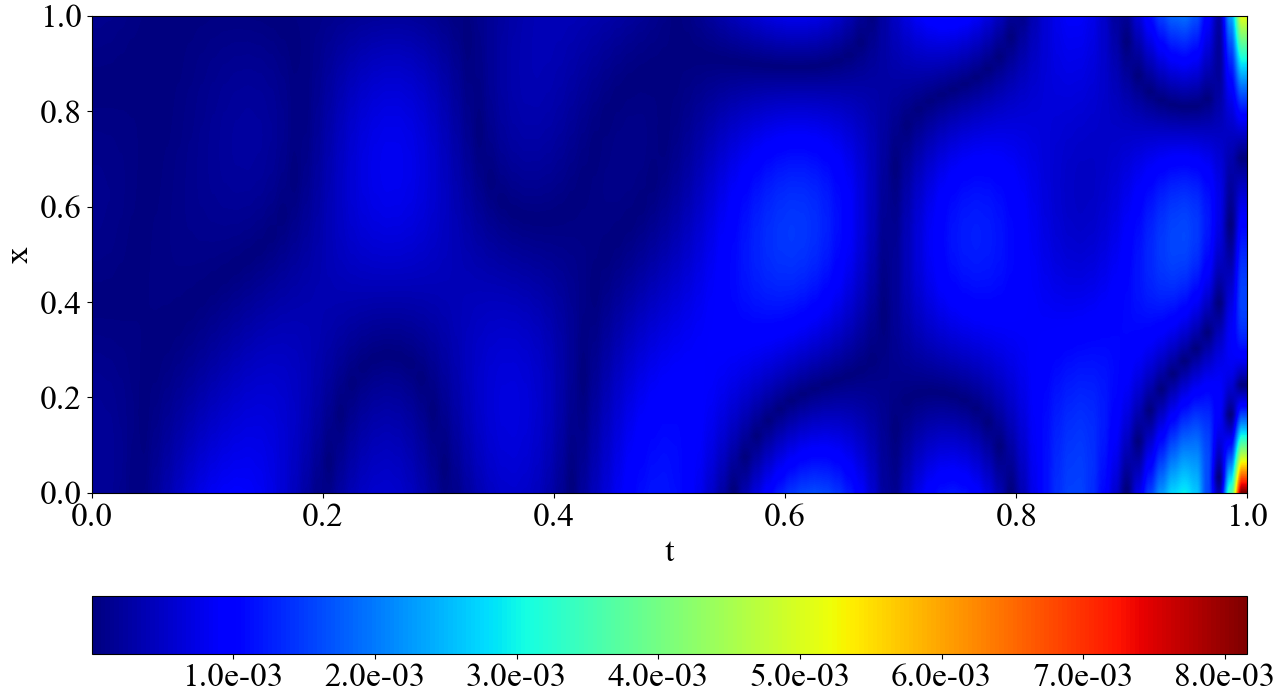}
  \caption{A-PINN}
\end{subfigure}\hfill
\begin{subfigure}[t]{0.48\textwidth}
  \centering
  \includegraphics[width=0.95\linewidth,height=4cm]{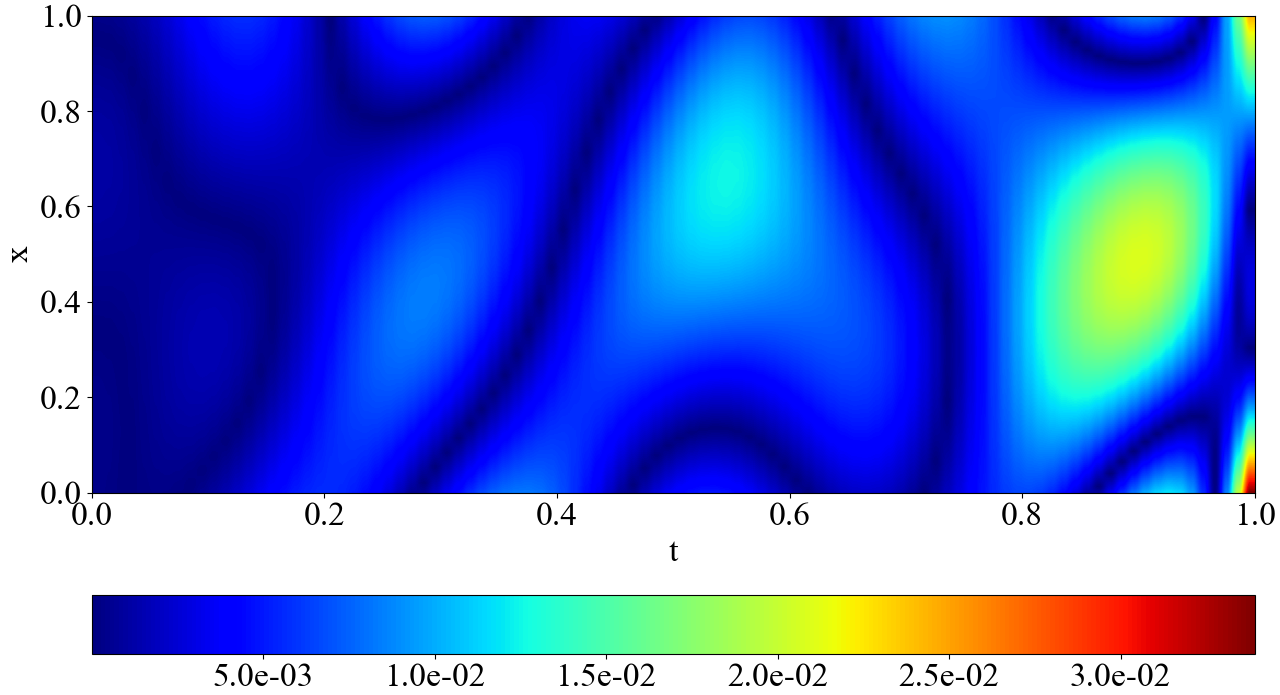}
  \caption{PINN}
\end{subfigure}
\\[0.5em]
\begin{subfigure}[t]{0.48\textwidth}
  \centering
  \includegraphics[width=0.95\linewidth,height=4cm]{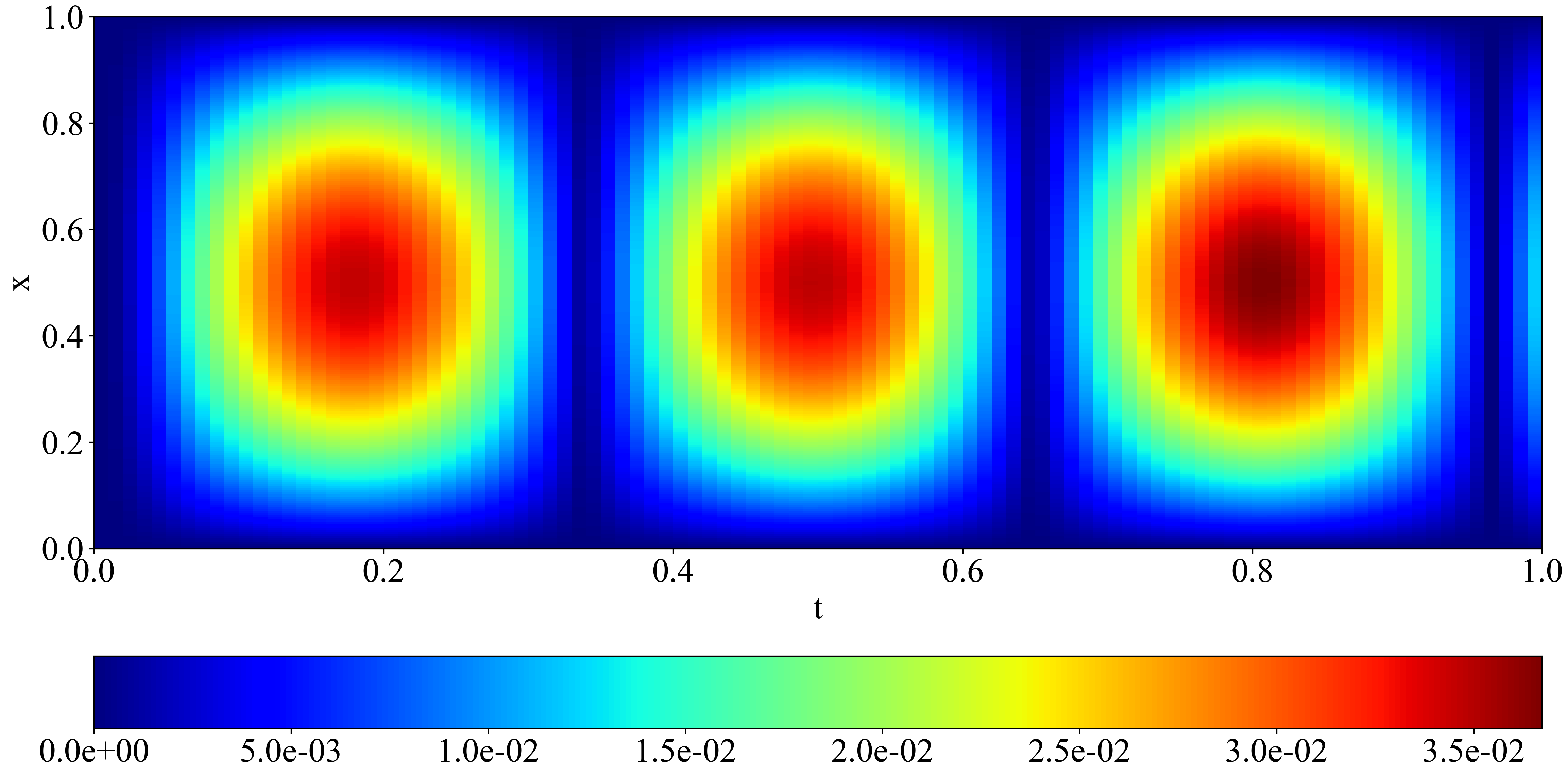}
  \caption{SANN}
\end{subfigure}\hfill
\begin{subfigure}[t]{0.48\textwidth}
  \centering
  \includegraphics[width=0.95\linewidth,height=4cm]{ 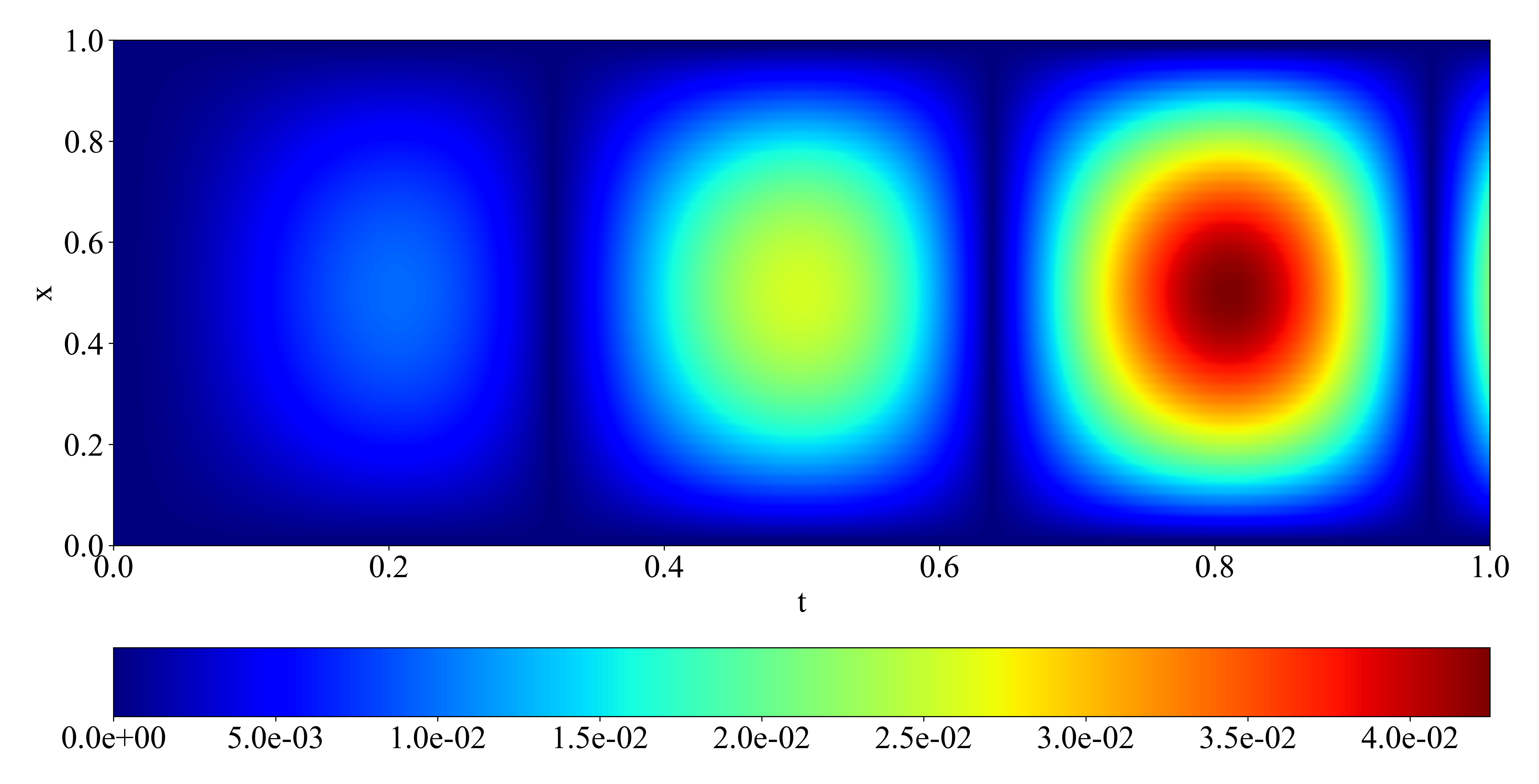}
  \caption{FDM}
\end{subfigure}
\\[1em]
\begin{subfigure}[t]{0.48\textwidth}
  \centering
  \includegraphics[width=0.95\linewidth,height=4.9cm]{ 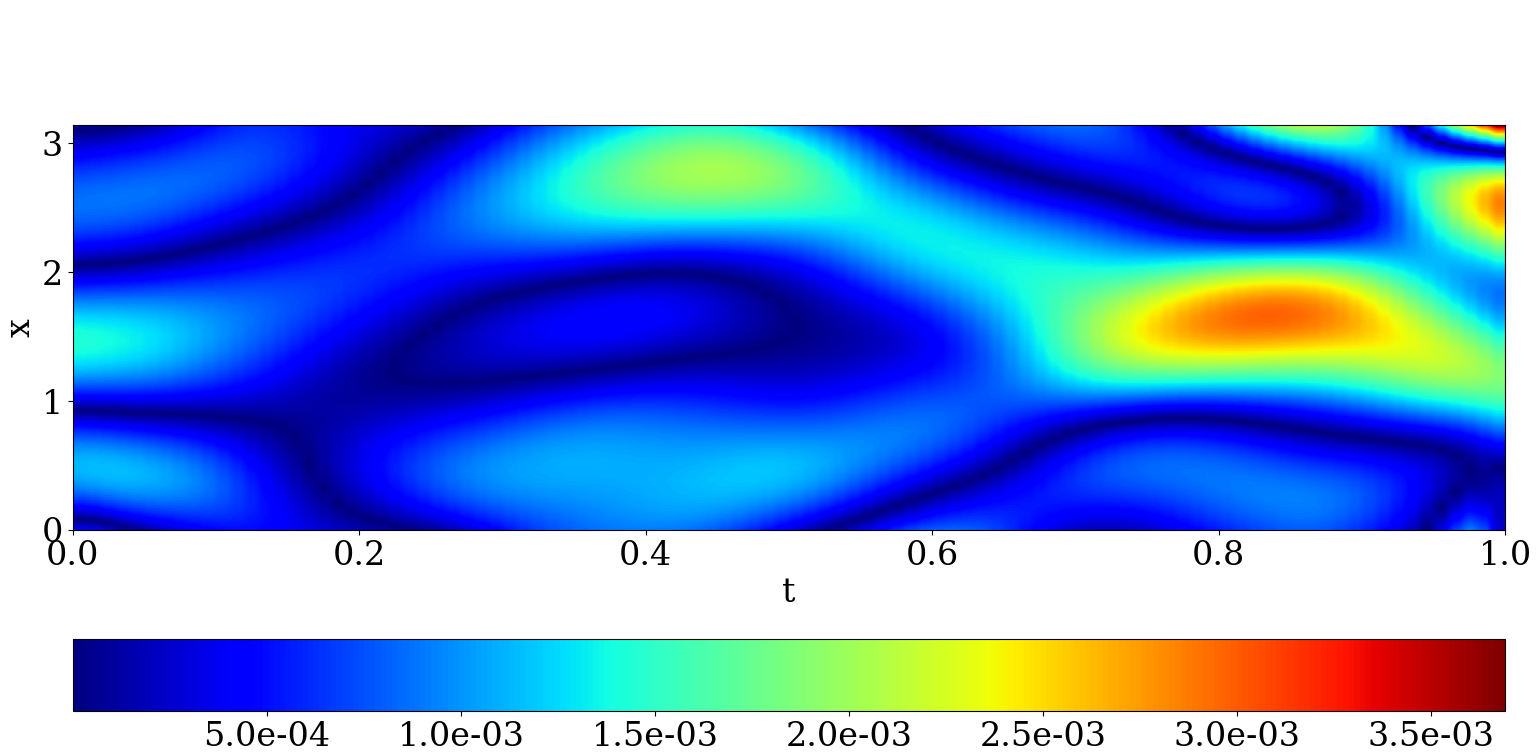}
  \caption{A-PINN}
\end{subfigure}\hfill
\begin{subfigure}[t]{0.48\textwidth}
  \centering
  \includegraphics[width=0.95\linewidth,height=4.9cm]{ 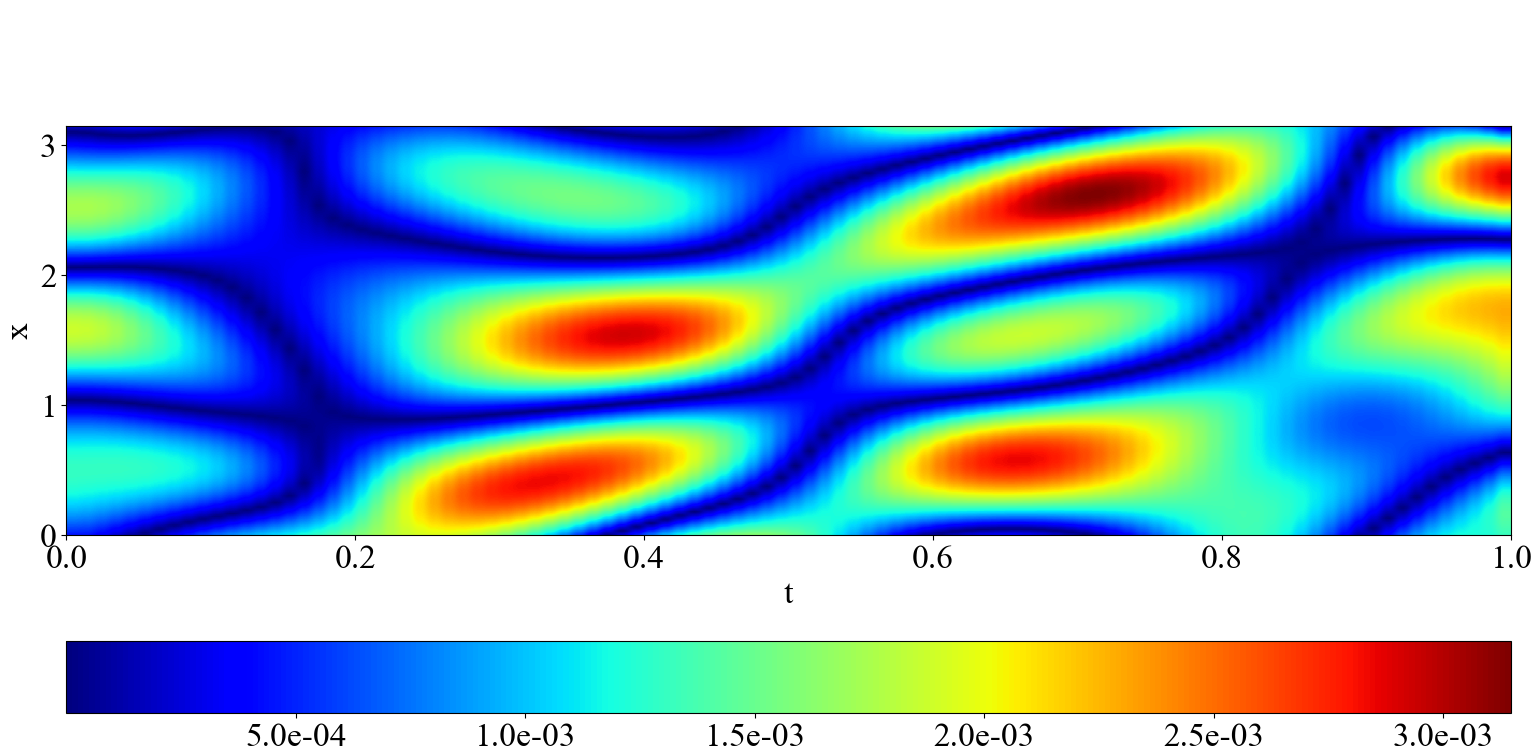}
  \caption{PINN}
\end{subfigure}
\\[0.5em]

\begin{subfigure}[t]{0.48\textwidth}
  \centering
  \includegraphics[width=0.95\linewidth,height=4.9cm]{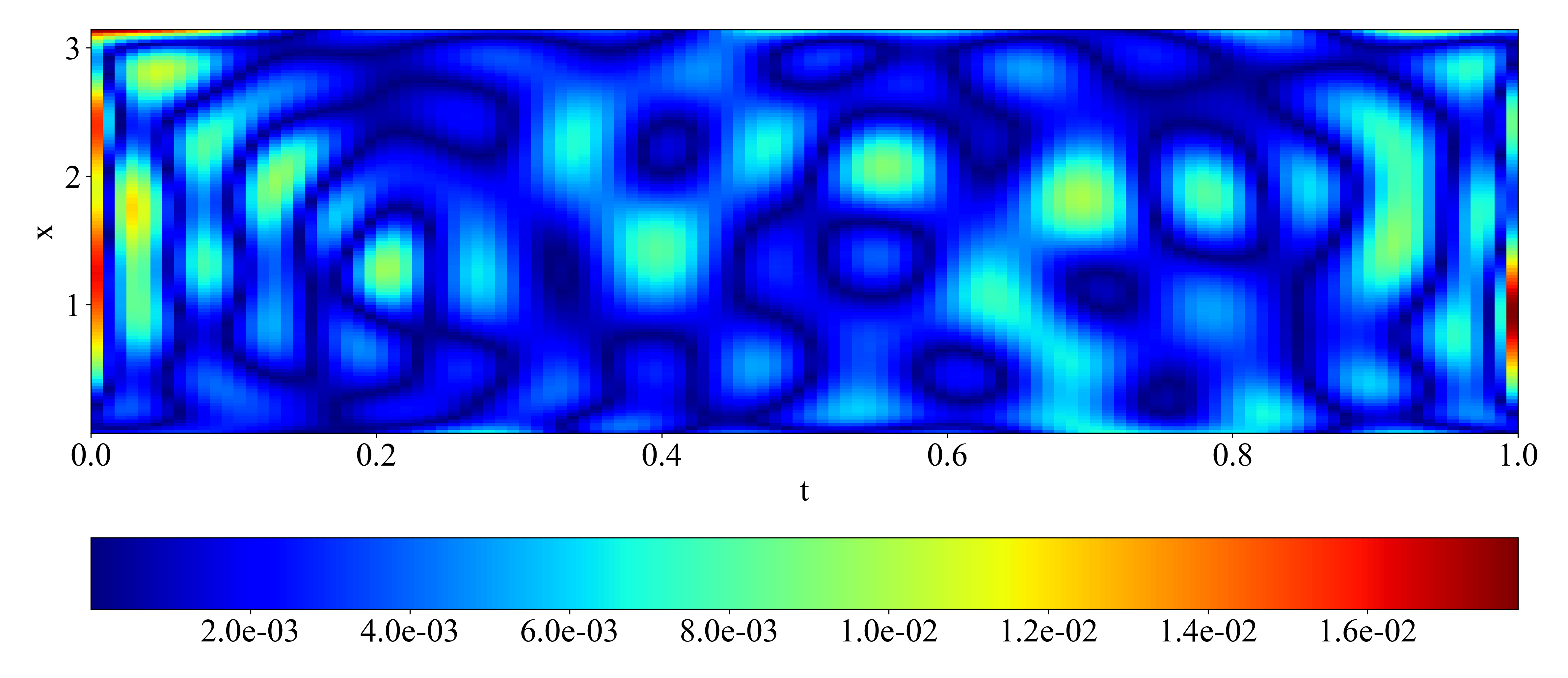}
  \caption{SANN}
\end{subfigure}\hfill
\begin{subfigure}[t]{0.48\textwidth}
  \centering
  \includegraphics[width=0.95\linewidth,height=4.9cm]{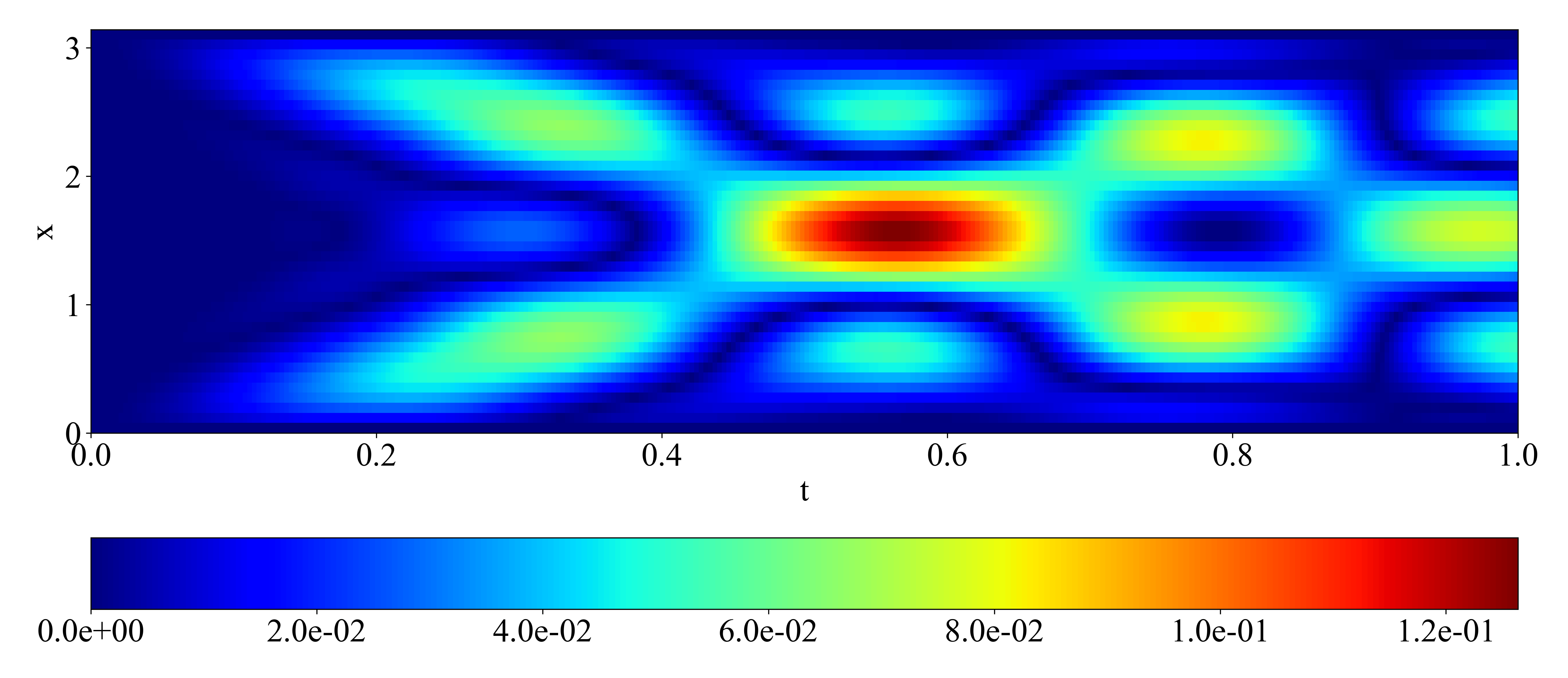}
  \caption{FDM}
\end{subfigure}
\\[1em]

\begin{subfigure}[t]{0.48\textwidth}
  \centering
  \includegraphics[width=0.95\linewidth,height=4cm]{ 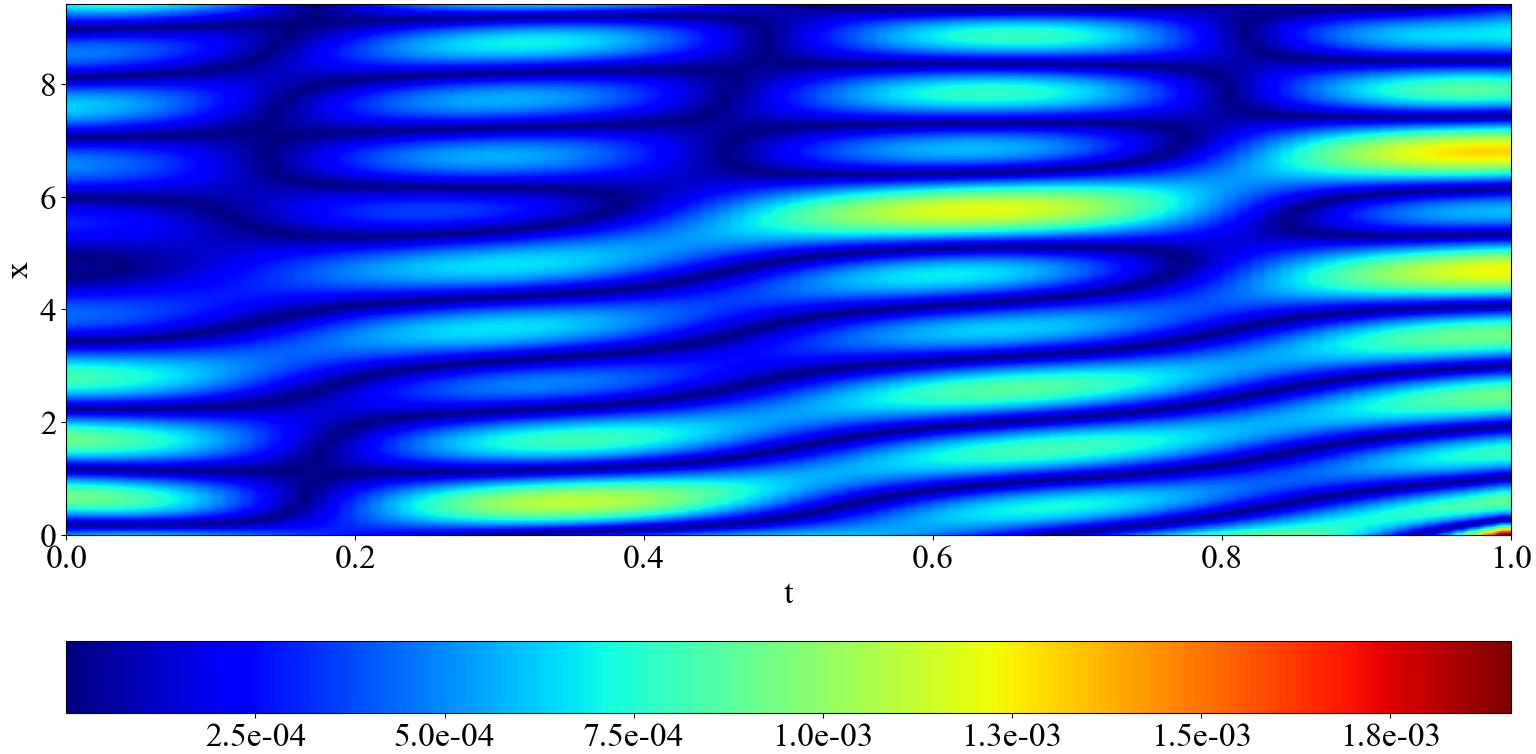}
  \caption{A-PINN}
\end{subfigure}\hfill
\begin{subfigure}[t]{0.48\textwidth}
  \centering
  \includegraphics[width=0.95\linewidth,height=4cm]{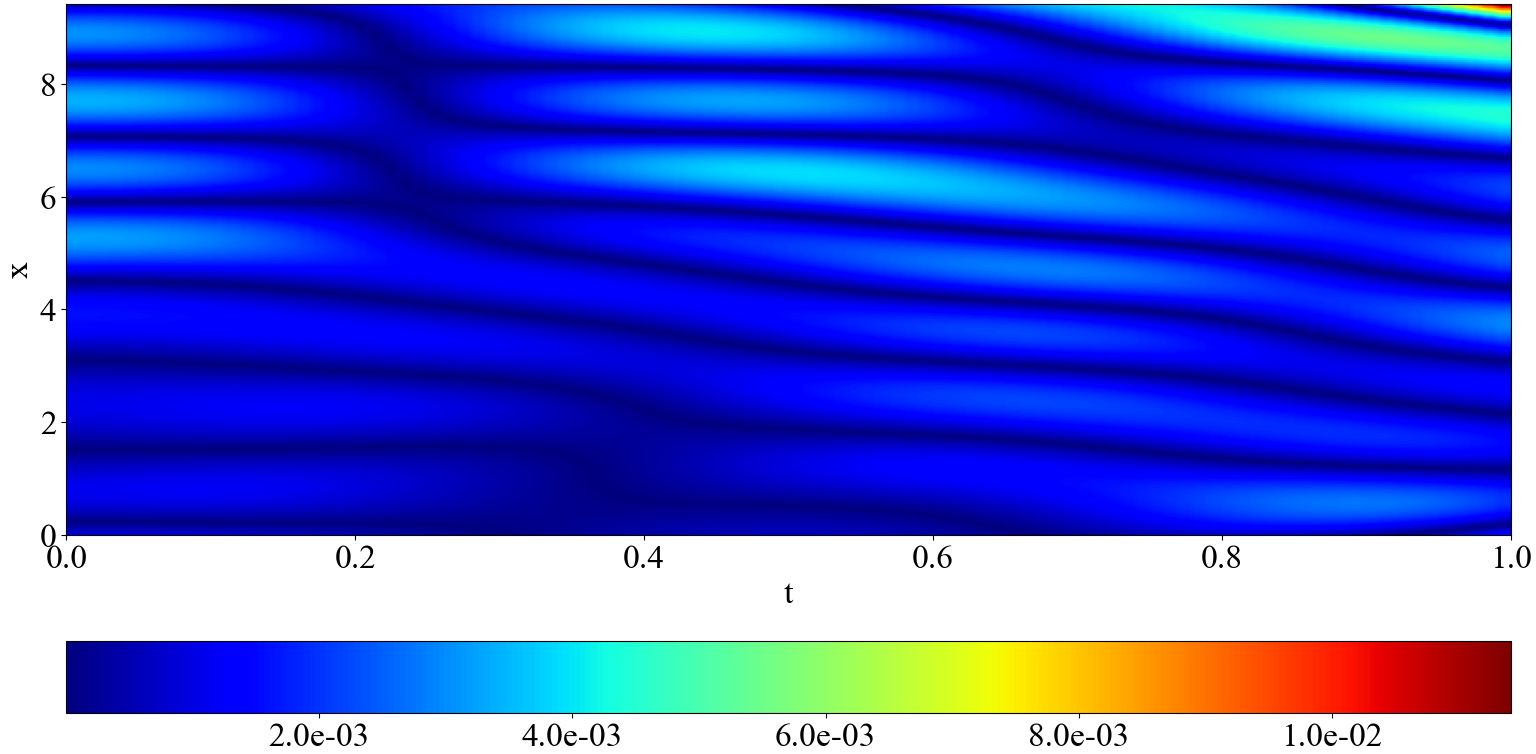}
  \caption{PINN}
\end{subfigure}
\\[0.5em]

\begin{subfigure}[t]{0.48\textwidth}
  \centering
  \includegraphics[width=0.95\linewidth,height=4cm]{ 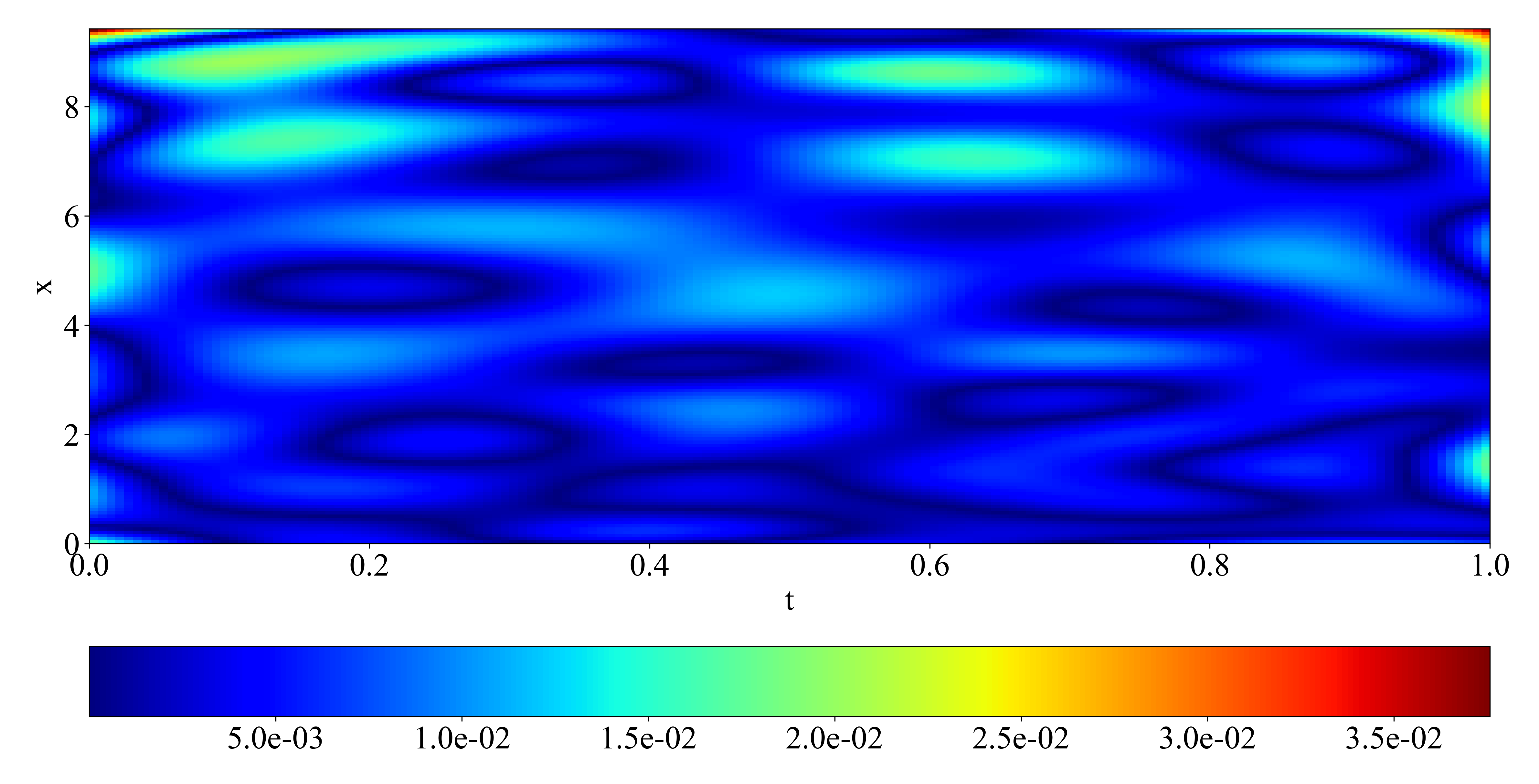}
  \caption{SANN}
\end{subfigure}\hfill
\begin{subfigure}[t]{0.48\textwidth}
  \centering
  \includegraphics[width=0.95\linewidth,height=4cm]{ 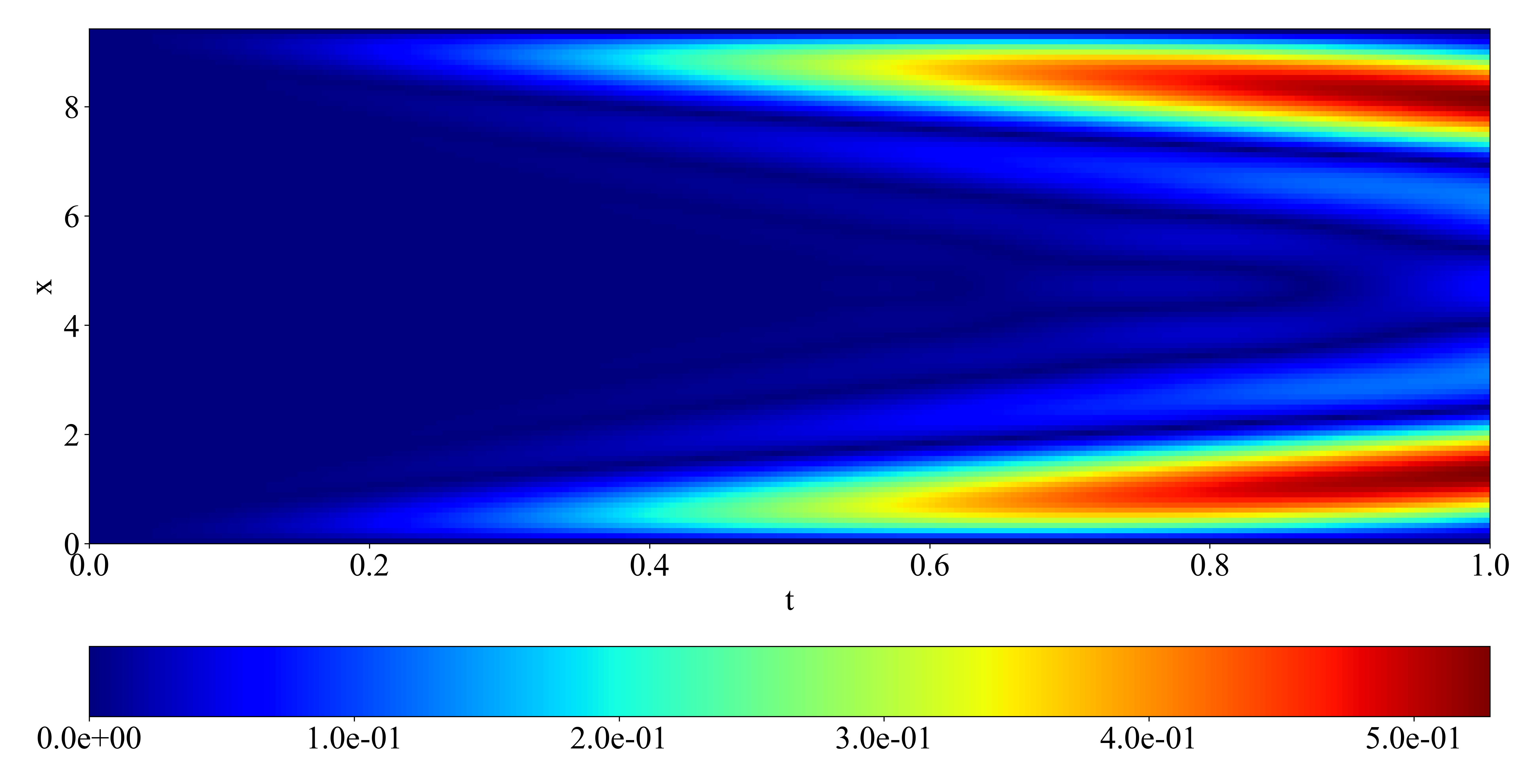}
  \caption{FDM}
\end{subfigure}

\caption{2D absolute error distributions for three benchmark vibration problems:
(a–d) Undamped free vibration, (e–h) Undamped force vibration, (i–l) Undamped force vibration in Winkler foundation. Each row shows A-PINN, PINN, SANN, and FDM.}
\label{ALL_abs_Error}
\end{figure}

Consistent with the tabulated error metrics, the spatial distributions of $E_1$ shown in Figure~\ref{ALL_abs_Error} clearly demonstrate the superior predictive accuracy of the A-PINN framework. Across all three EBB models, the A-PINN demonstrates the low error intensities throughout the spatio-temporal domain, with mean $E_1$ values consistently of the order of $10^{-3}$. 
This uniform magnitude of error highlights the model’s robustness and its ability to maintain stability and precision across different vibration problems. In contrast, the PINN and SANN reveal significantly higher error amplitudes in the range of $10^{-2}$ to $10^{-1}$, particularly near the beam supports and boundary areas where the residual field amplifies. 
The FDM results also display grid-induced discontinuities and oscillatory patterns along the edges, with magnitudes typically between $10^{-2}$ and $10^{-1}$. The meshless nature of the A-PINN effectively suppresses discretization artifacts, producing smooth and physically consistent error distributions across the entire domain. Furthermore, the residuals corresponding to the initial, boundary, and auxiliary constraints remain consistently below the order of $10^{-5}$, and even reach the $10^{-6}$ level for the P3, as is shown by the training losses in Table~\ref{loss_epochs}. This demonstrates that the A-PINN framework enforces the governing physics with high fidelity, achieving the stable and physically consistent convergence across all EBB cases.

To further assess robustness, small perturbations of approximately $\pm5\%$ were introduced in parameters such as the foundation stiffness and forcing amplitude. The trained A-PINN maintained its predictive accuracy with negligible variation in all error norms without requiring retraining, while the performance of PINN and FDM deteriorated notably. This observation highlights the excellent generalization of the A-PINN approach. Overall, the A-PINN achieves roughly an order of magnitude lower errors compared to PINN and two to four orders lower than the SANN and FDM methods, while converging nearly 30–40\% faster. Moreover, we get an average of 99 \% improvement over FDM, 95\% improvement over SANN and 75\% improvement over PINN.
These results collectively demonstrate that the A-PINN is a robust, meshless, and physically interpretable DL framework capable of accurately solving higher-order vibration and structural dynamics problems governed by PDEs.


\begin{table}[ht]
\centering
\caption{Comparison of training loss values for NN models.}
\label{loss_epochs}
\renewcommand{\arraystretch}{1.35} 
\setlength{\tabcolsep}{8pt} 
\begin{tabular}{lccc}
\toprule
\textbf{Problem} & \textbf{SANN} & \textbf{PINN} & \textbf{A-PINN} \\ 
\midrule
P1 & 2.8054 $\times$10$^{0}$ & 9.1991$\times$10$^{-5}$ & 5.0695$\times$10$^{-5}$ \\[2pt]

P2 & 2.8956$\times$10$^{-5}$ & 4.1122$\times$10$^{-5}$ & 5.2000$\times$10$^{-5}$ \\[2pt]

P3 & 4.5741$\times$10$^{-5}$ & 3.3862$\times$10$^{-5}$ & 6.7722$\times$10$^{-6}$ \\ 
\bottomrule
\end{tabular}
\end{table}

\section{Conclusion}\label{sec7}

As an alternative to traditional numerical-based formulations, prior research had data-driven NNs to address the limitations inherent in numerical models. These can accurately capture the complex dynamic behavior of structural dynamics within their training domain; however, their predictive performance deteriorates dramatically when applied outside that domain. To mitigate this limitation, recent works have employed PINNs. By incorporating physics loss components, PINNs integrate given domain knowledge with the expressive capacity of NNs, thereby enhancing model applicability without compromising the accuracy. Nonetheless, their generalization capability across the domains remains restricted by the dataset distribution that governs the loss function. To address this, A-PINNs have been introduced, leveraging a balanced adaptive optimizer and an auxiliary loss to enhance the stability and accuracy of PINNs. Applied to the EBB problem with simply supported beams, our approach successfully EBB models while limiting long-term error accumulation, highlighting its potential for reliable and efficient physics-informed models. The results highlight the strong potential of PINNs in solving high-order PDEs and demonstrate that the A-PINN framework achieves an average reduction of nearly 80--90\% in global error norms and converges approximately 30--40\% faster than PINN, while maintaining stable training behavior across all three vibration scenarios. The $E_1$ generally lies around the order of $10^{-4}$, with localized peaks reaching up to $10^{-3}$. The $E_2$ values stabilize within the range of $10^{-6}$ to $10^{-7}$, whereas the $E_3$ and $E_4$ metrics remain ordered $10^{-3}$ to $10^{-2}$ across all EBB models, reflecting highly accurate, smooth, and physically consistent approximations of the analytical beam solutions.


The introduction of auxiliary variables allows for the better representation of higher-order spatial derivatives, whereas adaptive loss balancing ensures stable optimization between physics, boundary, and auxiliary residuals. Consequently, the predicted displacement remains smooth, physically meaningful, and close to the GT, confirming the enhanced numerical stability, convergence efficiency, and physical fidelity of the A-PINN model.

In exchange for its enhanced accuracy and stability, the A-PINN framework leads to a higher computational burden arising from the NN architecture and auxiliary optimization terms. The performance may further depend on the appropriate hyperparameters, such as hidden neurons, learning rate, and number of collocation points. Moreover, the present investigation focuses on linear, undamped vibration cases; further validation is required for nonlinear or damped systems under complex BCs. Future work will extend the framework to nonlinear and damped beam vibrations to assess scalability under complex dynamic environments. Integrating adaptive collocation sampling and automated hyperparameter tuning could substantially improve computational efficiency. The combination of A-PINN with operator-learning paradigms such as DeepONet or fourier neural operators may also facilitate fast and generalized modeling for multi-physics and real-time vibration analysis. Incorporating uncertainty quantification and error-controlled learning strategies will further enhance the reliability and applicability of the framework for large-scale structural dynamics problems.


\section*{Conflict of Interest Statement}
The authors confirm that they have no conflicts of interest related to this paper.

\section*{Funding Declaration}
The authors confirm that they did not receive any funding to carry out this work.

\section*{Authors’ Contributions}\justifying
\textbf{S.S:} Conceptualization, Visualization, Software, Experiment,  Writing--original draft \& editing. \textbf{R.K.V:} Supervision, Investigation, Formal analysis, Validation,  Writing-review \& editing. \textbf{A.K.S:} Software, Methodology, Experiment, Validation, Writing-original draft

\section*{Data Availability Statement}
No datasets were used or generated during the preparation of this article.

\appendix
\section*{Appendix}
\addcontentsline{toc}{section}{Appendix}

This appendix presents two complementary methods that provide additional context for the development and validation of our A-PINN framework. The first is the classical FDM, a well-established numerical technique for approximating solutions of DEs, including the EBB model. The second is the SANN, a physics-based DL approach specifically designed to preserve the geometric and energy-conserving structures inherent in dynamical systems.

While FDM serves as a benchmark classical discretization scheme, SANN demonstrates how structure-preserving NN architectures can achieve stable and physically consistent approximations. Together, these methods provide a broader perspective for evaluating the performance of the A-PINN framework.

\section{Finite Difference Method}
In FDM, the computational domain is discretized by introducing a uniform spatial grid of size $\Delta x$ and a temporal grid with the step $\Delta t$. The continuous spatial derivatives are replaced by appropriate finite-difference approximations. Specifically, the second and fourth-order derivatives in the spatial variable $x$ are approximated as follows~\cite{sahoo2025cl}:

\begin{equation}
\frac{\partial^2 u}{\partial x^2}\Big|_{x=x_i} \approx 
\frac{u_{i+1}(t) - 2u_i(t) + u_{i-1}(t)}{(\Delta x)^2},
\end{equation}
\begin{equation}
\frac{\partial^4 u}{\partial x^4}\Big|_{x=x_i} \approx 
\frac{u_{i-2}(t) - 4u_{i-1}(t) + 6u_i(t) - 4u_{i+1}(t) + u_{i+2}(t)}{(\Delta x)^4},
\end{equation}
where $u_i(t) \equiv u(x_i,t)$ denotes the displacement at grid point $x_i$. Similarly, the second derivative in time is approximated as
\begin{equation}
\frac{\partial^2 u}{\partial t^2}\Big|_{t=t_j} \approx 
\frac{u(x,t_{j+1}) - 2u(x,t_j) + u(x,t_{j-1})}{(\Delta t)^2}.
\end{equation}
Substituting these approximations into the governing EBB equation yields a system of algebraic equations that can be advanced iteratively to compute $u(x,t)$ at all grid points.

Although FDM is conceptually straightforward and widely applicable, its accuracy depends strongly on the discretization size. Fine grids are often required to resolve higher-order derivatives, leading to large algebraic systems and increased computational cost. Additionally, stability issues may arise in long-time simulations, particularly for the EBB models. These limitations motivate physics-informed such as PINN and A-PINN, which reduce reliance on dense discretization while enforcing the beam physics.

\section{Symplectic Artificial Neural Network}
Let $\phi_\tau(x,t;\theta)$ denote the ANN-based approximation of the displacement field $u(x,t)$ represented by a NN with parameters $\theta$. The residual of the governing beam equation is
\begin{equation}
R(x,t;\theta) = E I\,\frac{\partial^4 \phi_\tau}{\partial x^4}
+ \rho A\,\frac{\partial^2 \phi_\tau}{\partial t^2}
+ k\,\phi_\tau - F(x,t),
\label{residual_beam}
\end{equation}
where $E$ is the Young’s modulus, $I$ the second moment of area, $\rho$ the density, $A$ the cross-sectional area, $k$ the stiffness of the Winkler foundation, and $F(x,t)$ the applied external force.

Training seeks parameters that minimize the residual at collocation points $(x_i,t_i)$:
\begin{equation}
\theta^\star = \arg\min_{\theta} \sum_{(x_i,t_i)\in\overline{D}} \big[ R(x_i,t_i;\theta) \big]^2,
\label{cost_beam}
\end{equation}
where $\overline{D}$ is the set of a collection of training points distributed over the space–time domain.

To enforce prescribed ICs/BCs, the approximate solution is written as \cite{chakraverty2025artificial}
\begin{equation}
\phi_\tau(x,t;\theta) = \alpha(x,t) + \Gamma\!\big(x,t,\,\mathrm{SANN}(x,t;\theta)\big),
\label{beam_phi}
\end{equation}
where $\alpha(x,t)$ is a trial function that satisfies the constraints exactly, while $\mathrm{SANN}(x,t;\theta)$ represents the symplectic network output responsible for learning the beam dynamics consistent with~\eqref{residual_beam}. The corresponding results are reported in Section \ref{sec5} and compared with the A-PINN framework.


\begin{table}[H]
    \centering
    \caption{Comparison among the proposed model and baselines at $t = 0.5$ and $t = 0.9$ for P1}
    \label{Apndx_Table_(P1)}
    \scriptsize
    \begin{tabular}{c ccc ccc}
        \toprule
        \multirow{2}{*}{$x$} & \multicolumn{3}{c}{At $t = 0.5$} & \multicolumn{3}{c}{At $t = 0.9$} \\
        \cmidrule(lr){2-4} \cmidrule(lr){5-7}
         & GT & $E_1$ (FDM) & $E_1$ (SANN) & GT & $E_1$ (FDM) & $E_1$ (SANN) \\
        \midrule
        0.00 & 0.000000 & 0.000000e+00 & 0.000000e+00 &  0.000000 & 0.000000e+00 & 0.000000e+00 \\
        0.10 & 0.068164 & 7.685920e-03 & 1.072866e-02 & -0.264707 & 7.817564e-03 & 7.141103e-03 \\
        0.20 & 0.129656 & 1.463649e-02 & 2.040712e-02 & -0.503502 & 1.494212e-02 & 1.358319e-02 \\
        0.30 & 0.178456 & 2.016217e-02 & 2.808799e-02 & -0.693012 & 2.063729e-02 & 1.869565e-02 \\
        0.40 & 0.209788 & 2.371391e-02 & 3.301942e-02 & -0.814684 & 2.431093e-02 & 2.197806e-02 \\
        0.50 & 0.220584 & 2.493844e-02 & 3.471867e-02 & -0.856610 & 2.557969e-02 & 2.310910e-02 \\
        0.60 & 0.209788 & 2.371391e-02 & 3.301942e-02 & -0.814684 & 2.431093e-02 & 2.197806e-02 \\
        0.70 & 0.178456 & 2.016217e-02 & 2.808799e-02 & -0.693012 & 2.063729e-02 & 1.869565e-02 \\
        0.80 & 0.129656 & 1.463649e-02 & 2.040712e-02 & -0.503502 & 1.494212e-02 & 1.358319e-02 \\
        0.90 & 0.068164 & 7.685920e-03 & 1.072866e-02 & -0.264707 & 7.817564e-03 & 7.141103e-03 \\
        1.00 & 0.000000 & 0.000000e+00 & 0.000000e+00 &  0.000000 & 0.000000e+00 & 0.000000e+00 \\
        \bottomrule
    \end{tabular}
\end{table}

\begin{table}[H]
    \centering
    \caption{Comparison among the proposed model and baselines at $t = 0.4$ and $t = 0.8$ for P2}
    \label{Apndx_Table_(P2)}
    \scriptsize
    \begin{tabular}{c ccc ccc}
        \toprule
        \multirow{2}{*}{$x$} & \multicolumn{3}{c}{At $t = 0.4$} & \multicolumn{3}{c}{At $t = 0.8$} \\
        \cmidrule(lr){2-4} \cmidrule(lr){5-7}
         &  GT & $E_1$ (FDM) & $E_1$ (SANN) & GT & $E_1$ (FDM) & $E_1$ (SANN) \\
        \midrule
        0.00 &  0.000000 & 0.000000e+00 & 4.047181e-04 &  0.000000 & 0.000000e+00 & 4.136627e-03 \\
        0.31 &  0.095492 & 5.536938e-03 & 7.423084e-04 & -0.250000 & 1.949568e-03 & 1.666048e-03 \\
        0.63 &  0.181636 & 4.243004e-02 & 1.536017e-03 & -0.475528 & 5.921781e-02 & 1.230216e-03 \\
        0.94 &  0.250000 & 5.773069e-02 & 1.819511e-03 & -0.654508 & 7.309885e-02 & 3.057164e-03 \\
        1.26 &  0.293893 & 4.210958e-02 & 4.923562e-03 & -0.769421 & 2.767040e-02 & 7.305961e-04 \\
        1.57 &  0.309017 & 2.890874e-02 & 4.137578e-03 & -0.809017 & 4.044077e-03 & 4.892847e-03 \\
        1.88 &  0.293893 & 4.210958e-02 & 1.309016e-03 & -0.769421 & 2.767040e-02 & 6.823044e-03 \\
        2.20 &  0.250000 & 5.773069e-02 & 5.090236e-03 & -0.654508 & 7.309885e-02 & 2.106177e-03 \\
        2.51 &  0.181636 & 4.243004e-02 & 1.492469e-03 & -0.475528 & 5.921781e-02 & 1.192766e-03 \\
        2.83 &  0.095492 & 5.536938e-03 & 1.780536e-03 & -0.250000 & 1.949568e-03 & 2.395861e-04 \\
        3.14 &  0.000000 & 0.000000e+00 & 2.172169e-03 &  0.000000 & 0.000000e+00 & 4.831877e-03 \\
        \bottomrule
    \end{tabular}
\end{table}

\begin{table}[H]
    \centering
    \caption{Comparison among the proposed model and baselines at $t = 0.3$ and $t = 0.9$ for P3}
    \label{Apndx_Table_(P3)}
    \scriptsize
    \begin{tabular}{c ccc ccc}
        \toprule
        \multirow{2}{*}{$x$} & \multicolumn{3}{c}{At $t = 0.3$} & \multicolumn{3}{c}{At $t = 0.9$} \\
        \cmidrule(lr){2-4} \cmidrule(lr){5-7}
         & GT & $E_1$ (FDM) & $E_1$ (SANN) & GT & $E_1$ (FDM) & $E_1$ (SANN) \\
        \midrule
        0.00 &  0.000000 & 0.000000e+00 & 3.725129e-04 &  0.000000 & 0.000000e+00 & 7.750684e-03 \\
        0.94 &  0.475528 & 4.404559e-02 & 2.094864e-03 & -0.769421 & 4.994990e-01 & 2.639728e-03 \\
        1.88 &  0.559017 & 3.161000e-03 & 3.109684e-03 & -0.904508 & 2.154100e-01 & 3.094135e-03 \\
        2.83 &  0.181636 & 8.600000e-05 & 3.273516e-03 & -0.293893 & 1.201100e-01 & 5.860095e-03 \\
        3.77 & -0.345492 & 1.843000e-03 & 7.008275e-03 &  0.559017 & 8.602000e-03 & 2.448282e-03 \\
        4.71 & -0.587785 & 3.015000e-03 & 8.654133e-04 &  0.951057 & 1.055400e-02 & 9.168675e-03 \\
        5.65 & -0.345492 & 1.843000e-03 & 1.102804e-02 &  0.559017 & 8.602000e-03 & 7.958598e-03 \\
        6.60 &  0.181636 & 8.600000e-05 & 1.524239e-03 & -0.293893 & 1.201100e-01 & 1.926272e-03 \\
        7.54 &  0.559017 & 3.161000e-03 & 7.521665e-03 & -0.904508 & 2.154100e-01 & 1.568183e-03 \\
        8.48 &  0.475528 & 4.404559e-02 & 6.746654e-03 & -0.769421 & 4.994990e-01 & 4.155456e-03 \\
        9.42 &  0.000000 & 0.000000e+00 & 7.299128e-03 &  0.000000 & 0.000000e+00 & 1.706771e-02 \\
        \bottomrule
    \end{tabular}
\end{table}

\bibliographystyle{elsarticle-num} 
\bibliography{References.bib} 

@article{sahoo2024unsupervised,
  title={An unsupervised scientific machine learning algorithm for approximating displacement of object in mass-spring-damper systems},
  author={Sahoo, Arup Kumar and Kumar, Sandeep and Chakraverty, Snehashish},
  journal={IEEE Access},
  year={2024},
  publisher={IEEE}
}

@article{kapoor2024transfer,
  title={Transfer learning for improved generalizability in causal physics-informed neural networks for beam simulations},
  author={Kapoor, Taniya and Wang, Hongrui and N{\'u}{\~n}ez, Alfredo and Dollevoet, Rolf},
  journal={Engineering Applications of Artificial Intelligence},
  volume={133},
  pages={108085},
  year={2024},
  publisher={Elsevier}
}

@article{alli2003solutions,
  title={The solutions of vibration control problems using artificial neural networks},
  author={Alli, Hasan and U{\c{c}}ar, Ay{\c{s}}eg{\"u}l and Demir, Yakup},
  journal={Journal of the Franklin Institute},
  volume={340},
  number={5},
  pages={307--325},
  year={2003},
  publisher={Elsevier}
}

@book{rao2019vibration,
  title        = {Vibration of Continuous Systems},
  author       = {Rao, Singiresu S.},
  year         = {2019},
  publisher    = {John Wiley \& Sons},
  address      = {Hoboken, New Jersey}
}

@article{deng2023dynamic,
  title={Dynamic stability and responses of beams on elastic foundations under a parametric load},
  author={Deng, Jian and Shahroudi, Mohammadmehdi and Liu, Kefu},
  journal={International Journal of Structural Stability and Dynamics},
  volume={23},
  number={02},
  pages={2350018},
  year={2023},
  publisher={World Scientific}
}

@article{kumar2025comparative,
  title={A comparative study of Center-Radius and Lower-Upper type interval neural network methods in uncertainty modeling},
  author={Kumar, Sandeep and Chakraverty, S and Sethi, Narayan},
  journal={Applied Soft Computing},
  pages={113347},
  year={2025},
  publisher={Elsevier}
}

@book{chakraverty2025artificial,
  title={Artificial Neural Networks and Type-2 Fuzzy Set: Elements of Soft Computing and Its Applications},
  author={Chakraverty, Snehashish and Sahoo, Arup Kumar and Mohapatra, Dhabaleswar},
  year={2025},
  publisher={Elsevier}
}

@book{goodfellow2016deep,
  title={Deep learning},
  author={Goodfellow, Ian and Bengio, Yoshua and Courville, Aaron and Bengio, Yoshua},
  volume={1},
  number={2},
  year={2016},
  publisher={MIT press Cambridge}
}

@article{raissi2019physics,
  title={Physics-informed neural networks: A deep learning framework for solving forward and inverse problems involving nonlinear partial differential equations},
  author={Raissi, Maziar and Perdikaris, Paris and Karniadakis, George E},
  journal={Journal of Computational Physics},
  volume={378},
  pages={686--707},
  year={2019},
  publisher={Elsevier}
}

@article{cai2021physics,
  title={Physics-informed neural networks {(PINNs)} for fluid mechanics: A review},
  author={Cai, Shengze and Mao, Zhiping and Wang, Zhicheng and Yin, Minglang and Karniadakis, George Em},
  journal={Acta Mechanica Sinica},
  volume={37},
  number={12},
  pages={1727--1738},
  year={2021},
  publisher={Springer}
}

@article{chen2023second,
  title={Second-order analysis of beam-columns by machine learning-based structural analysis through physics-informed neural networks},
  author={Chen, Liang and Zhang, Hao Yi and Liu, Si Wei and Chan, Siu Lai},
  journal={Advanced Steel Construction},
  volume={19},
  number={4},
  pages={411--420},
  year={2023},
  publisher={Hong Kong Institute of Steel Construction}
}

@article{roquemen2025recent,
  title={Recent advancements and applications of physics-informed machine learning in biomedical research},
  author={R. Echeverri, Valentina and M. Lopez, Clara},
  journal={Current Opinion in Biomedical Engineering},
  volume={35},
  pages={100612},
  year={2025},
  publisher={Elsevier}
}

@article{chen2020physics,
  title={Physics-informed neural networks for inverse problems in nano-optics and metamaterials},
  author={Chen, Yuyao and Lu, Lu and Karniadakis, George Em and Dal Negro, Luca},
  journal={Optics Express},
  volume={28},
  number={8},
  pages={11618--11633},
  year={2020},
  publisher={OSA}
}

@article{cheng2025exponential,
  title={Exponential stabilization and discrete approximation for axially moving {Euler--Bernoulli} beam under nonlinear boundary control},
  author={Cheng, Yi and Wu, Yingnan and Guo, Bao Zhu},
  journal={SIAM Journal on Control and Optimization},
  volume={63},
  number={1},
  pages={227--255},
  year={2025},
  publisher={SIAM}
}

@article{kharazmi2021hp,
  title={{hp-VPINNs}: Variational physics-informed neural networks with domain decomposition},
  author={Kharazmi, Ehsan and Zhang, Zhongqiang and Karniadakis, George Em},
  journal={Computer Methods in Applied Mechanics and Engineering},
  volume={374},
  pages={113547},
  year={2021},
  publisher={Elsevier}
}

@article{zhou2024data,
  title={Data-guided physics-informed neural networks for solving inverse problems in partial differential equations},
  author={Zhou, Wei and Xu, YF},
  journal={arXiv preprint arXiv:2407.10836},
  year={2024}
}

@article{ye2025mixed,
  title={A Mixed Finite Element Method for Vibration Equations of Structurally Damped Beam and Plate},
  author={Ye, Yuqian and Yin, Zhe and Zhu, Ailing},
  journal={Numerical Methods for Partial Differential Equations},
  volume={41},
  number={2},
  pages={e70005},
  year={2025},
  publisher={Wiley Online Library}
}

@article{xiao2025meshless,
  title={A meshless {Runge-Kutta}-based Physics-Informed Neural Network framework for structural vibration analysis},
  author={Xiao, Shusheng and Bai, Jinshuai and Jeong, Hyogu and Alzubaidi, Laith and Gu, Yuantong},
  journal={Engineering Analysis with Boundary Elements},
  volume={170},
  pages={106054},
  year={2025},
  publisher={Elsevier}
}

@book{jena2024structural,
  title={Structural dynamics in uncertain environments: micro, nano, and functionally graded beam analysis},
  author={Jena, Subrat Kumar and Chakraverty, Snehashish},
  year={2024},
  publisher={CRC Press}
}

@article{olawale2023response,
  title={Response of an {Euler--Bernoulli} beam subject to a stochastic disturbance},
  author={Olawale, Lukman and Gao, Tao and George, Erwin and Lai, Choi Hong},
  journal={Engineering with Computers},
  volume={39},
  number={6},
  pages={4185--4197},
  year={2023},
  publisher={Springer}
}

@inproceedings{kapoor2024physics,
  title={Physics-informed machine learning for moving load problems},
  author={Kapoor, Taniya and Wang, Hongrui and N{\'u}{\~n}ez, Alfredo and Dollevoet, Rolf},
  booktitle={Journal of Physics: Conference Series},
  volume={2647},
  number={15},
  pages={152003},
  year={2024},
  organization={IOP Publishing}
}

@article{xu2023transfer,
  title={Transfer learning based physics-informed neural networks for solving inverse problems in engineering structures under different loading scenarios},
  author={Xu, Chen and Cao, Ba Trung and Yuan, Yong and Meschke, G{\"u}nther},
  journal={Computer Methods in Applied Mechanics and Engineering},
  volume={405},
  pages={115852},
  year={2023},
  publisher={Elsevier}
}

@article{zhang2020differences,
  title={Differences between {Euler-Bernoulli and Timoshenko} beam formulations for calculating the effects of moving loads on a periodically supported beam},
  author={Zhang, Xianying and Thompson, David and Sheng, Xiaozhen},
  journal={Journal of Sound and Vibration},
  volume={481},
  pages={115432},
  year={2020},
  publisher={Elsevier}
}

@article{kumar2025deep,
  title={A deep kolmogorov--arnold network framework for solving time--fractional partial differential equations},
  author={Kumar, Yatin and Vats, Ramesh Kumar},
  journal={Applied Soft Computing},
  pages={114379},
  year={2025},
  publisher={Elsevier}
}

@inproceedings{kumar2023physics,
  title={Physics-informed machine learning framework for approximating the modified degasperis-procesi equation},
  author={Kumar, Sandeep and Sahoo, Arup Kumar and Chakraverty, S},
  booktitle={2023 International Conference on Ambient Intelligence, Knowledge Informatics and Industrial Electronics (AIKIIE)},
  pages={1--6},
  year={2023},
  organization={IEEE}
}

@article{soyleyici2025physics,
  title={A Physics-Informed Deep Neural Network based beam vibration framework for simulation and parameter identification},
  author={S{\"o}yleyici, Cem and {\"U}nver, Hakk{\i} {\"O}zg{\"u}r},
  journal={Engineering Applications of Artificial Intelligence},
  volume={141},
  pages={109804},
  year={2025},
  publisher={Elsevier}
}

@book{weaver1991vibration,
  title={Vibration Problems in Engineering},
  author={Weaver Jr, William and Timoshenko, Stephen P and Young, Donovan Harold},
  year={1991},
  publisher={John Wiley \& Sons}
}

@article{Rao2024,
	title={Numerical study of nonlinear time-fractional Caudrey--Dodd--Gibbon--Sawada--Kotera equation arising in propagation of waves},
	author={Rao, A. and Vats, R. K. and Yadav, S.},
	journal={Chaos, Solitons \& Fractals},
	volume={184},
	pages={114941},
	year={2024}
}

@book{wu2013analytical,
  title={Analytical and Numerical Methods for Vibration Analyses},
  author={Wu, Jong Shyong},
  year={2013},
  publisher={John Wiley \& Sons}
}

@book{thomas2013numerical,
	title={Numerical Partial Differential Equations: Finite Difference Methods},
	author={Thomas, James William},
	volume={22},
	year={2013},
	publisher={Springer Science \& Business Media}
}

@book{goldman1999vibration,
	title={Vibration Spectrum Analysis: A Practical Approach},
	author={Goldman, Steve},
	year={1999},
	publisher={Industrial Press Inc.}
}

@article{jena2018free,
  title={Free vibration analysis of variable cross-section Single-Layered Graphene Nano-Ribbons {(SLGNRs)} using differential quadrature method},
  author={Jena, Subrat Kumar and Chakraverty, Snehashish},
  journal={Frontiers in Built Environment},
  volume={4},
  pages={63},
  year={2018},
  publisher={Frontiers Media SA}
}

@article{yadav2025application,
  title={Application of extended residual power series method for time-fractional Zakharov--Kuznetsov equations in ocean-based coastal wave},
  author={Yadav, Sanjeev and Vats, Ramesh Kumar and Rao, Anjali},
  journal={Pramana},
  volume={99},
  number={3},
  pages={97},
  year={2025},
  publisher={Springer}
}

@article{sahoo2025cl,
  title={{CL-PINN}: A Physics-informed Neural Networks Framework for Equatorial Tsunami Wave Propagation in Geophysical Ocean},
  author={Sahoo, Arup Kumar and Sahoo, Swarup Kumar and Chakraverty, S},
  journal={IEEE Access},
  year={2025},
  publisher={IEEE}
}

@article{cao2025wbpinn,
  title={{wbPINN}: Weight balanced physics-informed neural networks for multi-objective learning},
  author={Cao, Fujun and Guo, Xiaobin and Dong, Xinzheng and Yuan, Dongfang},
  journal={Applied Soft Computing},
  volume={170},
  pages={112632},
  year={2025},
  publisher={Elsevier}
}

@article{zheng2024vtpinn,
  title={{VT-PINN}: Variable transformation improves physics-informed neural networks for approximating partial differential equations},
  author={Zheng, Jiachun and Yang, Yunlei},
  journal={Applied Soft Computing},
  volume={167},
  pages={112370},
  year={2024},
  publisher={Elsevier}
}

@article{wang2021learning,
  title={Learning the solution operator of parametric partial differential equations with physics-informed DeepONets},
  author={Wang, Sifan and Wang, Hanwen and Perdikaris, Paris},
  journal={Science advances},
  volume={7},
  number={40},
  pages={eabi8605},
  year={2021},
  publisher={American Association for the Advancement of Science}
}

@article{song2024admm,
  title={The ADMM-PINNs algorithmic framework for nonsmooth {PDE}-constrained optimization: a deep learning approach},
  author={Song, Yongcun and Yuan, Xiaoming and Yue, Hangrui},
  journal={SIAM Journal on Scientific Computing},
  volume={46},
  number={6},
  pages={C659--C687},
  year={2024},
  publisher={SIAM}
}

@article{yu2022gradient,
  title={Gradient-enhanced physics-informed neural networks for forward and inverse problems},
  author={Yu, Chen and Lu, Lu and Meng, Xuhui and Karniadakis, George Em},
  journal={Computer Methods in Applied Mechanics and Engineering},
  volume={389},
  pages={114399},
  year={2022},
  publisher={Elsevier}
}

@article{jagtap2020extended,
  title        = {Extended Physics-Informed Neural Networks {(XPINNs)}: A Generalized Space-Time Domain Decomposition Based Deep Learning Framework for Nonlinear Partial Differential Equations},
  author       = {Jagtap, Ameya D. and Karniadakis, George Em},
  journal      = {Communications in Computational Physics},
  volume       = {28},
  number       = {5},
  pages        = {2002--2041},
  year         = {2020}
}

@book{randall2021vibration,
  title={Vibration-based condition monitoring: industrial, automotive and aerospace applications},
  author={Randall, Robert Bond},
  year={2021},
  publisher={John Wiley \& Sons}
}

@article{skudrzyk1980mean,
  title={The mean-value method of predicting the dynamic response of complex vibrators},
  author={Skudrzyk, Eugen},
  journal={The Journal of the Acoustical Society of America},
  volume={67},
  number={4},
  pages={1105--1135},
  year={1980},
  publisher={Acoustical Society of America}
}

@article{chu2018human,
  title={Human pulse diagnosis for medical assessments using a wearable piezoelectret sensing system},
  author={Chu, Yao and Zhong, Junwen and Liu, Huiliang and Ma, Yuan and Liu, Nathaniel and Song, Yu and Liang, Jiaming and Shao, Zhichun and Sun, Yu and Dong, Ying and Wang, Xiaohao and Lin, Liwei},
  journal={Advanced Functional Materials},
  volume={28},
  number={40},
  pages={1803413},
  year={2018},
  publisher={Wiley Online Library}
}

@article{caetano2011vision,
  title={A VISION SYSTEM FOR VIBRATION MONITORING OF CIVIL ENGINEERING STRUCTURES.},
  author={Caetano, E and Silva, S and Bateira, J},
  journal={Experimental Techniques},
  volume={35},
  number={4},
pages={74--82},
  year={2011}
}

@article{kapoor2023physics,
  title={Physics-informed neural networks for solving forward and inverse problems in complex beam systems},
  author={Kapoor, Taniya and Wang, Hongrui and N{\'u}{\~n}ez, Alfredo and Dollevoet, Rolf},
  journal={IEEE Transactions on Neural Networks and Learning Systems},
  volume={35},
  number={5},
  pages={5981--5995},
  year={2023},
  publisher={IEEE}
}

@article{kumar2025robust,
  title={Robust Deep Learning Framework Using Hermite Interpolation for Time-Dependent Caputo Fractional Differential Equations},
  author={Kumar, Yatin and Kumar Vats, Ramesh and Kumar Nain, Ankit and Yadav, Sanjeev},
  journal={Mathematical Methods in the Applied Sciences},
  year={2025},
  publisher={Wiley Online Library}
}


\end{document}